%% file: main.tex
\documentclass{article}

\usepackage{arxiv} 

\usepackage{multicol}

\usepackage{CJKutf8}

\usepackage[
    backend=biber,
    style=nature,
]{biblatex}
\usepackage{listings}
\usepackage[svgnames]{xcolor} 
\usepackage{tabularx} 

\usepackage{booktabs} 
\usepackage{array}
\usepackage[frozencache,cachedir=minted-cache]{minted}

\newminted[sjson]{json}{
  frame=lines,
  framesep=2mm,
  linenos,
  breaklines,
}

\newminted[mlmd]{text}{
  frame=lines,
  framesep=2mm,
  linenos,
  breaklines,
}

\newminted[stext]{text}{
    breaklines=true,
    breaksymbol=,
}
\usepackage{mdframed}
\newmdenv[nobreak=true]{stextenv}  
\BeforeBeginEnvironment{stext}{\begin{stextenv}}
\AfterEndEnvironment{stext}{\end{stextenv}}

\usepackage{amsmath}

\usepackage{algorithm}
\usepackage{algpseudocode}

\usepackage{url}

\usepackage{float}
\usepackage{graphicx}

\usepackage{caption}
\usepackage{authblk}
\usepackage{bbding}

\usepackage{booktabs}

\usepackage{enumitem} 

\setlist[itemize]{leftmargin=*}
\setlist[enumerate]{leftmargin=*}

\newlist{ruleen}{enumerate}{3}
\setlist[ruleen,1]{label=\textbf{Rule \arabic*: }, leftmargin=*, labelsep=0pt, itemsep=0pt} 
\setlist[ruleen,2]{label=\textbf{Rule \arabic{ruleeni}.\arabic*: }, leftmargin=*, labelsep=0pt, itemsep=0pt}
\setlist[ruleen,3]{label=\textbf{Rule \arabic{ruleeni}.\arabic{ruleenii}.\arabic*: }, leftmargin=*, labelsep=0pt, itemsep=0pt}

\newlist{rulezh}{enumerate}{3}
\setlist[rulezh,1]{label=\textbf{规则\arabic*：}, leftmargin=*, labelsep=0pt, itemsep=0pt} 
\setlist[rulezh,2]{label=\textbf{规则\arabic{rulezhi}.\arabic*：}, leftmargin=*, labelsep=0pt, itemsep=0pt}
\setlist[rulezh,3]{label=\textbf{规则\arabic{rulezhi}.\arabic{rulezhii}.\arabic*：}, leftmargin=*, labelsep=0pt, itemsep=0pt}

\usepackage{pifont}
\newcommand{\cmark}{\makebox[1cm][c]{\ding{52}}} 
\newcommand{\xmark}{\makebox[1cm][c]{\ding{56}}}

\usepackage{tikz}
\usetikzlibrary{positioning, arrows.meta}

\newcommand{\fnrad}{.2}
\usepackage{scrextend,tikz,scalerel}
\newcommand\circled[2][\fnrad]{\kern.5pt\scalerel*{\tikz{\useasboundingbox (-#1,-#1) rectangle (#1,#1);\node{#2};\draw circle[radius=#1];}}{X}\kern.5pt}

\usepackage{bm} 

\usepackage { hyperref } 
\hypersetup { colorlinks=true,allcolors=DarkCyan }

\usepackage{url}

\newcommand{\boldcolor}[1]{\textbf{\textcolor{DarkCyan}{#1}}}

\usepackage[nameinlink]{cleveref}
\crefname{figure}{Fig.}{Fig.}
\crefname{table}{Table}{Table}
\crefname{listing}{Listing}{Listing}

\title{ShennongAlpha: an AI-driven sharing and collaboration platform for intelligent curation, acquisition, and translation of natural medicinal material knowledge}

\author[1, 2, 4, 5, 6]{YANG Zijie}
\author[3, 6]{YIN Yongjing}
\author[2, 4, 6]{KONG Chaojun}
\author[2, 4, 6]{CHI Tiange}
\author[2, 4, 6, \Envelope]{TAO Wufan}
\author[3, 5, 6, \Envelope]{ZHANG Yue}
\author[2, 4, 5, 6, 7, \Envelope]{XU Tian}

\affil[1]{
    Fudan University, Shanghai, China
}
\affil[2]{
    Key Laboratory of Growth Regulation and Translation Research of Zhejiang Province, School of Life Sciences, Westlake University, Hangzhou, Zhejiang, China
}
\affil[3]{
    School of Engineering, Westlake University, Hangzhou, Zhejiang, China
}
\affil[4]{
    Westlake Laboratory of Life Sciences and Biomedicine, Hangzhou, Zhejiang, China.
}
\affil[5]{
    Research Center for Industries of the Future, Westlake University, Hangzhou, Zhejiang, China
}
\affil[6]{
    Shennong Program, Westlake University, Hangzhou, Zhejiang, China
}
\affil[7]{
    Changping Laboratory, Beijing, China
}
\affil[\Envelope]{
    Corresponding authors
    \authorcr
    \{\tt yangzijie, taowufan, zhangyue, xutian\}@westlake.edu.cn
}

\date{}

\begin{document}
\begin{CJK*}{UTF8}{gbsn}

\include{chapter/main}

\include{chapter/fig}

\part*{Supplementary information}
\appendix
\begin{refsection}
\setcounter{figure}{0}
\renewcommand{\thefigure}{S\arabic{figure}}
\setcounter{table}{0} 
\renewcommand{\thetable}{S\arabic{table}}
\setcounter{listing}{0}
\renewcommand{\thelisting}{S\arabic{listing}}

\include{chapter/sfig}
\include{chapter/stable}

\include{chapter/snnmm}
\include{chapter/snnmma}
\include{chapter/mlmd}
\include{chapter/cgs}

\printbibliography[title={Supplementary references}, heading=bibnumbered]
\end{refsection}

\end{CJK*}
\end{document}

%% file: chapter/main.tex
\maketitle

\begin{abstract}
Natural Medicinal Materials (NMMs) have a long history of global clinical applications and a wealth of records and knowledge. Although NMMs are a major source for drug discovery and clinical application, the utilization and sharing of NMM knowledge face crucial challenges, including the standardized description of critical information, efficient curation and acquisition, and language barriers. To address these, we developed ShennongAlpha, an AI-driven sharing and collaboration platform for intelligent knowledge curation, acquisition, and translation. For standardized knowledge curation, the platform introduced a Systematic Nomenclature to enable accurate differentiation and identification of NMMs. More than fourteen thousand Chinese NMMs have been curated into the platform along with their knowledge. Furthermore, the platform pioneered chat-based knowledge acquisition, standardized machine translation, and collaborative knowledge updating. Together, our study represents the first major advance in leveraging AI to empower NMM knowledge sharing, which not only marks a novel application of AI for Science, but also will significantly benefit the global biomedical, pharmaceutical, physician, and patient communities.
\end{abstract}

\vspace{3\baselineskip}

\begin{multicols}{2}

\section{Main}

Natural Medicinal Materials (NMMs) have been a rich reservoir of therapeutic agents \cite{newman2020natural}. Their importance is highlighted by the diversity and biological relevance of the compounds they produce \cite{lautie2020unraveling,chen2018characterization}. The compounds isolated from NMMs or their derivatives have been instrumental in the treatment of various pathological conditions, ranging from infectious diseases to cancer, and continue to serve as a prolific source of novel drug leads \cite{harvey2015re, atanasov2021natural}. In fact, around 50\% of the FDA-approved drugs are natural products or related molecules \cite{newman2020natural}. Furthermore, NMMs have been directly used for therapeutic purposes throughout human history and continue to play a significant role in different societies. The broad clinical application of NMMs has been documented in various countries and regions, including China \cite{china2016}, Japan \cite{motoo2011traditional}, South Korea \cite{park2012traditional}, India \cite{sen2017revival}, Iran \cite{naghizadeh2020unaprod}, Europe \cite{leonti2017traditional, joos2012herbal}, the Americas \cite{dominguez2015mexican, dutra2016medicinal}, Africa \cite{mahomoodally2013traditional, bultum2019etm}, and the Arab region \cite{pan2014historical, azaizeh2010traditional}. For example, Chinese historical records alone documented nearly ten thousand NMMs and their applications \cite{ZhongHuaBenCao1999}. This vast repository of knowledge has proven to be valuable in discovering new therapies \cite{jang1946ch,tu2016artemisinin,zhang1973preliminary}. Therefore, the global utilization, sharing, and collaboration of NMM knowledge are crucial for advancing biomedical and pharmacological research and applications.

However, the great potential of NMMs has not been fully explored despite their diverse chemical reservoirs, rich history of human usage, and vast amounts of accumulated records and knowledge. The biomedical, pharmaceutical, physician, and patient communities face several challenges in utilizing and sharing NMM knowledge (\boldcolor{\cref{fig:sna-intro}a}). At the center of the challenges is the conspicuous absence of a standardized and systematic nomenclature capable of accurately differentiating and identifying each NMM. This magnifies the difficulty of efficiently procuring accurate and reliable knowledge about NMMs. The enriched information and records for NMMs further pose challenges for curation and acquisition. Finally, language barriers also hinder the global dissemination and utilization of NMM knowledge.

The problem of lacking standardized and systematic nomenclature for NMMs can be exemplified by the discovery of the anti-malaria drug artemisinin, which was awarded the 2015 Nobel Prize in Physiology or Medicine. This discovery emerged from historically documented malaria-treating NMM with the Chinese name ``Qing Hao". However, this name could refer to at least six different plants in the genus \textit{Artemisia}. The artemisinin molecule was eventually found only in \textit{Artemisia annua} with the Chinese name of ``Huang Hua Hao", but not in five other plants, including the modern-day botanic plant with the Chinese name of ``Qing Hao" (\textit{Artemisia caruifolia}) \cite{tu2016artemisinin}. Although the \textit{Chinese Pharmacopoeia (2020 edition)} now specifies \textit{Artemisia annua} as the sole NMM suitable for medicinal use \cite{chp2020vol1}, over three-quarters of the NMMs in the current edition still have ambiguous names (\boldcolor{\cref{stable:chp-name}}). For example, the NMM ``Ma Huang" (also known as ``Ephedra" or ``Ephedrae Herba") corresponds to three species: \textit{Ephedra sinica}, \textit{Ephedra intermedia}, or \textit{Ephedra equisetina} (\boldcolor{\cref{fig:sna-intro}c}). The same problem exists for NMMs in other countries. In the \textit{Indian Pharmacopoeia}, for instance, the name ``Acacia" pertains to multiple species within the \textit{Acacia} genus \cite{inp2010vol3}. Similar ambiguity problems also exist regarding the descriptions of the parts used for medicinal purposes and the preparation processes of NMMs (\boldcolor{\cref{stable:chp-name}}).

The absence of a standardized and systematic nomenclature for defining NMMs further complicates both the curation of related information and the acquisition of reliable and consistent knowledge about these NMMs. For example, Wikipedia often provides misleading entries for NMMs. The entry for ``Ephedra" or ``Ma Huang" misleadingly describes ``Ephedra as a medicinal preparation from the plant \textit{Ephedra sinica}" (\boldcolor{\cref{sfig:wikipedia}}). Chat-based platforms like ChatGPT also exacerbate these inaccuracies, as demonstrated by their incorrect assertion that \textit{Ephedra sinica} is the sole species origin for Ephedra (\boldcolor{\cref{sfig:gpt}}). The root cause of these issues lies in the current absence of a standardized and systematic approach to accurately differentiate and identify each unique NMM with the information regarding its species origin, medicinal part, special description, and processing method, and to correctly catalog its related knowledge. Given the vast number of NMMs, along with their extensive and yet dispersed records, it is, therefore, a daunting task to curate the related information using traditional approaches and to facilitate efficient knowledge acquisition. 

The lack of a multilingual platform for NMMs presents a significant barrier for the global biomedical and pharmaceutical communities, as well as physicians and patients. The absence of a standardized and systematic nomenclature also impedes the global sharing and dissemination of knowledge about NMMs. This issue is evident when translating the Chinese sentence regarding the NMM entry ``Ma Huang" (麻黄是一种天然药材) via Google Translate. The resulting translation of ``Ephedra is a natural medicine" implies a single plant rather than the multi-species intricacies of ``Ma Huang" (\boldcolor{\cref{sfig:google}}). Such mistranslations could misguide English-speaking users to use ``Ephedra" for further knowledge searches and potentially lead them to the aforementioned misleading Wikipedia page.

To systematically tackle these challenges, we have developed an AI-driven, web-based sharing platform named ShennongAlpha for intelligent curation, acquisition, and translation of NMM knowledge (\boldcolor{\cref{fig:sna-intro}}). The platform is also an open, collaborative, and evolving hub, which enables the global community to participate and contribute to NMM knowledge. As the first major advance in leveraging AI to empower NMM knowledge sharing, the ShennongAlpha platform currently contains knowledge of more than 14,000 Chinese NMMs. By presenting an exemplary case in this novel AI for Science field, our study has not only propelled AI methodologies and applications but also will significantly accelerate the NMM research and applications for the global biomedical, pharmaceutical, and physician and patient communities.

\section{Results}

\subsection{Systematic Nomenclature defines unique identities and critical information for NMMs}

In order to accurately differentiate and identify each NMM, we introduced a Systematic Nomenclature to define unique identities and critical information for NMMs (\boldcolor{Supplementary text \labelcref{appendix:snnmm,appendix:snnmm_zh}}). The Systematic Nomenclature specifies each NMM with a Systematic Name, a Generic Name, and a unique NMM ID (\boldcolor{\cref{fig:snnmm}}). The Systematic Name is an authoritative designation for NMMs to ensure academic accuracy. It comprises four components: I. the Latin name that defines the species origin of NMM; II. the part of the organism used for medicinal purposes; III. the special description providing information about the initial preparation at the production site and other specific characteristics; and IV. the processing method to produce the NMM for medicinal use (\boldcolor{\cref{fig:snnmm}a}, \boldcolor{\cref{sfig:nmmsn}}). Information for the special description (III) and the processing method (IV) is included when this critical information is necessary. The Generic Name follows a commonly used traditional name, facilitating everyday communication and aiding physicians in prescribing. The unique ID is a 4-digit base-36 code with the prefix “NMM-”, providing a unique identity for each NMM on the digital platform.

For instance, the Systematic Name for ``Jing Pian Jiang Huang" (净片姜黄) is ``Curcuma wenyujin Rhizome Freshly-sliced Cleaned" \cite{nmm-0016} (\boldcolor{\cref{fig:snnmm}c}). This indicates that the NMM ``Jing Pian Jiang Huang" is derived from the species \textit{Curcuma wenyujin}, which utilizes the rhizome as its medicinal part, and is freshly sliced at the production site and then further processed by specific cleaning methods for medicinal use. The Generic Name uses its traditional Chinese name, ``Jing-pian-jiang-huang". The assigned unique ID is ``NMM-0016". Another example of a Systematic Name is ``Ephedra sinica Stem-herbaceous Segmented and Aquafried-honey" for ``Mi Cao Ma Huang" (蜜草麻黄) \cite{nmm-000b} (\boldcolor{\cref{fig:snnmm}a}). It contains the species origin, medicinal part, and processing method, but no critical information for the special description. 

For the species origin, some NMMs come from more than one species. In this case, the Systematic Name will incorporate the Latin names of all pertinent species to ensure clarity. For example, the Systematic Name for ``Ma Huang" (麻黄) includes the species of \textit{Ephedra equisetina}, \textit{Ephedra intermedia}, and \textit{Ephedra sinica} (“Ephedra equisetina vel intermedia vel sinica Stem-herbaceous”) \cite{nmm-0006}. If, in the historical record, the species origin for an NMM is not clear, the term ``unspecified" will be included in the Systematic Name. For example, ``Pu Gong Ying" (蒲公英) in the \textit{Chinese Pharmacopoeia (2020 edition)} is described as an NMM using its whole plant. However, it could originate from any species within the \textit{Taraxacum} genus, which includes thousands of species \cite{Taraxacum}. It is unclear which single or multiple species within \textit{Taraxacum} have the medicinal effect and can be attributed to ``Pu Gong Ying". Therefore, we designate the Systematic Name for ``Pu Gong Ying" as ``Taraxacum unspecified Herb" \cite{nmm-01yf}. For the medicinal part, different parts of the same species can potentially be used as NMMs with varying medicinal properties. Therefore, it is necessary to specify the exact part used for medicine in the Systematic Name to ensure clarity. For example, both the herbaceous stem and the root of \textit{Ephedra sinica} can be used medicinally; hence, their Systematic Names are designated as ``Ephedra sinica Stem-herbaceous" \cite{nmm-0003} and ``Ephedra sinica Root" \cite{nmm-000g}, respectively.

The information for local production and processing methods for medicinal use is also critical for NMM differentiation and identification. NMMs often exist in three types (\boldcolor{\cref{fig:snnmm}b}): Raw NMMs, Agricultural NMMs, and Processed NMMs. Raw NMMs are unprocessed, representing NMMs in their natural state. Agricultural NMMs are derived from Raw NMMs through initial preparation at the production site and are not intended for direct medicinal use. Agricultural NMMs are often regulated as agricultural products. On the other hand, Processed NMMs are further processed using specific methods for medicinal use. These processing methods or procedures often change the molecular nature and composition of the NMMs, which are critical for medicinal efficacy. Processed NMMs are typically regulated as pharmaceuticals. In countries like China \cite{chp2020vol1}, Agricultural and Processed NMMs are subject to different regulatory frameworks due to their distinct applications in agriculture and pharmaceuticals. Historically, the information for the initial preparation and processing method is not a part of the identities of NMMs. Due to this, consumers often mistakenly purchase Agricultural NMMs instead of the necessary medicinal-grade Processed NMMs. This oversight also results in regulatory challenges for industry and government (\boldcolor{Supplementary text \labelcref{appendix:snnmm,appendix:snnmm_zh}}). 

To implement the Systematic Nomenclature for a large number of NMMs for the researcher and consumer communities, we developed an algorithm and its online web application, ShennongName, for interactive utility (\boldcolor{\cref{fig:snn}}, \boldcolor{Supplementary method \labelcref{appendix:snnmma}}, \url{https://shennongalpha.westlake.edu.cn/name}). Researchers and users can contribute new NMM entries or modify existing ones through ShennongName, which will be integrated into the AI platform after review (\boldcolor{\cref{fig:snn} \circled{8}}, also see below). The Systematic Nomenclature, accompanied by the ShennongName algorithm and application, provides a standardized, unambiguous, and efficient platform for differentiating and identifying NMMs.

\subsection{ShennongAlpha: an AI-driven sharing and collaboration platform for intelligent curation, acquisition, and translation of NMM knowledge}

Following Systematic Nomenclature, we embarked on a comprehensive process to collect and curate NMM knowledge. The initial phase of the endeavor described here involved gathering a total of 14,256 Chinese NMM entries (\boldcolor{\cref{stable:snkb}}), as they are the largest collection of NMMs. The resources include the 2020 and 2015 editions of the \textit{Chinese Pharmacopeia} \cite{chp2020vol1, chp2015vol1}, and the Chinese Medicinal Information Platform \cite{dayiorg}. Using this knowledge, we applied the Systematic Nomenclature and ShennongName to compute the relevant information and generated the Systematic Name, Generic Name, and ID for each NMM. Meanwhile, this collected knowledge was digitalized, structured, reviewed, and curated into an NMM Knowledge Base (\boldcolor{\cref{sfig:snkb}}). To share this valuable Knowledge Base with worldwide users and facilitate its utilization and collaboration, we developed an AI-powered web platform named ``ShennongAlpha" for the efficient and intelligent curation, acquisition, and translation of related NMM knowledge. 

ShennongAlpha, named after the pioneer of Chinese medicine, is an advanced AI platform characterized by its comprehensive integration. It combines ShennongName with a hexa-domain modular system to establish its foundational architecture. This architecture encompasses a custom-designed NMM Web, a Knowledge Base, a Search Engine, and a Large Language Model (LLM) system, along with an application for chat-based human-machine interaction and an application for standardized machine translation (\boldcolor{\cref{fig:sna-intro}b}).

We believe that the cornerstone of an excellent AI platform lies in its user-friendliness and accessibility. With this in mind, ShennongAlpha has been crafted to ensure that even users with no background in AI, programming, or algorithms can easily navigate and utilize the platform. In the following sections, we will illustrate how users can effortlessly and intelligently curate, acquire, and translate NMM-related knowledge through the straightforward process of accessing and harnessing the power of ShennongAlpha via the Web. The sophisticated technical details, implementations, and algorithms that underpin ShennongAlpha are discussed extensively in the ``Methods" and ``Supplementary information" sections, showcasing the substantial work and expertise that have gone into developing such an accessible yet advanced platform.

\subsection{Users access the curated NMM knowledge on ShennongAlpha}

To allow users to access and interact with the curated NMM knowledge easily, we launched the ShennongAlpha Web (\url{https://shennongalpha.westlake.edu.cn}). The Web serves as the primary portal for users to engage with ShennongAlpha, and it is distinguished by its modern and user-centric web design, allowing a consistently superior user experience regardless of device screen dimensions (\boldcolor{\cref{fig:sna-intro}c}). The Web was crafted to be bilingual, toggling seamlessly between English and Chinese, thereby catering to both Chinese and international users (\boldcolor{\cref{fig:sna-intro}d, e}). The bilingual feature of ShennongAlpha, for the first time, enables the international biomedical, pharmaceutical, physician, and patient communities to access the knowledge of Chinese NMMs, as this knowledge has largely been limited to the local community in the past. The standardized and structured Knowledge Base in ShennongAlpha is critical for correctly understanding and utilizing NMM knowledge. Therefore, ShennongAlpha is particularly significant for future research and application of NMMs. 

To access the NMM knowledge through the ShennongAlpha Web, users can easily browse different NMM entries through distinct knowledge pages. The URLs of these pages are structured by unique NMM IDs, as exemplified below:

\url{https://shennongalpha.westlake.edu.cn/knowledge/<nmm-id>}

For instance, ``Ma Huang" is assigned the NMM ID: ``NMM-0006", leading to the knowledge page:

\url{https://shennongalpha.westlake.edu.cn/knowledge/nmm-0006}

Besides accessing information via direct URLs, users can leverage a search function. This feature is prominently available within a search box on the homepage (\boldcolor{\cref{fig:sna} \circled{1}}) or at the top navigation bar of every other page (\boldcolor{\cref{fig:sna} \circled{2}}), all powered by the ShennongAlpha Search Engine. Upon executing a search, results are presented on a dedicated search page, where each entry contains pertinent details, including the NMM ID, Systematic Name, Generic Name, and brief introductory abstract (\boldcolor{\cref{fig:sna} \circled{4}}). This information aids users in discerning the relevance of their search results.

Knowledge pages within ShennongAlpha utilize a hierarchical tree structure, with content organized into sequenced sections and allowing for nested subsections (\boldcolor{\cref{fig:sna} \circled{5}, \circled{6}}). The ``Table of Contents" in the sidebar navigation facilitates users in swiftly pinpointing and navigating to distinct knowledge sections (\boldcolor{\cref{fig:sna} \circled{7}}). 

In our bid to ensure knowledge accessibility across language barriers, we designed a new text format known as Multilingual Markdown (MLMD) (\boldcolor{Supplementary method \labelcref{appendix:mlmd}}) for curating multilingual NMM knowledge. This approach allows ShennongAlpha to universally support four distinct language display modes: Chinese-English (zh-en), English-Chinese (en-zh), Chinese (zh), and English (en) (\boldcolor{\cref{sfig:modes}}). Users can effortlessly switch between these modes using the language button (\boldcolor{\cref{fig:sna}} \circled{8}). The functionalities of these modes are detailed below:

\begin{itemize}[nosep]
    \item zh-en: The Platform interface is in Chinese, with the Knowledge Base content rendered in a bilingual Chinese-English format.
    \item en-zh: The Platform interface is in English, with the Knowledge Base content rendered in a bilingual English-Chinese format.
    \item zh: Both the Platform interface and the Knowledge Base content are exclusively in Chinese.
    \item en: Both the Platform interface and the Knowledge Base content are exclusively in English.
\end{itemize}

Furthermore, within knowledge pages, users are afforded the option to save (\boldcolor{\cref{fig:sna} \circled{9}}), cite (\boldcolor{\cref{fig:sna} \circled{11}}), or download (\boldcolor{\cref{fig:sna} \circled{12}}) the content of the page. The downloadable content is structured in JSON format, enabling users to programmatically utilize the original NMM knowledge for advanced processing and applications.

\subsection{Users contribute NMM knowledge through ShennongAlpha}

Unlike traditional knowledge platforms, ShennongAlpha is uniquely designed as an open and collaborative hub. It invites global users to contribute their invaluable knowledge and findings about NMMs. On every knowledge page, users can find a prominent ``Leave your name and knowledge!" button (\boldcolor{\cref{fig:sna} \circled{13}}). This feature encourages users to submit new information or propose modifications to existing content. Each section of NMM knowledge, such as ``Abstract", ``Chinese Pharmacopoeia", and ``MLMD Encyclopedia" is accessible for enhancement or amendment through the ``Edit Content" button (\boldcolor{\cref{fig:sna} \circled{14}}). These user-generated modifications will be integrated into the Knowledge Base upon review and approval by ShennongAlpha. This feature will continually advance the Knowledge Base by leveraging and empowering the community.

Furthermore, the ``Show Edit History" button (\boldcolor{\cref{fig:sna} \circled{15}}) provides a tracking method and a transparent view of all historical modifications. Notably, once a user's contribution is included in the Knowledge Base, it will be acknowledged. Their names and avatars are prominently displayed under the ``Contributors" section (\boldcolor{\cref{fig:sna} \circled{16}}), not only as a token of appreciation but also as a badge of honor for their collaborative efforts. By pooling the collective expertise of the global NMM community, ShennongAlpha will rapidly evolve, driven by shared knowledge, collaborative spirit, and global community dedication.

\subsection{Chat-based NMM knowledge acquisition on ShennongAlpha}

In addition to the structured knowledge pages for a comprehensive overview, ShennongAlpha also heralds a major shift in NMM knowledge acquisition with the introduction of its AI-powered Chat application (ShennongChat, \boldcolor{\cref{fig:snc}}). By integrating the advanced LLM system into the Chat application, users are empowered with chat-based acquisition for NMM knowledge. Users can pose their questions to the Chat in natural language, eliminating the need for a technical background or specific search syntax. Leveraging the power of the Search Engine, the Chat automatically searches the Knowledge Base in real-time and articulates the gathered knowledge into a natural language response for the user (\boldcolor{\cref{sfig:snc-meth}}).

Users can access the Chat via the following URL on the ShennongAlpha Web:

\url{https://shennongalpha.westlake.edu.cn/chat}

The Chat supports multi-turn conversations. For example, a user poses the following question (\boldcolor{\cref{fig:snc} \circled{7}}):

\begin{stext}
What is the species origin of Ma Huang?
\end{stext}

Then, the Chat interprets the user's intent, automatically searches the Knowledge Base, and provides the following response (\boldcolor{\cref{fig:snc} \circled{8}}):

\begin{stext}
The species origin of Ma Huang is Ephedra equisetina, Ephedra intermedia, or Ephedra sinica.
\end{stext}

It's noteworthy that this answer aptly highlights the multi-species character of “Ma Huang”, distinguishing it from the response given by GPT-4 (\boldcolor{\cref{sfig:gpt}}).

Furthermore, the Chat's interface has been specially designed to enhance the interpretability of responses. Search results are embedded within the Chat’s answers in a collapsible format, allowing users to expand these sections for deeper insights into the background knowledge supporting the answers (\boldcolor{\cref{fig:snc} \circled{9}}).

Moreover, leveraging the advanced intelligence of the LLM system, the Chat supports interactions in multiple languages. Users can specify their preferred response style, whether in a different language or a particular structural format (\boldcolor{\cref{sfig:snc-style}}). This adaptability enhances the overall user experience, broadens knowledge accessibility, and promotes global knowledge sharing.

\subsection{Standardized and interpretable translation of NMM knowledge on ShennongAlpha}

Another important factor in NMM research and applications is language barriers, as much of the accumulated experience and knowledge is often recorded in a monolingual format. Therefore, a key feature of the ShennongAlpha platform is an AI-based standardized and interpretable Translate application for NMM knowledge (ShennongTranslate, \boldcolor{\cref{fig:snt}}). The Translate application is underpinned by an AI translation algorithm that we developed (Neural Machine Translation based on Coreference Primary Term, NMT-CPT, \boldcolor{\cref{sfig:snt-meth}}). It addresses language barriers and also prevents semantic drift caused by non-standardized translations. 

Users can access the Translate via the following URL on the ShennongAlpha Web:

\url{https://shennongalpha.westlake.edu.cn/translate}

The Translate upholds consistency in translations from Chinese to English or vice versa.  For example, a user submits the following text to be translated (\boldcolor{\cref{fig:snt} \circled{2}}):

\begin{stext}
麻黄是一种天然药材。
\end{stext}

Then, the Translate automatically searches the Knowledge Base for the related NMM entities in the text, retrieves their standardized translations, and generates the final translation for the sentence to ensure it is standardized and interpretable for both language and NMM knowledge requirements (\boldcolor{\cref{fig:snt} \circled{5}}):

\begin{stext}
Ephedra equisetina vel intermedia vel sinica Stem-herbaceous (NMM-0006, Ma-huang) is a kind of Natural Medicinal Material. 
\end{stext}

Here, ``麻黄" is translated using its standardized NMM nomenclature: ``Ephedra equisetina vel intermedia vel sinica Stem-herbaceous (NMM-0006, Ma-huang)". Similarly, “天然药材” incorporates a user-defined glossary as ``Natural Medicinal Material".

Uniquely, within the Translate's interface, distinct color highlights are applied to both the standardized translation ``Ephedra equisetina vel intermedia vel sinica Stem-herbaceous (NMM-ooo6, Ma-huang)" (\boldcolor{\cref{fig:snt} \circled{6}}) and the user-defined glossary ``Natural Medicinal Material" (\boldcolor{\cref{fig:snt} \circled{7}}). This offers users instant clarity on specialized terms, thus enhancing the standardization and interpretability of the translation results. Even more significantly, the Translate automatically identifies the NMM associated with the standardized translation, enabling access to its knowledge page on ShennongAlpha. By merely hovering over the standardized translation term, users are presented with a bilingual summary tooltip (adaptable based on the user's language display preference on ShennongAlpha) (\boldcolor{\cref{fig:snt} \circled{6}'}). Clicking the term within the tooltip directly navigates to its detailed knowledge page. This interactive approach not only provides users with standardized translations but also fosters further exploration of related knowledge, thereby enhancing understanding. It extends the traditional role of the Translate beyond mere translation, transforming it into an application for knowledge exploration, learning, and sharing.

\section{Discussion}

Global knowledge sharing is particularly important for specialized fields such as NMMs, which are traditionally rich with historical records documented in local texts and formats. The emergence of large language models (LLMs) offers unique opportunities to facilitate knowledge sharing. However, it also raises concerns that incorrect information could be spread by advanced AI technologies at unprecedented rates. Furthermore, LLMs trained with misinformation could undermine knowledge integrity in specialized fields. Therefore, it is urgent that knowledge in such fields be correctly curated and translated, thereby enabling intelligent acquisition and further dissemination through AI methods.  

A critical issue for sharing NMM knowledge and promoting research and applications is the lack of standardized and systematic nomenclature. This deficiency hampers accurate differentiation and identification of NMMs, and obstructs effective communication and collaboration within the field. In comparison, the pharmaceutical sector has established systematic nomenclature for chemical drugs to ensure accuracy and consistency in chemical identification and research. Specifically, the chemical systematic naming system includes a systematic name for accurately describing the chemical structure of each drug (the International Union of Pure and Applied Chemistry nomenclature \cite{leigh2011principles}); a generic name for convenient usage (the International Nonproprietary Names \cite{serafini2021s}); and a unique identification number for digital platforms (the Chemical Abstracts Service registry number \cite{morgan1965generation}, the PubChem Compound ID number \cite{kim2023pubchem}).

In this study, we have introduced a Systematic Nomenclature for NMMs. The Systematic Name describes an NMM by its species origin, medicinal part, special description, and processing method. The Generic Name provides convenience for NMM usage and also observes local tradition. The NMM ID facilitates use on digital platforms. Therefore, for the first time, the Systematic Nomenclature provides a standardized system that uniquely differentiates and identifies each NMM. Using the Systematic Nomenclature and its companion ShengnonAlpha platform (see below), we have assigned identities for more than 14,000 Chinese NMMs and curated their related knowledge in the Knowledge Base. The Systematic Nomenclature could also be applied to additional NMMs from other regions of the world.

To systematically empower NMM knowledge sharing, we have developed the AI-driven platform ShengnonAlpha. First, ShennongAlpha automates the implementation of the Systematic Nomenclature for NMMs through the tailored algorithm and application. Second, to address language barriers, all knowledge entries in ShennongAlpha are curated bilingually in English and Chinese, making them accessible to both Chinese and global users. This is supported by our Translate application, which provides standardized and interpretable translation for the cross-lingual sharing of NMM knowledge. Although the current platform primarily supports English and/or Chinese interface displays and bidirectional text translation, the MLMD syntax and the NMT-CPT algorithm allow the platform to be easily expanded for multilingual presentation and translation into other languages. Third, by leveraging the LLM, ShennongAlpha also offers the Chat application for NMM knowledge acquisition. Its conversational interface enables users to obtain standardized NMM knowledge in an interactive, chat-based manner. Fourth, ShennongAlpha has introduced a unique collaboration mechanism for updating NMM knowledge. By encouraging global users to contribute their valuable new information and findings about NMMs, the Knowledge Base within ShennongAlpha will continue to grow. This represents not only a mechanism for sharing new knowledge but also fosters collaboration within the global NMM community. Finally, despite its sophisticated architecture, algorithms, and implementation, ShennongAlpha offers an integrated, user-friendly interface through the ShennongAlpha Web. This design skillfully conceals the underlying complexities, making it easily accessible for users.

In summary, ShennongAlpha epitomizes the exciting fusion of AI with NMM knowledge sharing in the novel AI for Science field. ShennongAlpha’s hexa-domain modular architecture (\boldcolor{\cref{fig:sna-intro}b}) also offers a model for knowledge sharing in other specialized fields. ShennongAlpha also provides the world’s largest NMM Knowledge Base with standardized and curated knowledge, serving as a unique resource for NMM researchers, physicians, patients, and users, as well as for future LLM training. We believe that as AI increasingly emerges as a primary knowledge provider in scientific research \cite{yang2023ai}, our platform will significantly promote global knowledge sharing and collaboration in the NMM field, ultimately benefiting global health and human well-being.

\section{Methods}

\subsection{Systematic Nomenclature for NMMs with its algorithm and application}

Detailed rules for Systematic Nomenclature for NMMs (SNNMM) can be found in the \boldcolor{Supplementary text \labelcref{appendix:snnmm,appendix:snnmm_zh}}.

Systematic Nomenclature for NMMs Algorithm (SNNMMA) is coded in Python, with its algorithm stringently following the rules of SNNMM. It is open-source, and the source code is available on GitHub: \url{https://github.com/shennong-program/shennongname}. 

The associated Python package for SNNMMA is published and released on PyPI: \url{https://pypi.org/project/shennongname}.

For a concise overview of the data structure and algorithms of SNNMMA, refer to the \boldcolor{Supplementary method \labelcref{appendix:snnmma}}.

\subsection{Multilingual Markdown}

Multilingual Markdown (MLMD) is a newly designed lightweight markup language explicitly tailored for managing multilingual text. 

The essential syntax of MLMD can be found in the \boldcolor{Supplementary method \labelcref{appendix:mlmd}}.

The MLMD HTML parser is built using TypeScript. It is open-source, with the source code accessible on GitHub: \url{https://github.com/shennong-program/mlmd}. 

The associated TypeScript package is published and released on npm: \url{https://www.npmjs.com/package/mlmd}.

\subsection{ShennongAlpha Knowledge Base}

ShennongAlpha Knowledge Base (ShennongKB) represents the core knowledge base within the ShennongAlpha and resides in the 5th layer known as the knowledge base layer (\boldcolor{\cref{fig:sna-intro}b}). 

ShennongKB, built on the MongoDB document-oriented database, integrates and organizes a diverse data set. The current version of ShennongKB predominantly sources its data from the 2020 \cite{chp2020vol1} and 2015 \cite{chp2015vol1} editions of the \textit{Chinese Pharmacopoeia}, and the Chinese Medicinal Information Platform \cite{dayiorg}. This NMM-associated data is organized into several interrelated collections, including SNNMM, Text, Knowledge, Glossary, and Relation (\boldcolor{\cref{sfig:snkb}}). This organization follows a thorough process of collection, structuring, standardization, and review, all carried out by our ShennongAlpha team.

\subsection{ShennongAlpha Web}

The ShennongAlpha Web serves as the principal interface for users with the ShennongAlpha, situated within ShennongAlpha's 1st layer, the web and user interaction layer (\boldcolor{\cref{fig:sna-intro}b}). 

Embracing a responsive design approach, ShennongAlpha Web assures users of a uniform and superior user experience on screens of any dimension. ShennongAlpha Web is built on TypeScript, employing Next.js as its principal framework. It amalgamates the ShennongAlpha Knowledge Base and Search Engine. Additionally, it incorporates applications such as ShennongName, ShennongChat, and ShennongTranslate. ShennongAlpha takes advantage of a microservices architecture, facilitating the decoupling of diverse service units for enhanced manageability and extensibility. The backend for applications like ShennongName, ShennongChat, and ShennongTranslate is developed using Python, leveraging the Flask or FastAPI framework. These applications interface with ShennongAlpha through APIs that adhere to RESTful standards, fostering overall system stability and maintainability. For deployment, ShennongAlpha exploits Docker for container orchestration and Docker Compose technology for integrating and managing multiple containerized applications.

\subsection{ShennongAlpha Search Engine}

ShennongAlpha Search Engine (ShennongSearch) \boldcolor{(\cref{sfig:sns})} is the core of knowledge retrieval in the ShennongAlpha, located in the 4th layer referred to as the search engine layer (\boldcolor{\cref{fig:sna-intro}b}). 

ShennongSearch can be seen as a data highway within the ShennongAlpha. To accommodate a wide array of queries and guarantee efficient and precise information retrieval, ShennongSearch is customized and designed with three advanced search methods: Coreference-based Graph Search (CGS) (\boldcolor{\cref{sfig:sns}a}), vector search (\boldcolor{\cref{sfig:sns}b}) and full-text search (\boldcolor{\cref{sfig:sns}c}). Each search method possesses distinct strengths and can be employed individually or collectively to yield accurate and thorough search results, thereby enhancing the overall usability of ShennongSearch. The three advanced search methods of ShennongSearch are primarily applied in the knowledge search process of ShennongChat. In the publicly accessible ShennongChat, knowledge searches apply the CGS method.

\textbf{Coreference-based Graph Search.} The CGS algorithm (detailed in \boldcolor{Supplementary method \labelcref{appendix:cgs}}), designed and developed by this project, centers on dispatching relationships between different terms within the relation collection of ShennongKB, automatically constructing a Coreference Primary Term Graph (CPTG) (\boldcolor{\cref{sfig:sns}a}). CPTG essentially operates as a directed acyclic graph. Given a term/named entity, graph search within CPTG locates its corresponding Primary Term. This Primary Term then allows querying its corresponding standard information within other collections in ShennongKB (such as Knowledge or Translation). This method is especially useful when standardizing queries. We construct the CPTG using the Python networkx package \cite{NetworkX}. The associated Python implementation for CGS is available on GitHub (\url{https://github.com/shennong-program/pycgs}) or PyPI (\url{https://pypi.org/project/pycgs}).

\textbf{Vector search.} The method utilizes a vector-based approach, transforming search queries and documents within ShennongKB into vector embeddings (\boldcolor{\cref{sfig:sns}b}). A unique strength of this method lies in its ability to accept entire sentences or paragraphs as search input. By converting this text into vector space through embedding, the semantic essence is captured within a vector representation. Subsequently, by evaluating the similarity between the query text vector embedding and the archived text vector embeddings within ShennongKB, information that is semantically relevant to the query text can be swiftly identified and retrieved. We employ OpenAI's embedding model for text vectorization: \lstinline{text-embedding-3-small}. Each text within ShennongKB is preprocessed with \lstinline{text-embedding-3-small}, generating a 1536-dimensional vector embedding. These embeddings and their original texts are then stored within ShennongKB. Upon each new search, the query text is processed with \lstinline{text-embedding-3-small} to generate a new vector embedding. Then, the search engine uses cosine similarity to measure the similarity between this new embedding and the vector embeddings archived in ShennongKB, returning the text corresponding to the most similar vector embedding as the search result. ShennongSearch utilizes MongoDB's \lstinline!$vectorSearch! operator for its vector search functionality. 

\textbf{Full-text search.} This method is rooted in full-text fuzzy searching, using a tokenization and inverted index process (\boldcolor{\cref{sfig:sns}c}). Operating by breaking down documents and search queries in ShennongKB into individual words or ``tokens", this method allows efficient searching and retrieval of pertinent information, accommodating approximate matches, variations in phrasing, typos, misspellings, or alternate spellings. This method notably enhances the system's flexibility and user-friendliness by leveraging the tokenization and inverted index. ShennongSearch utilizes MongoDB's \lstinline!$text! and \lstinline!$search! operators for its full-text search functionality. Documents within MongoDB are tokenized using Jieba \cite{Jieba}, after which a full-text inverted index is constructed. During queries, the text to be searched is also tokenized using Jieba before the full-text search is executed.

\subsection{ShennongAlpha Large Language Model}

Situated within the 3rd layer, the Artificial Intelligence layer, ShennongAlpha Large Language Model (ShennongLLM) acts as the pivotal AI hub for the ShennongAlpha (\boldcolor{\cref{fig:sna-intro}b}). Built on a foundation of LLMs, the hub is designed for adaptability, supporting both open-source and proprietary language models like Llama 2/3 \cite{touvron2023llama}, GPT-3.5/4/4o \cite{openai2023gpt4}, Gemini, Claude, ERNIE, Qwen, among others. This versatility is accessible either through local deployment or API calls.

\subsection{ShennongAlaph Chat application}

Acquiring knowledge through direct chat with LLMs heralds a transformative paradigm in information retrieval. However, LLMs often produce inaccurate or spurious outputs, especially in unfamiliar domains \cite{Survey_Hallu}. Given the fact that NMM knowledge had not been standardized prior to this study, the standardized knowledge we curated was unavailable for training LLMs. Therefore, even state-of-the-art LLMs like GPT-4 tend to provide misleading responses (\boldcolor{\cref{sfig:gpt}}). Recognizing this constraint, we adopt the retrieval augmented generation (RAG) methodology \cite{peng2023check, nakano2021webgpt} in the design of the ShennongAlaph Chat application (ShennongChat), substantially enhancing the accuracy of LLM responses.

ShennongChat utilizes the prompt chain technique \cite{LangChain, wei2022chain}, a method that uses a step-by-step process to ensure the LLM's responses to user queries are based on retrieved standardized NMM knowledge. This technique enhances the relevance and reliability of the knowledge acquired during chats.

For illustrative purposes, let's examine the user query mentioned in the ``Results" section (\boldcolor{\cref{fig:snc,sfig:snc-meth}}):

\begin{stext}
What is the species origin of Ma Huang?
\end{stext}

Upon receiving a query, ShennongChat processes it using ShennongLLM to assess if the current chat history contains sufficient knowledge to respond. If it does, ShennongLLM will directly answer the user’s question. In cases where more information is needed, ShennongSearch is activated to conduct a knowledge search. This process involves dispatching ShennongSearch to retrieve relevant standardized knowledge from ShennongKB. From this search, we glean:

\begin{stext}
species_origins: Ephedra equisetina, or, Ephedra intermedia, or, Ephedra sinica
\end{stext}

This contextual information is then amalgamated with the user's query and submitted to ShennongLLM to formulate an accurate response:

\begin{stext}
The species origin of Ma Huang is Ephedra equisetina, Ephedra intermedia, or Ephedra sinica.
\end{stext}

\subsection{ShennongAlaph Translate application}

To enhance the quality of standardized translations, particularly in the field of NMMs, capturing the original text's essence and employing standardized terminology in the target language is imperative. This way, we can ensure a coherent linguistic and terminological framework, facilitating standardized scholarly communication. However, despite the considerable advancements in Neural Machine Translation (NMT) technologies \cite{Transformer,li2022prompt,GPTonMT}, current NMT algorithms fall short of fulfilling our specialized requirements. To tackle this limitation, we introduced a new translation algorithm named ``Neural Machine Translation Based on Coreference Primary Term (NMT-CPT)". 

At the heart of the NMT-CPT is a dual-purpose strategy. When a user submits a translation request via ShennongTranslate, the system proactively identifies standardized terms within the text recorded in ShennongKB, referencing their Primary Terms and standardized translations in the target language. Subsequently, using the specialized annotation syntax of MLMD, ShennongTranslate, powered by ShennongLLM, produces a translation that incorporates these standardized translations.

To illustrate, let's examine the translation request mentioned in the ``Results" section (\boldcolor{\cref{fig:snt,sfig:snt-meth}}):

\begin{stext}
Text for translation:
麻黄是一种天然药材。
Translation direction:
zh -> en
User-customized dictionary:
天然药材 -> Natural Medicinal Material
\end{stext}

Once ShennongTranslate receives the translation request, it tokenizes the input text and utilizes ShennongSearch's CGS to identify potential NMMs mentioned within the text and search out the corresponding Primary Terms. In this case, the term ``麻黄" is detected, associated with the corresponding Primary Term ``nmm-0006". Based on the Primary Term (``nmm-0006") and the target language code (``en"), we can probe the ShennongKB's glossary collection, ultimately obtaining its standardized translation in the ``NMMSN (NMM ID, NMMGN)" format:

\begin{stext}
Ephedra equisetina vel intermedia vel sinica Stem-herbaceous (NMM-ooo6, Ma-huang)
\end{stext}

These standardized and user-customized translations are then formatted according to the unique MLMD syntax used in the NMT-CPT algorithm: ``[[xxx $|$ yyy]]". After processing, the dictionary is formatted as follows:

\begin{stext}
麻黄 -> [[nmm-0006 | Ephedra equisetina vel intermedia vel sinica Stem-herbaceous (NMM-ooo6, Ma-huang)]]
天然药材 -> [[Natural Medicinal Material]]
\end{stext}

In the MLMD syntax, the term ``麻黄" is translated as ``[[nmm-0006 $|$ Ephedra equisetina vel intermedia vel sinica Stem-herbaceous (NMM-ooo6, Ma-huang)]]". Here, ``nmm-0006" (i.e., the ``xxx" part of the annotation) signifies the Primary Term (or NMM ID) for the term ``麻黄", as determined through a ShennongSearch's CGS query within ShennongKB. Meanwhile, ``Ephedra equisetina vel intermedia vel sinica Stem-herbaceous (NMM-ooo6, Ma-huang)" (i.e., the ``yyy" part) represents its standardized translation in the target language. For translations in the user-customized dictionary, as they do not need to apply a ShennongSearch query to find the Primary Terms, the annotation adopts the form ``[[yyy]]" without the ``xxx" component.

By forwarding the user's text to be translated, the translation direction, and the formatted dictionary to ShennongLLM, we obtain the following translation with MLMD syntax:

\begin{stext}
[[nmm-0006 | Ephedra equisetina vel intermedia vel sinica Stem-herbaceous (NMM-ooo6, Ma-huang)]] is a kind of [[Natural Medicinal Material]].
\end{stext}

ShennongTranslate further parses this MLMD-styled translation to provide functionalities for term highlighting and tooltips.

ShennongTranslate utilizes the jieba package \cite{Jieba} for tokenization and leverages the LangChain framework \cite{LangChain} to construct translation prompt templates. This approach guarantees that translations adhere to a strict structure and consistent formatting. By employing a few-shot learning technique \cite{brown2020language}, the system can generalize to the new translation syntax without requiring extensive model fine-tuning. 

\section{Data availability}

Data from ShennongKB, as well as the web and applications of the ShennongAlpha, can be accessed through the ShennongAlpha Web (\url{https://shennongalpha.westlake.edu.cn}).

\section{Code availability}

The SNNMMA and ShennongName suite can be obtained on GitHub (\url{https://github.com/shennong-program/shennongname}). The relevant Python package has been published and released on PyPI (\url{https://pypi.org/project/shennongname}).

The MLMD parser can be obtained on GitHub (\url{https://github.com/shennong-program/mlmd}). The relevant TypeScript package has been published and released on npm (\url{https://www.npmjs.com/package/mlmd}).

The Python implementation of CGS can be obtained on GitHub (\url{https://github.com/shennong-program/pycgs}) or PyPI (\url{https://pypi.org/project/pycgs}).

\section{Contributions}

YANG Zijie, ZHANG Yue, and XU Tian conceived and designed the study. YANG Zijie, TAO Wufan, ZHANG Yue, and XU Tian designed the SNNMM. YANG Zijie, YIN Yongjing, KONG Chaojun, and CHI Tiange collected, structured, and reviewed information about NMMs. YANG Zijie designed and completed the architecture, syntax, and algorithms related to ShennongAlpha. YANG Zijie and YIN Yongjing completed the development and deployment of ShennongAlpha. TAO Wufan, ZHANG Yue, and XU Tian supervised the study. YANG Zijie, YIN Yongjing, TAO Wufan, ZHANG Yue, and XU Tian wrote the paper.

\section{Acknowledgements}

We appreciate XU Anlong (Beijing University of Chinese Medicine), CHEN Kaixian (Chinese Academy of Sciences) for advising on this study; YANG Dong, CONG Peikuan for participating in the discussions of this study; YU Ziyan for assisting in deploying the GPT-3.5/4 API; WU Kaile, CHEN Jienan, ZHOU Zhaozhao for assisting in the development of the ShennongAlpha Web; YU Qian, WANG Weicheng, ZHANG Hongji for assisting in the artistic design of the ShennongAlpha Web; WU Qiuping, YANG Xusheng, ZHANG Lin, WU Qiaozhi, XU Shuangyan for participating in review of NMM knowledge.

This work was supported by a grant from the National Natural Science Foundation of China (U21A20201); grants from the Department of Science and Technology of Zhejiang Province (2020E10027, 2021ZY1019, 2022ZY1005); the Research Program No. 202208011 of Westlake Laboratory of Life Sciences and Biomedicine; the Research Program No. WU2023C020 of Research Center for Industries of the Future, Westlake University; the Westlake Education Foundation.

\section{Competing interests}
YANG Zijie, ZHANG Yue, and XU Tian hold the relevant patents (2023117101437, 2023117101511, 2023117101526) for the ShennongAlpha. The other authors declare no competing interests.

\printbibliography[heading=bibnumbered]
\end{multicols}

%% file: chapter/fig.tex
\begin{figure}
\centering
\includegraphics[width=\textwidth]{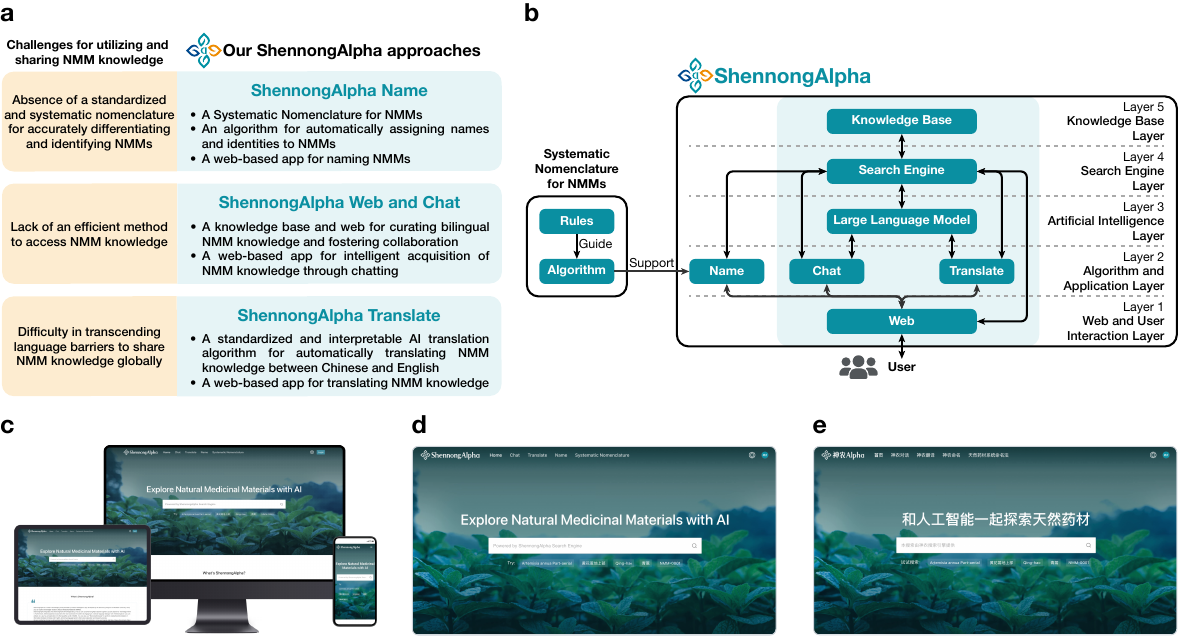}
\caption{
\textbf{ShennongAlpha: an AI-driven sharing and collaboration platform for intelligent curation, acquisition, and translation of NMM knowledge.}\\
\textbf{a.} The challenges for utilizing and sharing NMM knowledge and our ShennongAlpha approaches.
\textbf{b.} Architecture of the ShennongAlpha. ShennongAlpha applies the Systematic Nomenclature for NMMs (\boldcolor{\cref{fig:snnmm}}) and integrates ShennongName (\boldcolor{\cref{fig:snn}}) with a hexa-domain modular system to form its structure. The hexa-domain system is outlined in the light teal block. 
ShennongAlpha is structured into five layers, from shallow to deep: Layer 1: Web and user interaction layer. In this layer, users can access the system via our ShennongAlpha Web (\boldcolor{\cref{fig:sna}}). Layer 2: Algorithm and application layer. In this layer, we have specifically developed three applications customized for NMMs: ShennongName, ShennongChat (\boldcolor{\cref{fig:snc}}), and ShennongTranslate (\boldcolor{\cref{fig:snt}}). Users can access these applications on the corresponding pages of the ShennongAlpha Web. Layer 3: Artificial intelligence layer. In this layer, we have integrated the ShennongAlpha Large Language Model system, allowing the ShennongAlpha to process and respond to data from different layers intelligently. Layer 4: Search engine layer. In this layer, we have integrated the ShennongAlpha Search Engine customized for NMM-related data. Layer 5: Knowledge base layer. In this layer, we have integrated the ShennongAlpha Knowledge Base to curate NMM knowledge efficiently. 
Arrows represent the allowed data interactions between different layers. 
\textbf{c.} Cross-platform and user-friendly design of the ShennongAlpha.
\textbf{d.} The English homepage of the ShennongAlpha Web. 
\textbf{e.} The Chinese homepage of the ShennongAlpha Web. 
}
\label{fig:sna-intro}
\end{figure}

\begin{figure}
\centering
\includegraphics[width=\textwidth]{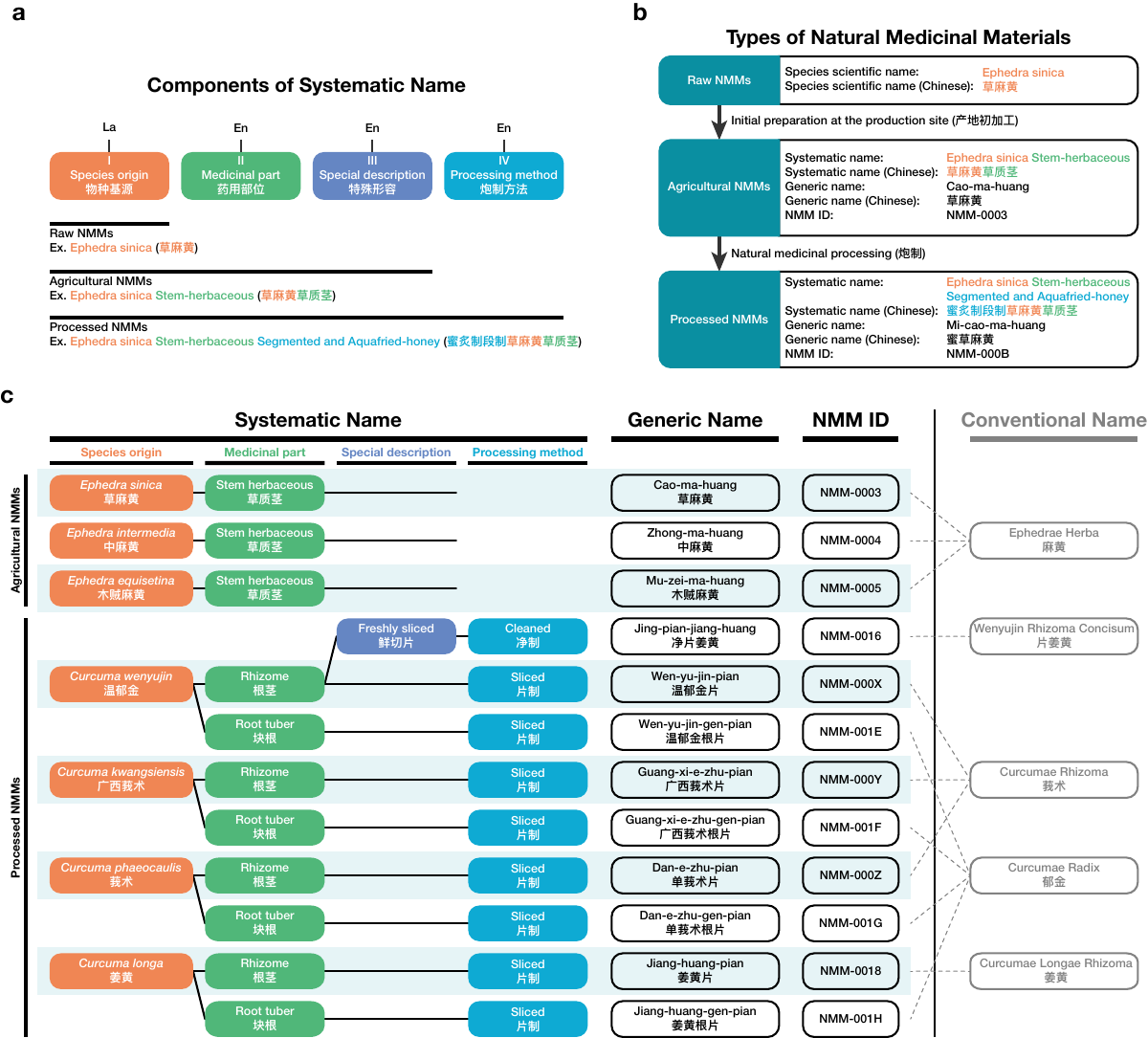}
\caption{
\textbf{Systematic Nomenclature for Natural Medicinal Materials.} \\
The Systematic Nomenclature assigns each NMM a unique Systematic Name, Generic Name, and NMM ID. 
\textbf{a.} Components of the Systematic Name. It consists of four components: I. Species origin, including species names in Latin; II. Medicinal part; III. Special description for initial preparations or specific characteristics; and IV. Processing method.
\textbf{b.} NMM Types. Raw NMMs are initially prepared at the production sites to produce Agricultural NMMs; Agricultural NMMs are often further processed to produce Processed NMMs. 
\textbf{c.} Examples of traditional Chinese NMMs in Systematic Nomenclature. Conventional names often lead to confusion by collectively referring to multiple NMMs that are not identical, due to missing or incorrect information about species origin, medicinal part, special description, and processing method. For example, the illustration shows three Agricultural NMMs from the \textit{Ephedra} genus with the herbaceous stem as the medicinal part, conventionally named ``Ephedrae Herba" (``麻黄"), leading to ambiguity. Similarly, nine Processed NMMs from the \textit{Curcuma} genus, with different medicinal parts, initial preparations and processing methods, are collectively referred to by four names: ``Wenyujin Rhizoma Concisum" (``片姜黄"), ``Curcumae Rhizoma" (``莪术"), ``Curcumae Radix" (``郁金"), and ``Curcumae Longae Rhizoma" (``姜黄"). In contrast, our Systematic Nomenclature accurately assigns distinct Systematic Names, Generic Names, and NMM IDs to these twelve different Agricultural and Processed NMMs, eliminating ambiguity. The dashed lines connect the conventional names to the different NMMs they collectively represent. 
}
\label{fig:snnmm}
\end{figure}

\begin{figure}
\centering
\includegraphics[width=\textwidth]{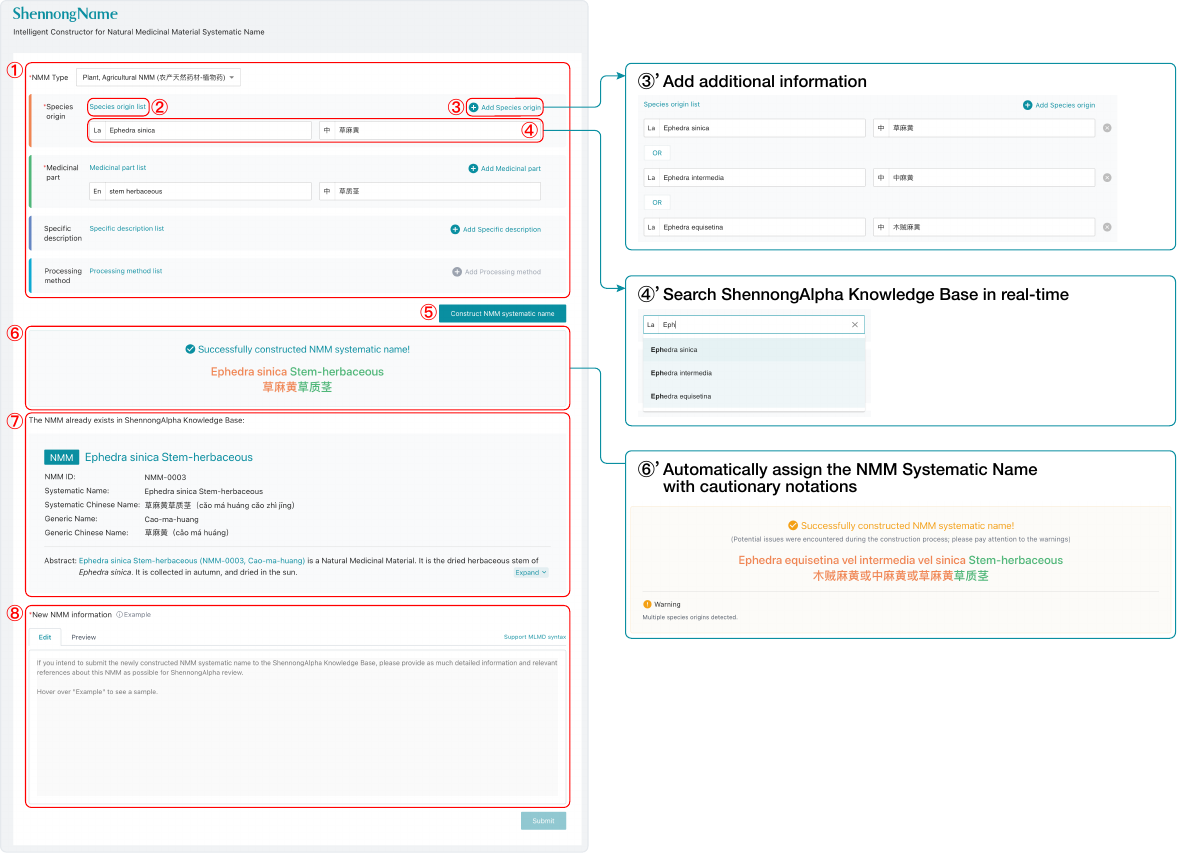}
\caption{
\textbf{Using ShennongName to automatically construct NMM Systematic Names.}\\
Users select the NMM type and provide information for the four name components in area \circled{1}. By clicking on hyperlinks like \circled{2}, users can view entries for the four name components already cataloged in the ShennongAlpha Knowledge Base. For each name component, users can add additional information by clicking on plus buttons like \circled{3}\ (\circled{3}'). When users begin entering name component information in text boxes like \circled{4}, ShennongName performs a real-time search in the Knowledge Base for relevant matching entries to enable auto-completion (\circled{4}'). After users have entered the necessary naming information for the NMM, they can click on the ``Construct NMM Systematic Name" button (\circled{5}), allowing ShennongName to automatically construct the Systematic Name using the algorithm. If the construction is successful, the generated information is displayed with a green background (\circled{6}). If any issues arise during construction, the relevant information is displayed with an orange background (\circled{6}'). For successfully constructed Systematic Names, ShennongName will also automatically perform a search for it in the Knowledge Base; if a matching NMM is found, users will be informed that the NMM is already recorded in the Knowledge Base, eliminating the need for redundant construction (\circled{7}). After successfully constructing a new Systematic Name, if users wish to add it to the Knowledge Base, they can provide relevant details about the NMM in the textbox in area \circled{8}\ and submit it. Once reviewed and approved by ShennongAlpha, the entry will be incorporated into the Knowledge Base.
}
\label{fig:snn}
\end{figure}

\begin{figure}
\centering
\includegraphics[width=0.95\textwidth]{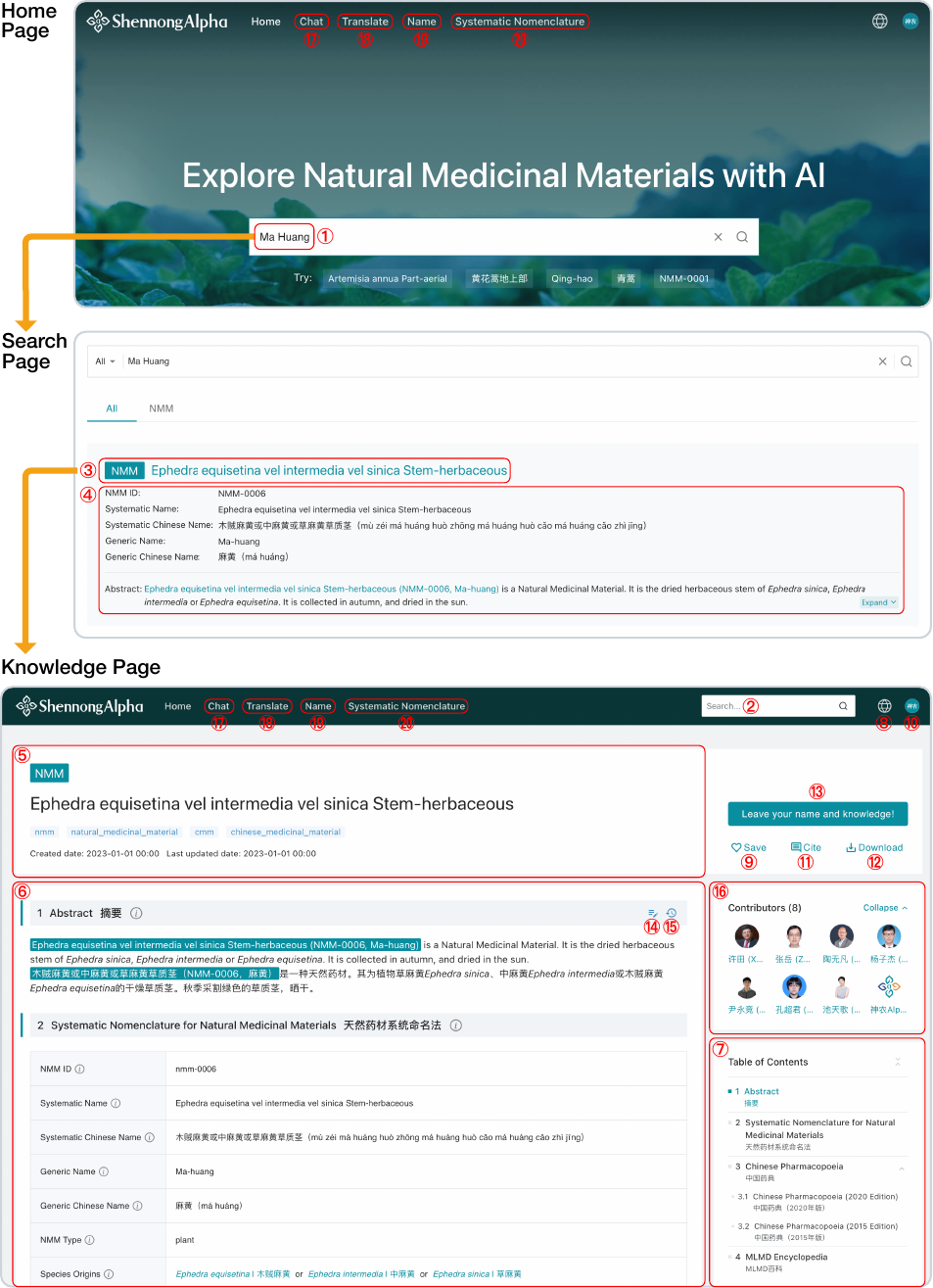}
\caption{
\textbf{Browsing NMM knowledge on the ShennongAlpha Web.} (Legend continued on next page.)
}
\label{fig:sna}
\end{figure}

\begin{figure}
\centering
\ContinuedFloat
\caption{
(Legend continued from the previous page.)
Users can initiate their exploration of NMM knowledge by using the search bar located either on the homepage (\circled{1}) or atop other pages (\circled{2}). Post-search, users are directed to the search page where they can glance through the title (\circled{3}) and summarized information (\circled{4}) of each NMM entry to ascertain its relevance. By clicking on the title of an entry, users are navigated to a detailed knowledge page dedicated to that specific NMM. The header of this knowledge page (\circled{5}) displays the Systematic Name of the NMM, while the main content area (\circled{6}) is organized in a structured, section-by-section layout. The ``Table of Contents" sidebar (\circled{7}) enables swift navigation between sections. To facilitate cross-language accessibility for global users, the Web offers four display modes (\circled{8}): Bilingual (Chinese-English), Bilingual (English-Chinese), Chinese only, and English only. The ``Save" button (\circled{9}) allows users to bookmark the knowledge page to their user dashboard (\circled{10}). To encourage academic references, the ``Cite" button (\circled{11}) offers citation formats in styles such as APA, MLA, GB/T 7714-2015, and BibTeX. The ``Download" button (\circled{12}) enables users to download the knowledge page's content in JSON format. Furthermore, with the ``Leave your name and knowledge!" button (\circled{13}), users can propose new or revised NMM-related knowledge. Contributions can also be made directly via the ``Edit Content" button (\circled{14}), allowing users to modify the content of each section. To review past modifications, the ``Edit History" button (\circled{15}) provides access to all historical changes. Approved user contributions are then integrated into the ShennongAlpha Knowledge Base, and contributors are recognized and acknowledged in the ``Contributors" area (\circled{16}), where their usernames and avatars are displayed. Users can navigate to the ShennongChat (\circled{17}, \boldcolor{\cref{fig:snc}}), ShennongTranslate (\circled{18}, \boldcolor{\cref{fig:snt}}), and ShennongName (\circled{19}, \boldcolor{\cref{fig:snn}}) applications in ShennongAlpha, as well as the detailed rules of the Systematic Nomenclature for NMMs (\circled{20}), directly from the homepage or through the header navigation bar on any page of the Web.
}
\end{figure}

\begin{figure}
\centering
\includegraphics[width=\textwidth]{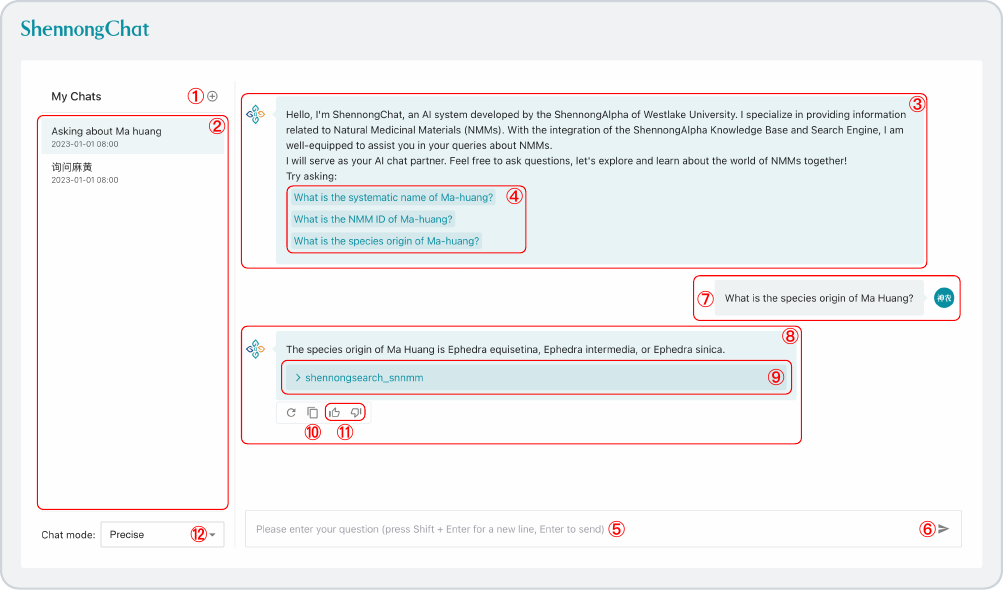}
\caption{
\textbf{Chat-based NMM knowledge acquisition through ShennongChat.}\\
Users can initiate a new chat by clicking the plus button (\circled{1}). The chat history area (\circled{2}) retains previous chat sessions, allowing users to navigate between historical chat sessions or delete unwanted chat sessions if needed. Within each chat session, the guidance section (\circled{3}) highlights the unique features and capabilities of ShennongChat, ensuring users can easily get started. To quickly experience the chatting features, users can select from a set of sample questions in the \circled{4}\ area. By inputting a question in the chat box (\circled{5}) and hitting the send button (\circled{6}), the user's question (\circled{7}) is relayed to ShennongChat for a response. In formulating its response, ShennongChat integrates the ShennongAlpha Search Engine to retrieve standardized knowledge about the NMM mentioned in the user's question from the ShennongAlpha Knowledge Base, providing a retrieval-augmented response (\circled{8}). Users can expand the search dropdown (\circled{9}) for details about the search results. The copy button (\circled{10}) enables users to duplicate the text of ShennongChat's response. Feedback on the responses (either positive or negative) can be provided by the user (\circled{11}), which helps in improving the quality of ShennongChat's responses. Depending on their preference, users can toggle between ShennongChat's chat modes: precise or quick (\circled{12}).
}
\label{fig:snc}
\end{figure}

\begin{figure}
\centering
\includegraphics[width=\textwidth]{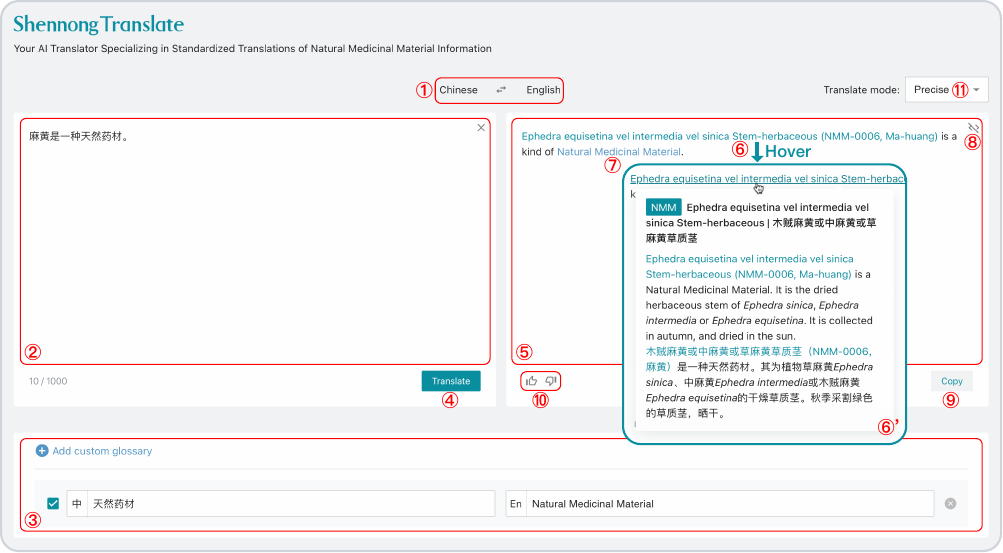}
\caption{
\textbf{Standardized and interpretable translation of NMM text through ShennongTranslate.}\\
Users can toggle between Chinese and English translation directions using \circled{1}. After inputting the NMM text to be translated in the text entry area (\circled{2}) and adding any user-customized glossary (\circled{3}, optional), users can click the ``Translate" button (\circled{4}) to submit the text for translation. The user-customized glossaries are saved, and users can add/delete, activate/deactivate these glossaries as needed, allowing ShennongTranslate to become increasingly tailored to their translation preferences. During the translation process, ShennongTranslate automatically uses the ShennongAlpha Search Engine to identify NMM terms in the text and then retrieve their Primary Terms and standardized translations from the ShennongAlpha Knowledge Base. In the translation results (\circled{5}). standardized translations of NMM terms (\circled{6}) and user-customized terms (\circled{7}) are highlighted in teal and blue, respectively. Hovering over a standardized translation of an NMM term (\circled{6}) prompts ShennongTranslate to display a tooltip (\circled{6}') containing its introductory abstract; clicking on the tooltip directs users to the knowledge page of the NMM. The button \circled{8}\ allows users to toggle between plain text and MLMD source code displays. In either display mode, users can copy the content by clicking the ``Copy" button (\circled{9}). Feedback on the translation (either positive or negative) can be provided by the user (\circled{10}), helping to improve ShennongTranslate's performance. Depending on their preference, users can toggle between ShennongTranslate's translate modes: precise or quick (\circled{11}).
}
\label{fig:snt}
\end{figure}

%% file: chapter/sfig.tex
\section{Supplementary figures}

\begin{figure*}[!ht]
\centering
\includegraphics[width=\textwidth]{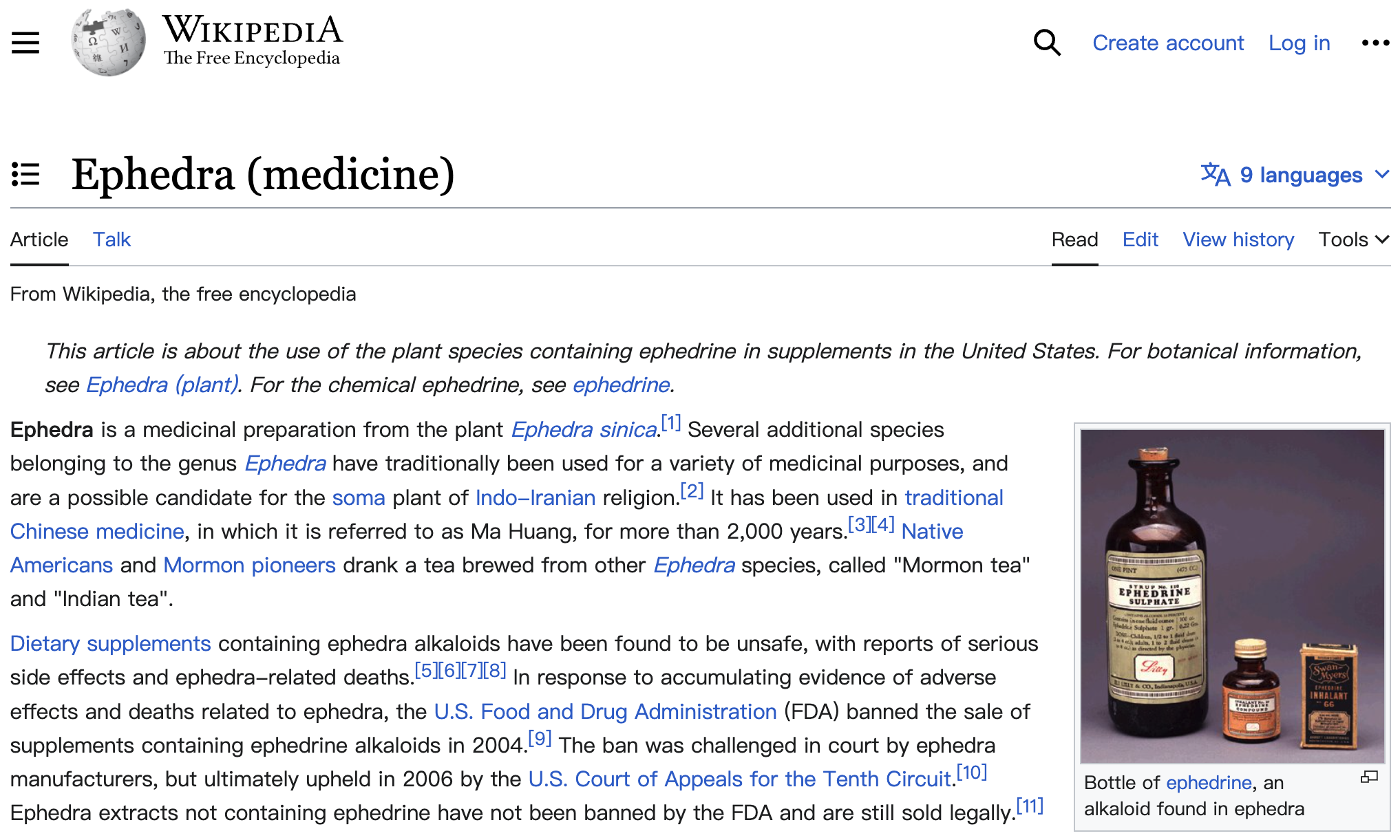}
\caption{
\textbf{Wikipedia entry: Ephedra.} \\
Wikipedia states that the species origin of Ephedra is \textit{Ephedra sinica} and claims that it is known as ``Ma Huang" in traditional Chinese medicine. However, this information is inaccurate. According to China's most authoritative pharmaceutical reference, \textit{Chinese Pharmacopoeia (2020 Edition)}, Ephedra is a multi-species NMM. Its species origins include \textit{Ephedra sinica}, \textit{Ephedra intermedia}, or \textit{Ephedra equisetina}.
}
\label{sfig:wikipedia}
\end{figure*}

\begin{figure*}
\centering
\includegraphics[width=\textwidth]{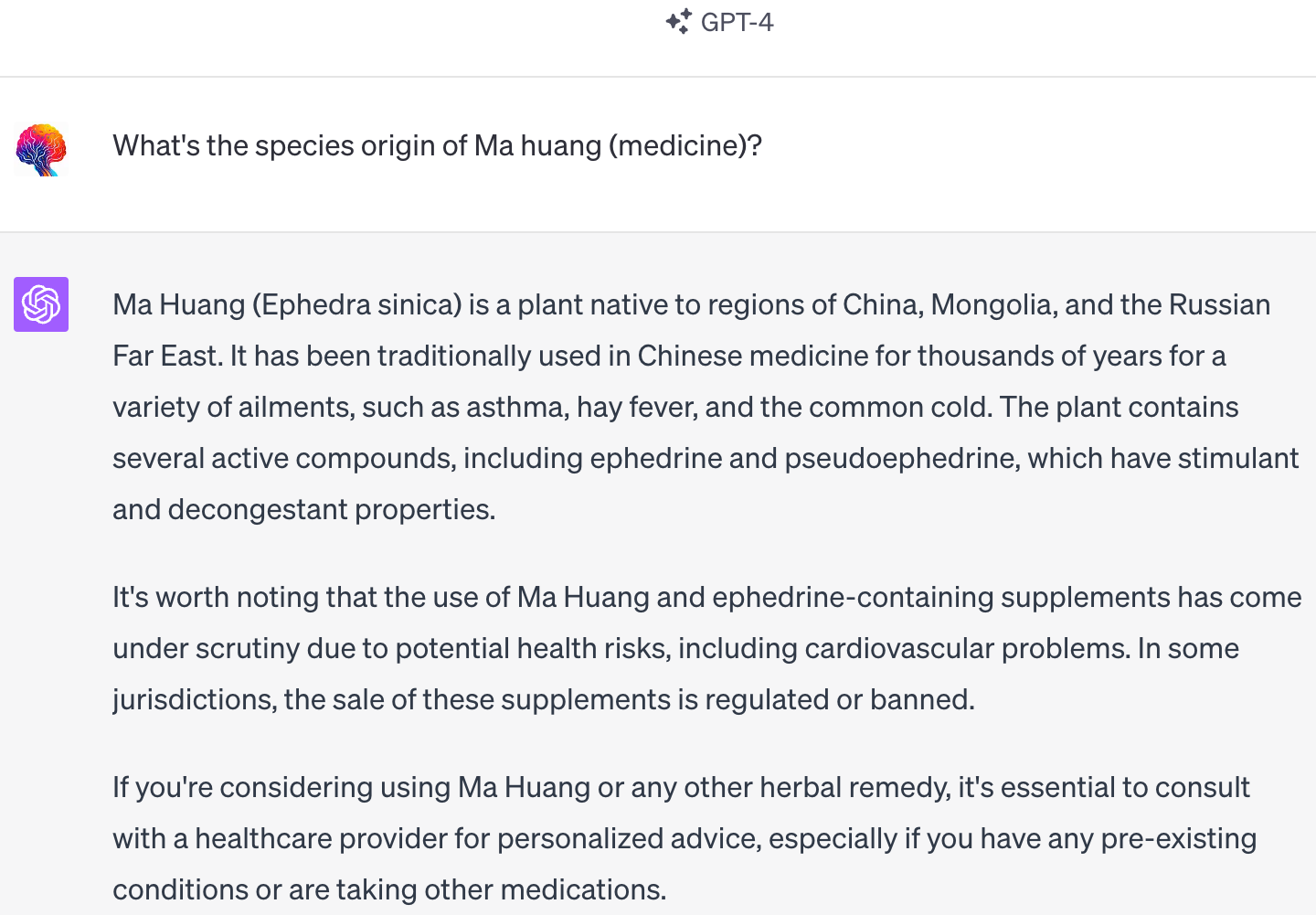}
\caption{
\textbf{GPT-4's answer on the species origin of Ma Huang.} \\
GPT-4 incorrectly asserts that Ma Huang is a single-species NMM based solely on \textit{Ephedra sinica}.
}
\label{sfig:gpt}
\end{figure*}

\begin{figure*}
\centering
\includegraphics[width=\textwidth]{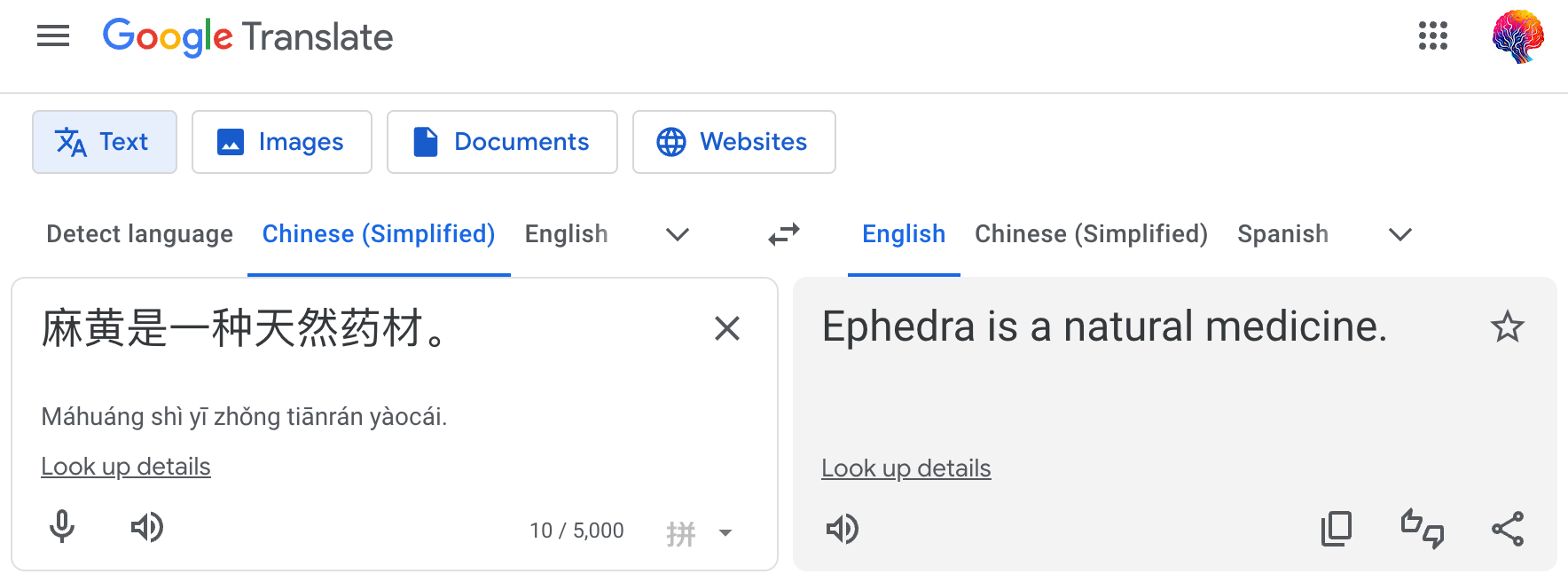}
\caption{
\textbf{Google Translate's translation of text related to ``Ma Huang" (麻黄).} \\
Due to the lack of a systematic nomenclature for NMMs, Google Translate simplifies the translation of ``麻黄" to ``Ephedra," failing to reflect the multi-species intricacies of ``麻黄". This mistranslation could lead English-speaking users to incorrectly use the term ``Ephedra" for further retrieval of knowledge, potentially directing them to the aforementioned Wikipedia entry containing inaccurate information, thereby hindering the globalization of NMM knowledge.
}
\label{sfig:google}
\end{figure*}

\begin{figure*}
\centering
\includegraphics[width=\textwidth]{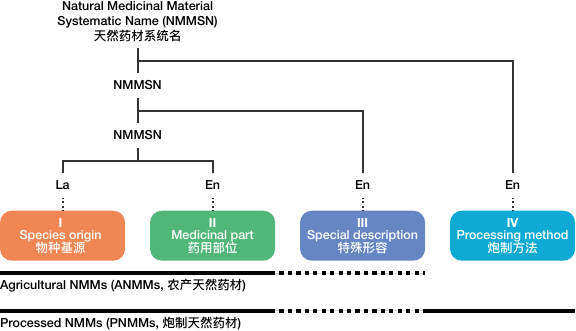}
\caption{
\textbf{Parsing structure of the NMM Systematic Name.}\\
The parsing structure of a valid NMM Systematic Name (NMMSN) involves a lawful combination of four naming components. At a minimum, an NMMSN includes components I and II. For Agricultural NMMs, an NMMSN may include components I, II, and III; for Processed NMMs, it may include components I, II, III, and IV. A solid line indicates that these components are required for that type of NMM; a dashed line indicates that component III is optional.
}
\label{sfig:nmmsn}
\end{figure*}

\begin{figure*}
\centering
\includegraphics[width=\textwidth]{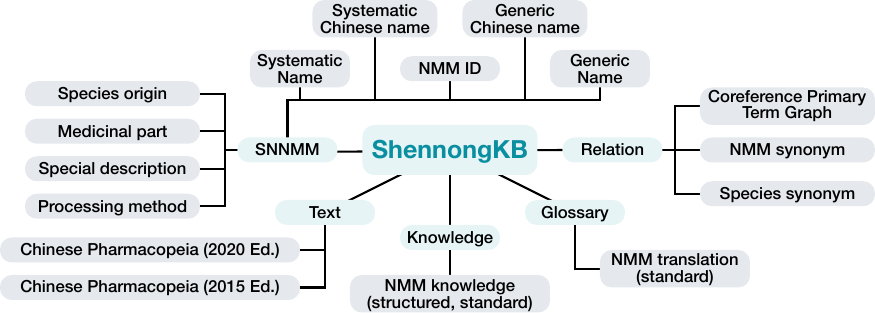}
\caption{
\textbf{NMM knowledge curated in the ShennongAlpha Knowledge Base}\\
The ShennongAlpha Knowledge Base (ShennongKB) encompasses five major categories of information, including the Systematic Nomenclature for Natural Medicinal Materials (SNNMM), text, knowledge, glossary, and relation. 
}
\label{sfig:snkb}
\end{figure*}

\begin{figure*}
\centering
\includegraphics[width=0.65\textwidth]{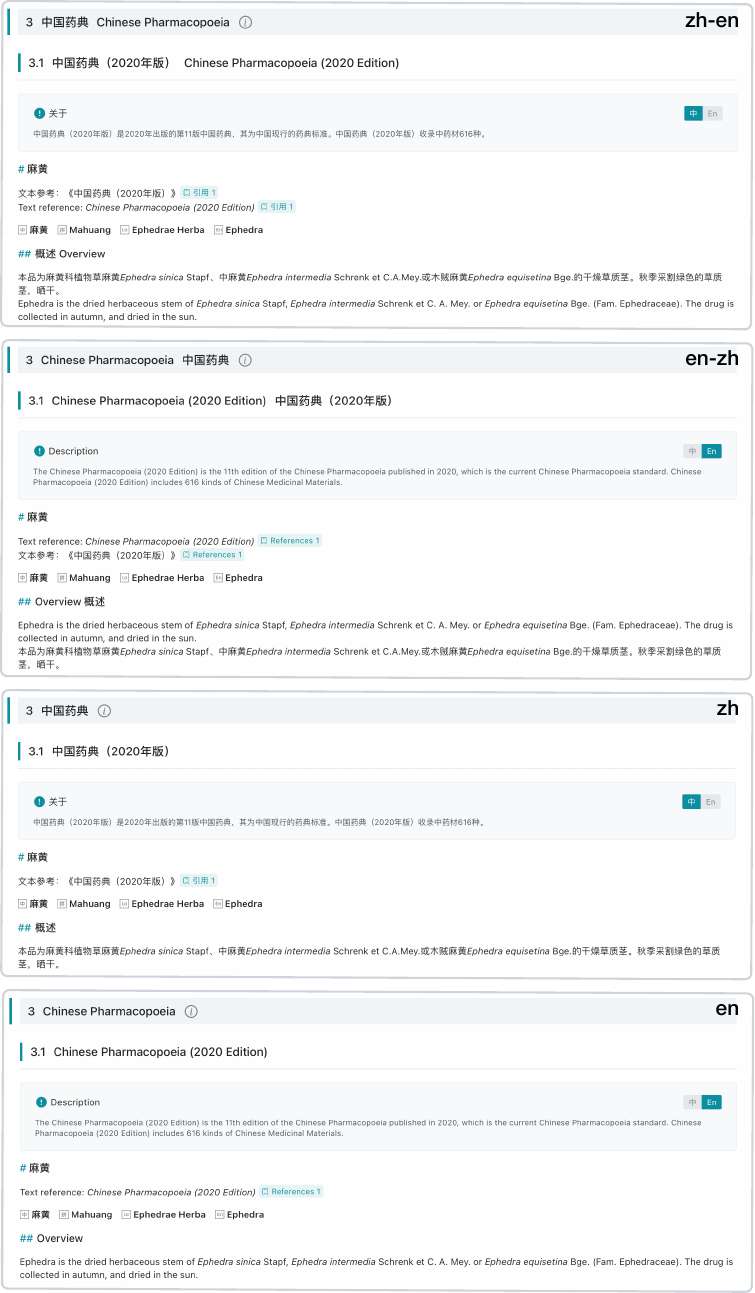}
\caption{
\textbf{Language display modes in the ShennongAlpha Web.} \\
The ShennongAlpha Web adopts a bilingual design in both Chinese and English, utilizing MLMD to curate parallel Chinese and English texts within a single document. ShennongAlpha supports four language display modes: zh-en, en-zh, zh, and en. Notably, in the zh-en and en-zh modes, both the section titles and text of the knowledge are displayed bilingually, differing only in their display order (in the zh-en mode, Chinese precedes English, whereas in en-zh, English comes before Chinese). In the zh and en modes, both the section titles and text of the knowledge are displayed solely in Chinese or English, respectively.
}
\label{sfig:modes}
\end{figure*}

\begin{figure*}
\centering
\includegraphics[width=\textwidth]{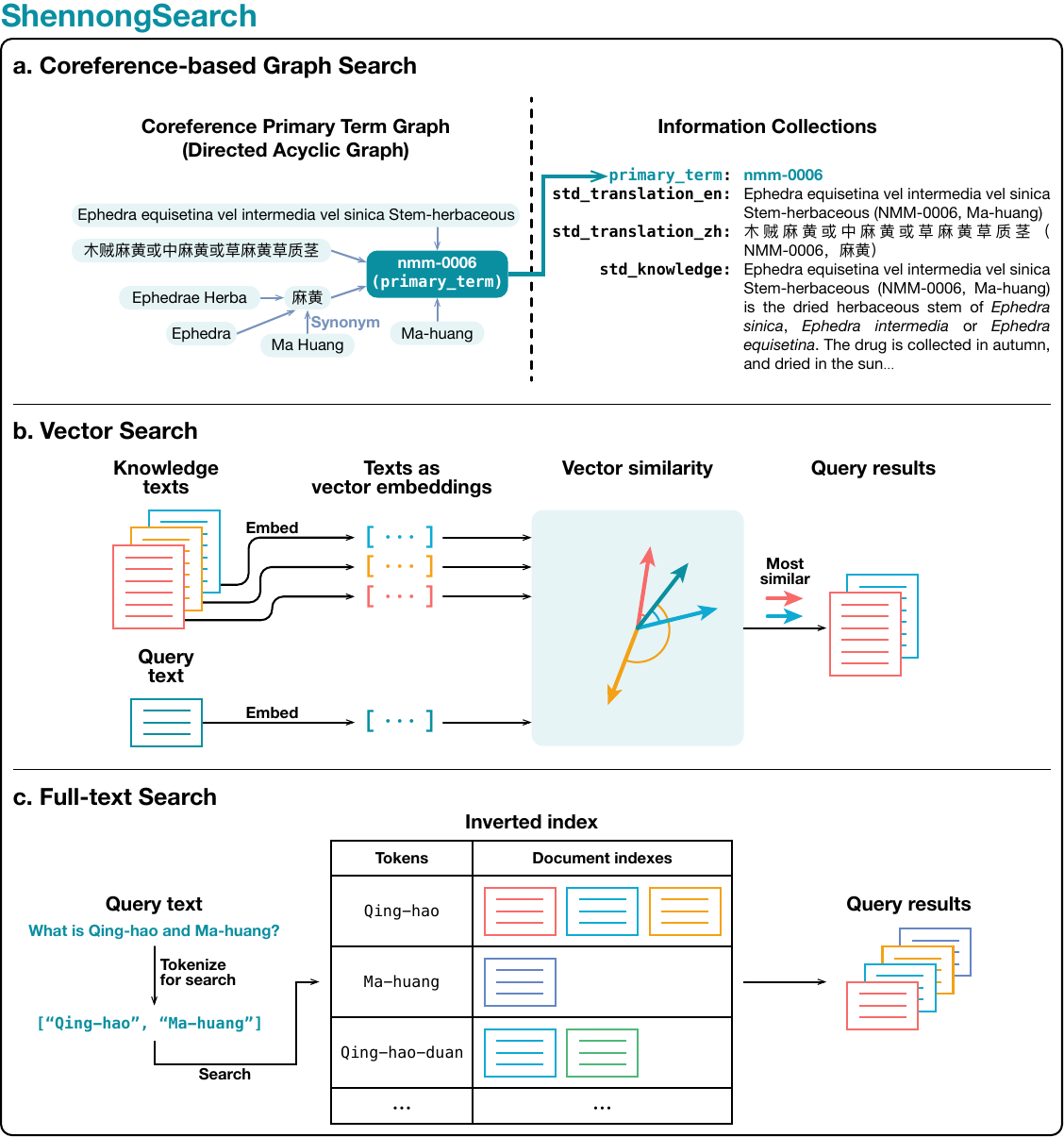}
\caption{
\textbf{Advanced search methods in the ShennongAlpha Search Engine.}\\
The ShennongAlpha Search Engine (ShennongSearch) has been designed to offer three distinct advanced search methods to cater to the specific search needs associated with NMMs: \textbf{a.} Coreference-based Graph Search (CGS), utilizing a Coreference Primary Term Graph (CPTG) to depict the relationships among NMMs' synonyms, thereby facilitating the retrieval of standardized knowledge and translations for a given NMM name; \textbf{b.} Vector search, which evaluates the similarity between the vector embeddings of queries and the texts archived in ShennongKB, allowing for semantic matches; \textbf{c.} Full-text search, employing tokenization and an inverted index to pinpoint and yield the pertinent fuzzy results.
}
\label{sfig:sns}
\end{figure*}

\begin{figure*}
\centering
\includegraphics[width=\textwidth]{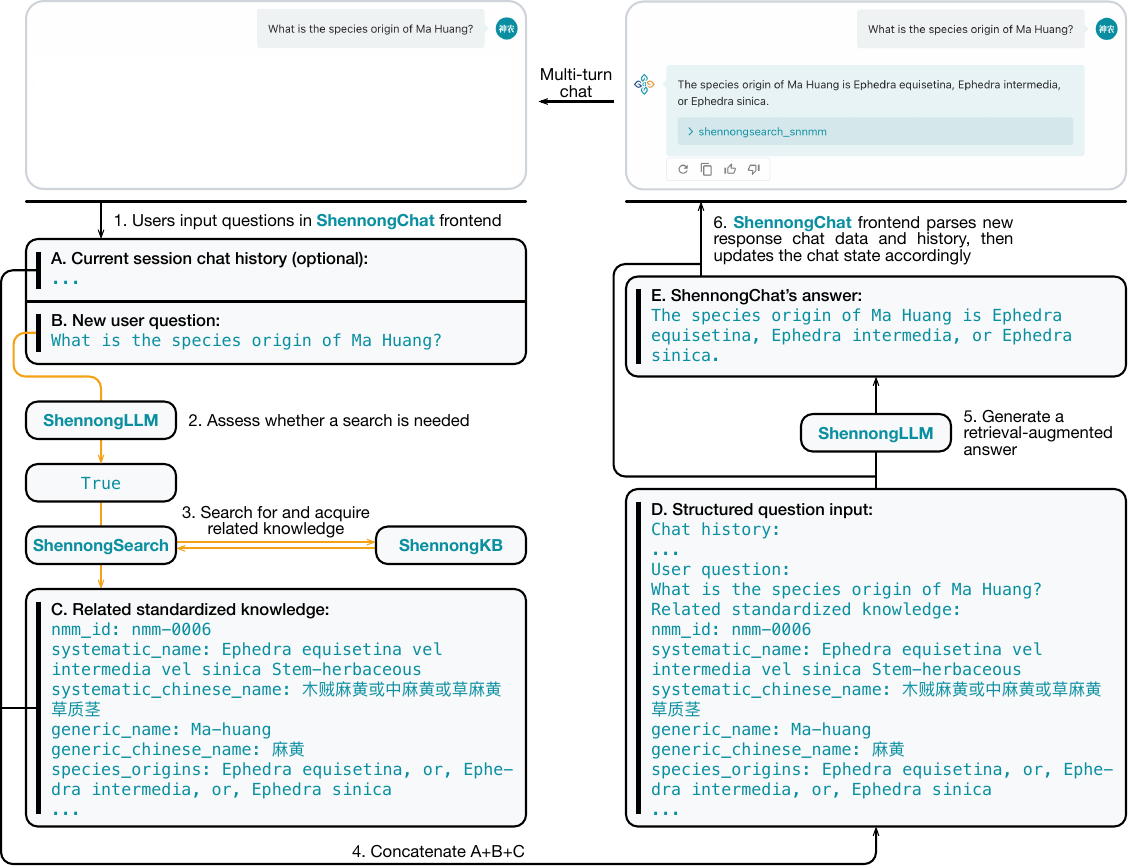}
\caption{
\textbf{Algorithm behind ShennongChat.} \\
Upon receiving a new question from a user, ShennongChat initially employs the ShennongAlpha Large Language Model system (ShennongLLM) to assess if the existing chat history provides sufficient knowledge to formulate an answer. In the presented case, the chat history lacks the necessary knowledge. Consequently, ShennongLLM decides to dispatch ShennongSearch, which delves into ShennongKB to find relevant standardized knowledge. Specifically, knowledge related to the species origin of ``Ma Huang” is retrieved. ShennongChat then integrates the user's previous chat history, the new question, and the retrieved standardized knowledge to generate a retrieval-augmented answer. This answer is subsequently displayed on the ShennongChat user interface.
}
\label{sfig:snc-meth}
\end{figure*}

\begin{figure*}
\centering
\includegraphics[width=\textwidth]{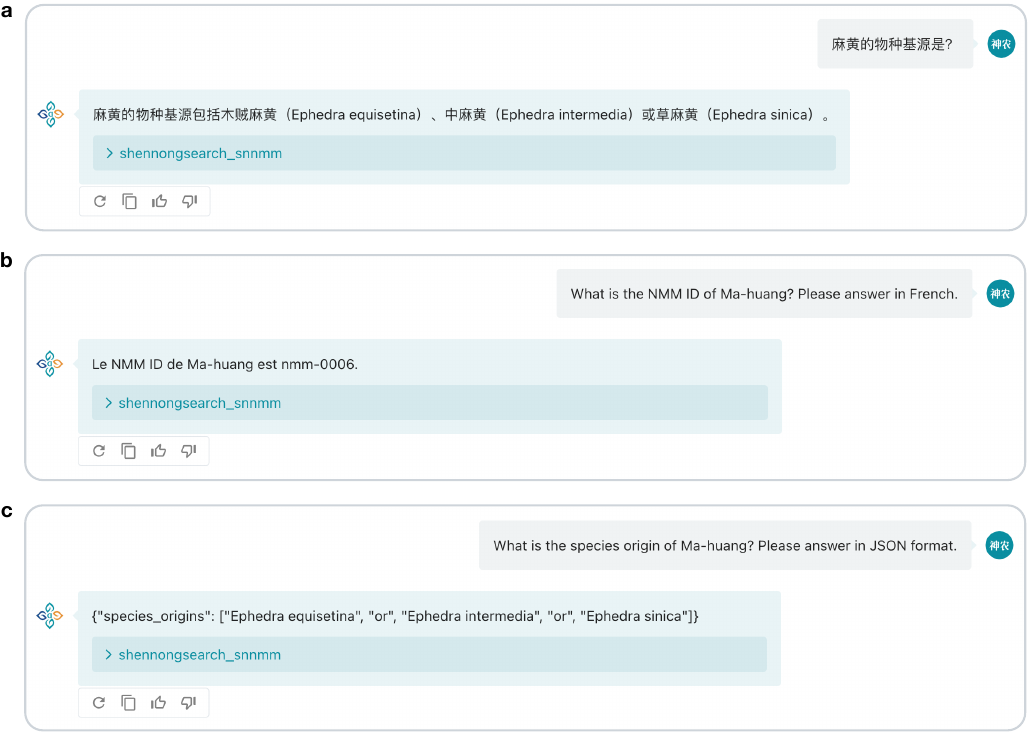}
\caption{
\textbf{ShennongChat offers personalized questioning and answering styles for users.} \\
Users can use their preferred language (such as Chinese) to question ShennongChat and receive answers in the same language (\textbf{a}). Users can also specify their desired answer style by describing it in natural language in their question, such as opting for a particular language (\textbf{b}) or a specific format (\textbf{c}). ShennongChat will tailor its answers to users' indicated styles.
}
\label{sfig:snc-style}
\end{figure*}

\begin{figure*}
\centering
\includegraphics[width=\textwidth]{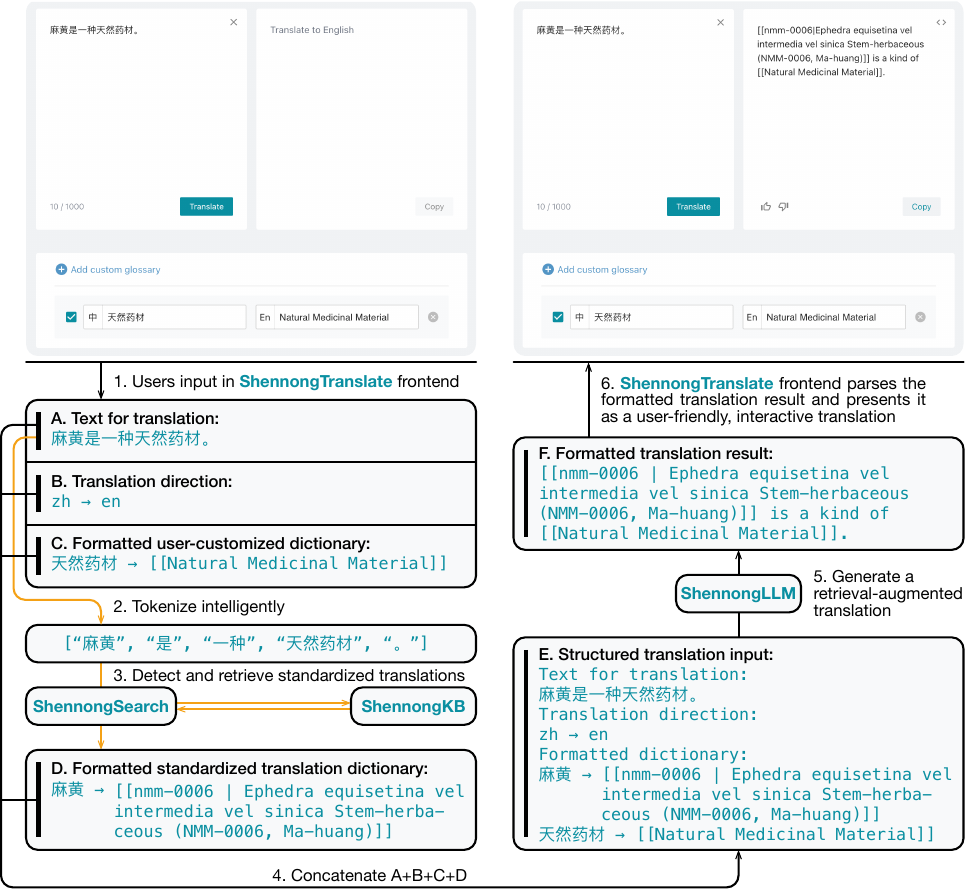}
\caption{
\textbf{Algorithm behind ShennongTranslate.} \\
ShennongTranslate employs the translation algorithm we proposed, named ``Neural Machine Translation based on Coreference Primary Term (NMT-CPT)". Following a sequence from steps 1-6, ShennongTranslate generates a standardized translation with the format shown in F. The annotation ``[[xxx $|$ yyy]]" is a proprietary MLMD syntax used in the NMT-CPT. Here, ``xxx" stands for the Primary Term retrieved from ShennongKB, while ``yyy" denotes its standardized translation in the target language. 
}
\label{sfig:snt-meth}
\end{figure*}

%% file: chapter/stable.tex
\section{Supplementary tables}

\begin{table}[!ht]
\centering
\caption{
\textbf{Problems in the names of 616 NMMs in the \textit{Chinese Pharmacopoeia: 2020 Edition: Volume I}.} \\
Please refer to the attached Excel spreadsheet for specific table contents.
}
\label{stable:chp-name}
\end{table}

\begin{table}[!ht]
\centering
\caption{
\textbf{ShennongAlpha Knowledge Base data counts.} \\
Data as of 2024-05-01. The up-to-date data counts are available on the ShennongAlpha Statistics page (\url{https://shennongalpha.westlake.edu.cn/statistics}).
}
\begin{tabularx}{\textwidth}{lrX}
\toprule
\bfseries Type & \bfseries Count & \bfseries Description \\
\midrule
NMM & 14,256 & Unique NMMs under the Systematic Nomenclature, with distinct NMM IDs, Systematic Names, and Generic Names \\
\hline
NMM knowledge & 14,256 & Structured, standardized knowledge of NMMs \\
\hline
NMM standardized translation & 14,256  & Standardized bilingual (Chinese-English) translations of NMMs \\
\hline
NMM text in ChP-2020 & 616 & NMM monographs and their text in the \textit{Chinese Pharmacopoeia (2020 edition)} \\
\hline
NMM text in ChP-2015 & 618 & NMM monographs and their text in the \textit{Chinese Pharmacopoeia (2015 edition)} \\
\hline
NMM synonym & 58,872 & Synonymous names of NMMs \\
\hline
Species origin & 7,125 & Relevant species origins of the curated NMMs \\
\hline
Medicinal part & 347 & Relevant medicinal parts of the curated NMMs \\
\hline
Processing method & 78 & Relevant processing methods of the curated NMMs \\
\bottomrule
\end{tabularx}
\label{stable:snkb}
\end{table}

%% file: chapter/snnmm.tex
\section{Supplementary text: Systematic Nomenclature for Natural Medicinal Materials (SNNMM)}
\label{appendix:snnmm}

\begin{ruleen}
    \item Two names one ID rule. Each Natural Medicinal Material (NMM) is assigned an NMM Systematic Name (NMMSN), an NMM Generic Name (NMMGN), and an NMM ID.
    \item Uniqueness rule. The NMMSN, NMMGN, and NMM ID generated under SNNMM are all unique.
    \item ANMM/PNMM distinct name rule. SNNMM classifies NMMs into 3 types: Raw NMMs (RNMMs), Agricultural NMMs (ANMMs), and Processed NMMs (PNMMs) \boldcolor{(\cref{fig:snnmm}b)}. RNMMs are generally regulated as agricultural products, while PNMMs are generally regulated as drugs.
    \item Legitimate character rule. To allow people from any country to easily input NMMSN or NMMGN using their standard keyboard, naming NMMSN and NMMGN can only use the following characters: a-zA-Z, and the hyphen ``-".
    \item Case-insensitive rule. In NMMSN or NMMGN, even though both uppercase and lowercase letters might be used, it's only for ease of reading. Different NMMs cannot be distinguished by different cases in the SNNMM.
    \item Upright type rule. NMMSN or NMMGN should be written in upright type.
    \item NMMSN rules.
    \begin{ruleen}
        \item Standard Chinese NMMSN correspondence rule. To facilitate communication in a Chinese context, for each NMM named with NMMSN, SNNMM will also provide a corresponding standard Chinese systematic name, i.e., NMMSN-zh.
        \item NMMSN composition Rule. NMMSN consists of the following 4 name components \boldcolor{(\cref{fig:snnmm}a)}: I. Species origin; II. Medicinal part; III. Special description; IV. Processing method.
        \item NMMSN syntax parsing rule. Based on the principles of natural language processing and computational linguistics, the syntax parsing structure of NMMSN is as shown in \boldcolor{\cref{sfig:nmmsn}}.
        \item Minimal NMMSN composition rule. The NMMSN of ANMM must contain at least the first two name components: I and II; The NMMSN of PNMM must contain at least three name components: I, II, and IV \boldcolor{(\cref{sfig:nmmsn})}.
        \item NMMSN order rule. The sequence of name components in an NMMSN must adhere to a specific order: I-II-III-IV. In the case of NMMSN-zh, which follows Chinese conventions, the components are ordered as IV-III-I-II.
        
        \item Species origin naming rules.
        \begin{ruleen}
            \item Complete taxonomic name rule. Prioritize the use of the full Latin taxonomic name of a species when naming. When a specific species can be identified, do not just use the genus name or species epithet for naming.
            \item[Ex. ] \cmark NMMSN: Solidago decurrens Herb \cite{nmm-0020}; NMMSN-zh: 一枝黄花全草 (species origin: \textit{Solidago decurrens}) \\
            \xmark Solidago Herb
            \item[Ex. ] \cmark NMMSN: Panax ginseng Rhizome and Root \cite{nmm-001l}; NMMSN-zh: 人参根茎与根 (species origin: \textit{Panax ginseng}) \\
            \xmark Ginseng Rhizome and Root
            \item Standard taxonomic name rule. When naming NMMSN, prioritize using the current standard taxonomic name. The standard taxonomic name should primarily refer to the following databases: ``The Catalogue of Life" (\url{https://www.catalogueoflife.org}) and ``Catalogue of Life China" (\url{http://www.sp2000.org.cn}).
            \item[Ex. ] \cmark \textit{Tetradium ruticarpum} \xmark \textit{Euodia ruticarpa} \xmark \textit{Euodia ruticarpa var. officinalis} \xmark \textit{Euodia ruticarpa var. bodinieri}\\
            In the ``Chinese Pharmacopoeia: 2020 Edition: Volume I" (hereinafter referred to as ``ChP") \cite{chp2020vol1}, the NMM ``吴茱萸" is recorded as a multi-species NMM, based on \textit{Euodia ruticarpa} or \textit{Euodia ruticarpa var. officinalis} or \textit{Euodia ruticarpa var. bodinieri}. However, all three of these scientific names are non-standard. The correct name for all of them is \textit{Tetradium ruticarpum}. Therefore, ``吴茱萸" is still actually a single-species NMM. Its correct NMMSN is:\\
            NMMSN: Tetradium ruticarpum Fruit \cite{nmm-012w}; NMMSN-zh: 吴茱萸果实
            \item Multiple species origins naming rule. When the same part of multiple species is used as the medicinal part of an NMM, and these species are interchangeable (i.e., in an ``or" relationship), in order to clarify the specific species origins of the NMM, all species origins should be fully listed in the NMMSN. Taxonomic names of species are listed side by side, connected by the term ``vel" (Latin conjunction meaning ``or") (In NMMSN-zh, the word ``或" is used), and arranged in alphabetical order by the Latin names. When the genus name appears repeatedly among the multiple species, the genus name can be omitted the second time it appears.
            \item[Ex. ] Multiple species originate from the same genus. In ChP, ``麻黄" has 3 species origins: \textit{Ephedra sinica} or \textit{Ephedra intermedia} or \textit{Ephedra equisetina}. Thus, it should be named as:\\
            Ephedra equisetina vel Ephedra intermedia vel Ephedra sinica Stem-herbaceous\\
            However, because they share the same genus name, it should be further abbreviated as:\\
            NMMSN: Ephedra equisetina vel intermedia vel sinica Stem-herbaceous \cite{nmm-0006}; NMMSN-zh: 木贼麻黄或中麻黄或草麻黄草质茎
            \item[Ex. ] Multiple species originate from different genera. For example:\\
            Cremastra appendiculata vel Pleione bulbocodioides vel Pleione yunnanensis Pseudobulb \\
            After abbreviating the genus name for the last two species, it becomes:\\
            NMMSN: Cremastra appendiculata vel Pleione bulbocodioides vel yunnanensis Pseudobulb \cite{nmm-00ac}; NMMSN-zh: 杜鹃兰或独蒜兰或云南独蒜兰假鳞茎
            \item Refinement of species origin rule. When a more refined species origin can be specified, we should prioritize using this more refined species origin for referring to the NMM.
            \item[Ex. ] For ``麻黄" with more refined species origin, we can adopt the following names:\\
            NMMSN: Ephedra sinica Stem-herbaceous \cite{nmm-0003}; NMMSN-zh: 草麻黄草质茎\\
            NMMSN: Ephedra intermedia Stem-herbaceous \cite{nmm-0004}; NMMSN-zh: 中麻黄草质茎\\
            NMMSN: Ephedra equisetina Stem-herbaceous \cite{nmm-0005}; NMMSN-zh: 木贼麻黄草质茎
            \item Species inclusion rule. For a multi-species NMM, if the taxonomic levels of its species origins are in a hierarchical relationship, only the species with the highest hierarchical level is used for naming.
            \item[Ex. ] ChP records the species origins of ``山楂" as \textit{Crataegus pinnatifida} and \textit{Crataegus pinnatifida var. major}. However, since \textit{Crataegus pinnatifida var. major} is a variety of \textit{Crataegus pinnatifida}, the latter includes the former, so we should not include this variety as a species origin of ``山楂" in the naming.\\
            \cmark NMMSN: Crataegus pinnatifida Fruit \cite{nmm-009x}; NMMSN-zh: 山楂果实\\
            \xmark Crataegus pinnatifida vel pinnatifida var major Fruit\\
            The NMM with more refined species origin can be named separately to clarify:\\
            \cmark NMMSN: Crataegus pinnatifida var major Fruit \cite{nmm-009y}; NMMSN-zh: 山里红果实
            \item Naming rule for NMMs with uncertain specific species information. For NMMs where the specific species information is uncertain, the naming can be done using only the genus name, followed by ``unspecified" (in NMMGN-zh, add ``未定种") to indicate that the species information has yet to be clarified.
            \item[Ex. ] There are thousands of species under the genus \textit{Taraxacum}, all of which could potentially be used as an NMM:\\
            NMMSN: Taraxacum unspecified Herb \cite{nmm-01yf}; NMMSN-zh: 蒲公英属未定种全草
            \item Rule for omitting non-legitimate characters in taxonomic names. If the taxonomic name of the species origin of the NMM contains non-legitimate characters, these should be omitted.
            \item[Ex. ] \cmark NMMSN: Ziziphus jujuba var spinosa Seed \cite{nmm-005x}; NMMSN-zh 酸枣种子\\
            \xmark Ziziphus jujuba var. spinosa Seed
        \end{ruleen}

        \item Medicinal part naming rules.
        \begin{ruleen}
            \item Syntax rule. The medicinal part in NMMSN should be named using singular nouns or noun phrases in English.
            \item[Ex. ] \cmark NMMSN: Panax ginseng Leaf \cite{nmm-001s}; NMMSN-zh: 人参叶\\
            \xmark Panax ginseng Leaves
            \item Capitalization and hyphenation rule. For readability, the first letter of the medicinal part should be capitalized. If the medicinal part consists of multiple words, they should be connected with a hyphen ``-".
            \item[Ex. ] NMMSN: Solidago decurrens Herb \cite{nmm-0020}; NMMSN-zh: 一枝黄花全草
            \item[Ex. ] NMMSN: Ephedra sinica Stem-herbaceous \cite{nmm-0003}; NMMSN-zh: 草麻黄草质茎
            \item Multiple medicinal parts naming rule. If an NMM uses multiple different medicinal parts from the same species, they can be connected with ``and" (for NMMSN-zh, using ``与"). If an NMM can use multiple interchangeable medicinal parts from the same species, they can be connected with ``or" (for NMMSN-zh, using ``或"). The order of medicinal parts is based on the alphabetical order.
            \item[Ex. ] NMMSN: Vincetoxicum pycnostelma Rhizome and Root \cite{nmm-01mz}; NMMSN-zh: 徐长卿根茎与根
            \item Refinement of medicinal part rule. When a more specific medicinal part can be identified, we should prioritize using the refined medicinal part to refer to the NMM. This naming rule helps us to clarify the medicinal part of the NMM further.
            \item[Ex. ] ``Vincetoxicum pycnostelma Rhizome and Root" can be further refined as: \\
            NMMSN: Vincetoxicum pycnostelma Rhizome \cite{nmm-01mx}; NMMSN-zh: 徐长卿根茎\\
            NMMSN: Vincetoxicum pycnostelma Root \cite{nmm-01my}; NMMSN-zh: 徐长卿根
        \end{ruleen}

        \item Special descriptions naming rules.
        \begin{ruleen}
            \item Syntax rule. Some NMMs require specific characteristics or need to undergo special initial preparation at the production site before they can be used as medicine. For these NMMs, adjectives, adjective phrases, nouns, or appositives can be used for naming in element III.
            \item Capitalization and hyphenation rule. For readability, the first letter of the special descriptions should be capitalized. If the descriptions consist of multiple words, they should be connected with a hyphen ``-".
            \item[Ex. ] NMMSN: Zingiber officinale Rhizome \cite{nmm-003g}; NMMSN-zh: 姜根茎\\
            NMMSN: Zingiber officinale Rhizome Fresh \cite{nmm-003m}; NMMSN-zh: 鲜姜根茎
            \item[Ex. ] NMMSN: Curcuma wenyujin Rhizome \cite{nmm-000t}; NMMSN-zh: 温郁金根茎\\
            NMMSN: Curcuma wenyujin Rhizome Freshly-sliced \cite{nmm-0015}; NMMSN-zh: 鲜切片温郁金根茎
            \item Rule for genuine regional NMMs (道地药材). For some NMMs that specifically need to be sourced from genuine regions, geographical nouns can be used as special descriptions. For genuine regional Chinese NMMs, the standard English names of the provincial capitals of the People's Republic of China are generally used, while in NMMSN-zh, the abbreviation of the provincial capital plus ``产" (meaning produced in) is used.
            \item[Ex. ] NMMSN: Fritillaria thunbergii Bulb Zhejiang \cite{nmm-024c}; NMMSN-zh: 浙产浙贝母鳞茎
        \end{ruleen}

        \item Processing methods naming rules.
        \begin{ruleen}
            \item Syntax rule. PNMM must include a processing method. PNMM is named based on its corresponding ANMM by adding an English adjective, adjective phrase, appositive, or abbreviation related to the processing method in name element IV. The corresponding Chinese word for the processing method should end with ``制".
            \item[Ex. ] ANMM: NMMSN: Artemisia annual Part-aerial \cite{nmm-0001}; NMMSN-zh: 黄花蒿地上部\\
            PNMM: NMMSN: Artemisia annual Part-aerial Segmented \cite{nmm-0002}; NMMSN-zh: 段制黄花蒿地上部
            \item Capitalization and hyphenation rule. For readability, the first letter of the processing method should be capitalized. If the processing method consists of multiple words, they should be connected using a hyphen ``-".
            \item Naming rule for NMM with multiple processing methods. For PNMMs that require multiple processing methods, they should be connected using ``and", with the term for the later processing method placed after (In NMMSN-zh, processing methods, because they end in ``制", do not need logical connectors; the term for the later processing method is placed before). The sequence of processing methods is related to the processing order, so the order of the words cannot be changed arbitrarily.
            \item[Ex. ] \cmark NMMSN: Ephedra sinica Stem-herbaceous Segmented and Aquafried-honey \cite{nmm-000b}; NMMSN-zh: 蜜炙制段制草麻黄草质茎\\
            \xmark Ephedra sinica Stem-herbaceous Aquafried-honey and Segmented
            \item Naming rule for processing methods of Chinese NMMs. The processing methods for Chinese NMMs are based on ``0231 General Rules for Processing" in the ``Chinese Pharmacopoeia: 2020 Edition: Volume IV" \cite{chp2020vol4}. When naming the processing method, if a specific type of processing can be determined, it should be used; if not, the general category can be used.
            \item[Ex. ] NMMSN: Crataegus pinnatifida Fruit Cleaned and Stirfried-golden \cite{nmm-00a1}; NMMSN-zh: 炒黄制净制山楂果实\\
            NMMSN: Crataegus pinnatifida Fruit Cleaned and Stirfried-charred \cite{nmm-00a3}; NMMSN-zh: 炒焦制净制山楂果实
            \item Classification rule for natural medicinal processing (炮制) methods of Chinese NMMs. There are three main categories of processing methods for Chinese NMMs: processing by cleaning (净制), processing by cutting (切制), and processing by preparing (备制). Chinese NMMs that have undergone processing by cutting are assumed to have been processed by cleaning. Any that needs processing by preparing must first go through processing by cleaning or cutting.
            \item[Ex. ] \cmark NMMSN: Ephedra sinica Stem-herbaceous Segmented and Aquafried-honey; NMMSN-zh \cite{nmm-000b}: 蜜炙制段制草麻黄草质茎\\
            \xmark Ephedra sinica Stem-herbaceous Aquafried-honey
            \item[Ex. ] \cmark NMMSN: Zingiber officinale Rhizome Cleaned and Stirfried-sand \cite{nmm-003l}; NMMSN-zh: 砂炒制净制姜根茎\\
            \xmark Zingiber officinale Rhizome Stirfried-sand
        \end{ruleen}

        \item Naming rule for NMMs of non-species origin or with unclear species origin. If the NMM to be named is non-species or its species origin is hard to determine, its NMMSN can be named using an English common name.
        \item[Ex. ] NMMSN: Talc \cite{nmm-01xx}; NMMSN-zh: 滑石\\
        NMMSN: Talc Pulverized \cite{nmm-01xz}; NMMSN-zh: 粉制滑石
        
    \end{ruleen}

    \item NMMGN Rules.
    \begin{ruleen}
        \item Conciseness rule. Owing to the detailed naming requirements of NMMSN, which include specifying the species origin, medicinal part, special description, and processing method, NMMSN tends to be lengthy. To facilitate everyday usage and clinical prescription, each NMM is also assigned a corresponding shorter NMMGN.
        \item Conventional naming rule for Chinese NMMGN-zh of Chinese NMMs. The NMMGN-zh of a Chinese NMM prefers the commonly used name. If the NMM is from the ChP, the Chinese name listed in the ChP is generally the NMMGN-zh. For NMMs originated from a refined single species, to prevent naming conflicts, the prefix ``单"/``独" (meaning ``single") is added to the beginning of some NMMGN-zh to differentiate them.
        \item Pinyin naming priority rule for NMMGN of Chinese NMM. To correspond with NMMGN-zh, we prefer the pinyin name of the NMMGN-zh when naming the NMMGN. The pinyin does not use tone marks or spaces but uses hyphens to connect the pinyin of different characters, capitalizes the first letter, and uses ``v" instead of ``ü".
        \item[Ex. ] NMMSN: Erycibe obtusifolia Stem \cite{nmm-0022}; NMMSN-zh: 丁公藤茎; NMMGN: Dan-ding-gong-teng; NMMGN-zh: 单丁公藤 (To differentiate with ``丁工藤", add prefix ``单")\\
        NMMSN: Erycibe obtusifolia vel schmidtii Stem \cite{nmm-0024}; NMMSN-zh: 丁公藤或光叶丁公藤茎; NMMGN: Ding-gong-teng; NMMGN-zh: 丁公藤
        \item[Ex. ] NMMSN: Ligustrum lucidum Fruit \cite{nmm-00d7}; NMMSN-zh: 女贞果实; NMMGN: Nv-zhen-zi; NMMGN-zh: 女贞子
        \item Minimum length rule for NMMGN-zh. NMMGN-zh must be named using two or more characters.
        \item[Ex. ] NMMSN: Prunus mume Fruit \cite{nmm-008w}; NMMSN-zh: 梅果实\\
        NMMGN-zh: \cmark 乌梅 \xmark 梅
        \item First-come-first-served rule. Due to the brevity of NMMGN, naming conflicts might arise. In such cases, we adhere to the priority of the first-come NMMGN, and subsequent NMMGNs must incorporate additional information for differentiation.
        \item[Ex. ] Suppose we have already named the following NMM:\\
        NMMSN: Ephedra sinica Stem-herbaceous \cite{nmm-0003}; NMMSN-zh: 草麻黄草质茎; NMMGN: Cao-ma-huang; NMMGN-zh: 草麻黄\\
        In the aforementioned case, the NMMGN-zh of the NMM did not mention its medicinal part information since it was prioritized, and thus omitted. But if we were to further name the root of \textit{Ephedra sinica}, its NMMGN, and NMMGN-zh would need appropriate differentiation:\\
        NMMSN: Ephedra sinica Root \cite{nmm-000g}; NMMSN-zh: 草麻黄根; NMMGN: Cao-ma-huang-gen; NMMGN-zh: 草麻黄根
    \end{ruleen}

    \item NMM ID rules. Each NMM is assigned a unique NMM ID. The encoding rule for NMM ID is: \lstinline{NMM-XXXX}, where \lstinline{XXXX} is a 4-digit number in base 36 (i.e., 0-9, A-Z), starting from \lstinline{0001} and increasing to \lstinline{ZZZZ}, encoding up to $36^4 - 1 = 1,679,615$ kinds of NMMs. NMM ID is case-insensitive, but for ease of reading, it is usually written in uppercase.
    \item[Ex. ] NMM-ID: NMM-0001 \cite{nmm-0001}\\
    NMMSN: Artemisia annua Part-aerial\\
    NMMSN-zh: 黄花蒿地上部\\
    NMMGN: Qing-hao\\
    NMMGN-zh: 青蒿
    \item[Ex. ] NMM-ID: NMM-0002 \cite{nmm-0002}\\
    NMMSN: Artemisia annua Part-aerial Segmented\\
    NMMSN-zh: 段制黄花蒿地上部\\
    NMMGN: Qing-hao-duan\\
    NMMGN-zh: 青蒿段
    
    \item Standard referencing rules.
    \begin{ruleen}
        \item First complete appearance rule. To ensure that the NMM name in a scientific text accurately corresponds to the actual NMM it refers to, it is recommended to present the ``NMMSN (NMM-ID, NMMGN)" format when an NMM first appears in an independent text (such as a paper or encyclopedia). In subsequent appearances of the NMM within the text, only NMMSN, NMMGN, or NMM ID may be used for reference.
        \item[Ex. ] Artemisia annua Part-aerial (NMM-0001, Qing-hao) is a commonly used Chinese natural medicinal material. Artemisia annua Part-aerial has the effect of treating malaria.
        \item[Ex. ] Artemisia annua Part-aerial (NMM-0001, Qing-hao) is a commonly used Chinese natural medicinal material. Qing-hao has the effect of treating malaria.
        \item[Ex. ] Artemisia annua Part-aerial (NMM-0001, Qing-hao) is a commonly used Chinese natural medicinal material. NMM-0001 has the effect of treating malaria.
        \item[Ex. ] 黄花蒿地上部（NMM-0001，青蒿）是一种常用的中药材。黄花蒿地上部可用于治疗疟疾。
        \item[Ex. ] 黄花蒿地上部（NMM-0001，青蒿）是一种常用的中药材。青蒿可用于治疗疟疾。
        \item[Ex. ] 黄花蒿地上部（NMM-0001，青蒿）是一种常用的中药材。NMM-0001可用于治疗疟疾。
        \item Appendix complete information rule. If a study involves a large number of NMMs, presenting all NMMSN, NMMGN, and NMM ID in the main text can make it excessively lengthy. In such cases, only one of the NMMSN, NMMGN, or NMM ID can be used to refer to the NMM in the main text; however, in the appendix, a complete list in the format ``NMM ID -- NMMSN -- NMMGN" should be provided.
    \end{ruleen}
        
\end{ruleen}

\clearpage

\section{Supplementary text: Systematic Nomenclature for Natural Medicinal Materials (SNNMM, 天然药材系统命名法, Chinese version)}
\label{appendix:snnmm_zh}

\begin{rulezh}
    \item 两名一ID规则。每个天然药材 (Natural Medicinal Material, NMM) 均被被赋予一个天然药材系统名 (NMM Systematic Name, NMMSN)、一个天然药材通用名 (NMM Generic Name, NMMGN) 和一个天然药材ID (NMM ID)。
    \item 唯一规则。在SNNMM下产生的NMMSN、NMMGN和NMM ID均唯一。
    \item ANMM/PNMM不同名规则。SNNMM将NMMs分为3类，原始天然药材 (Raw NMMs, RNMMs)、农产天然药材 (Agricultural NMMs, ANMMs) 和炮制天然药材 (Processed NMMs, PNMMs) \boldcolor{(\cref{fig:snnmm}b)}。RNMMs一般按照农产品进行管理，PNMMs一般按照药品进行管理。
    \item 合法字符规则。为便于世界上任何一个国家的人们都能够通过他们国家的标准键盘简单的输入NMMSN或NMMGN，NMMSN和NMMGN的命名只允许使用以下字符：大小写拉丁字母a-zA-Z、连词符“-”。
    \item 大小写不敏感规则。在NMMSN或NMMGN中，尽管会有大小写字母同时使用的情况，但这仅仅是为了便于NMMSN和NMMGN的阅读的便利。SNNMM中，不得通过采用不同大小写的方式以区分不同的NMM。
    \item 正体规则：NMMSN或NMMGN采用正体书写。
    \item NMMSN规则。
    \begin{rulezh}
        \item NMMSN标准中文对应名规则。为便于中文语境下的交流，每个NMM在命名NMMSN的同时，SNNMM也会给出NMMSN的对应的标准中文对应名，即天然药材系统中文名（NMMSN-zh）。
        \item NMMSN构词规则。NMMSN由以下4种命名组件构成 \boldcolor{(\cref{fig:snnmm}a)}：I. 物种基源 (Species origin)、II. 药用部位 (Medicinal part)、III. 特殊形容 (Special description)、IV. 炮制方法 (Processing method)。
        \item NMMSN语法分析规则。根据自然语言处理和计算语言学原理，NMMSN的语法分析 (Parsing) 结构如\boldcolor{\cref{sfig:nmmsn}}所示。
        \item NMMSN最小构词规则。ANMM的NMMSN至少包含以下2种命名组件：I和II；PNMM的NMMSN至少包含以下3种命名组件：I、II和IV \boldcolor{(\cref{sfig:nmmsn})}。
        \item NMMSN语序规则。NMMSN命名组件的语序不可改变，语序为：I-II-III-IV；对于NMMSN-zh，命名组件为符合中文习惯，语序为：IV-III-I-II。
        
        \item 物种基源命名规则。
        \begin{rulezh}
            \item 完整物种学名规则。命名时尽可能优先使用物种的完整的拉丁文物种学名。在可以明确具体物种时，不得仅采用物种的属名或种加词进行命名。
            \item[Ex. ] \cmark NMMSN: Solidago decurrens Herb \cite{nmm-0020}; NMMSN-zh: 一枝黄花全草（物种基源：\textit{Solidago decurrens}）\\
            \xmark Solidago Herb
            \item[Ex. ] \cmark NMMSN: Panax ginseng Rhizome and Root \cite{nmm-001l}; NMMSN-zh: 人参根茎与根（物种基源：\textit{Panax ginseng}）\\
            \xmark Ginseng Rhizome and Root
            \item 标准物种学名规则。NMMSN命名时优先使用现行的标准物种学名。标准物种学名优先参考以下数据库：“The Catalogue of Life”（\url{https://www.catalogueoflife.org}）、“中国生物物种名录”（\url{http://www.sp2000.org.cn}）。
            \item[Ex. ] \cmark \textit{Tetradium ruticarpum} \xmark \textit{Euodia ruticarpa} \xmark \textit{Euodia ruticarpa var. officinalis} \xmark \textit{Euodia ruticarpa var. bodinieri}\\
            在《中国药典·2020年版·一部》（下简称“ChP”） \cite{chp2020vol1}中记载NMM“吴茱萸”为多物种基源NMM：基于\textit{Euodia ruticarpa}或\textit{Euodia ruticarpa var. officinalis}或\textit{Euodia ruticarpa var. bodinieri}。然而，这三个物种学名均非标准物种学名，其正名均为\textit{Tetradium ruticarpum}。因此，“吴茱萸”实际仍然为单物种基源NMM。其正确NMMSN为：\\
            NMMSN: Tetradium ruticarpum Fruit \cite{nmm-012w}; NMMSN-zh: 吴茱萸果实
            \item 多物种基源命名规则。多种物种的相同部位用作一个NMM的药用部位，且这些物种互为可替代关系（即“或”关系）时，为了明确NMM的具体物种基源，NMMSN命名时要完整列出所有的物种基源。物种学名并列，中间采用“vel”（拉丁文连词，意为“或”）连接（NMMSN-zh中使用“或”连接），排序时以拉丁字母为顺序。当多物种中出现重复的属名时，属名第二次出现时可省略。
            \item[Ex. ] 多物种基源来自相同属物种。ChP中，“麻黄”有3种物种基源：\textit{Ephedra sinica}或\textit{Ephedra intermedia}或\textit{Ephedra equisetina}。因此应当命名为：\\
            Ephedra equisetina vel Ephedra intermedia vel Ephedra sinica Stem-herbaceous\\
            但由于属名一致，须进一步缩写为：\\
            NMMSN: Ephedra equisetina vel intermedia vel sinica Stem-herbaceous \cite{nmm-0006}; NMMSN-zh: 木贼麻黄或中麻黄或草麻黄草质茎
            \item[Ex. ] 多物种基源来自不同属物种。如：\\
            Cremastra appendiculata vel Pleione bulbocodioides vel Pleione yunnanensis Pseudobulb \\
            须进一步缩写后两种物种的属名：\\
            NMMSN: Cremastra appendiculata vel Pleione bulbocodioides vel yunnanensis Pseudobulb \cite{nmm-00ac}; NMMSN-zh: 杜鹃兰或独蒜兰或云南独蒜兰假鳞茎
            \item 细化物种基源规则。当能够明确更细化的物种基源时，我们应当优先使用更细化的物种基源对NMM进行指代。
            \item[Ex. ] 我们可以对具有更细化的物种基源的“麻黄”采取以下命名：\\
            NMMSN: Ephedra sinica Stem-herbaceous \cite{nmm-0003}; NMMSN-zh: 草麻黄草质茎\\
            NMMSN: Ephedra intermedia Stem-herbaceous \cite{nmm-0004}; NMMSN-zh 中麻黄草质茎\\
            NMMSN: Ephedra equisetina Stem-herbaceous \cite{nmm-0005}; NMMSN-zh 木贼麻黄草质茎
            \item 物种包含规则。对于一个多物种基源NMM，若其物种基源的物种分类等级存在包含关系，则仅使用具有最高包含等级的物种进行命名。
            \item[Ex. ] ChP记载“山楂”的物种基源为\textit{Crataegus pinnatifida}和\textit{Crataegus pinnatifida var. major}。但由于\textit{Crataegus pinnatifida var. major}是\textit{Crataegus pinnatifida}的变种，后者对前者是包含关系，因此我们在命名时不应该包含此变种作为“山楂”的物种基源。\\
            \cmark NMMSN: Crataegus pinnatifida Fruit \cite{nmm-009x}; NMMSN-zh 山楂果实\\
            \xmark Crataegus pinnatifida vel pinnatifida var major Fruit\\
            具有更细化物种基源的NMM可以单列，以明确：\\
            \cmark NMMSN: Crataegus pinnatifida var major Fruit \cite{nmm-009y}; NMMSN-zh: 山里红果实
            \item 具体物种信息不确定的NMM的命名规则。对于具体物种信息不确定的NMM，可以仅用属名进行命名，并在属名后加“unspecified”（NMMGN-zh中添加“未定种”）以提示物种信息尚未被明确。
            \item[Ex. ] 蒲公英属下有数千种物种，均可潜在用作NMM：\\
            NMMSN: Taraxacum unspecified Herb \cite{nmm-01yf}; NMMSN-zh: 蒲公英属未定种全草
            \item 物种学名非合法字符省略规则。如果NMM的物种基源的物种学名含有非合法字符，则省略。
            \item[Ex. ] \cmark NMMSN: Ziziphus jujuba var spinosa Seed \cite{nmm-005x}; NMMSN-zh 酸枣种子\\
            \xmark Ziziphus jujuba var. spinosa Seed
        \end{rulezh}
        
        \item 药用部位命名规则。
        \begin{rulezh}
            \item 语法规则。NMMSN药用部位使用英文单数名词或名词短语进行命名。
            \item[Ex. ] \cmark NMMSN: Panax ginseng Leaf \cite{nmm-001s}; NMMSN-zh: 人参叶\\
            \xmark Panax ginseng Leaves
            \item 首字母大写和连词符规则。为便于阅读，药用部位的首字母应当大写；药用部位由多个词组成时，中间需要使用连词符“-”连接。
            \item[Ex. ] NMMSN: Solidago decurrens Herb \cite{nmm-0020}; NMMSN-zh: 一枝黄花全草
            \item[Ex. ] NMMSN: Ephedra sinica Stem-herbaceous \cite{nmm-0003}; NMMSN-zh: 草麻黄草质茎
            \item 多药用部位命名规则。如果一个NMM同时使用同一物种的多种不同的药用部位入药，药用部位间可以使用“and”（“与”）连接。如果一个NMM可以使用同一物种的多种不同的药用部位入药，且互为可替代关系，则药用部位间可以使用“or”（“或”）连接。药用部位的排序根据拉丁字母顺序。
            \item[Ex. ] NMMSN: Vincetoxicum pycnostelma Rhizome and Root \cite{nmm-01mz}; NMMSN-zh: 徐长卿根茎与根
            \item 细化药用部位规则。当能够明确更细化的药用部位时，我们应当优先使用更细化的药用部位对NMM进行指代。该命名规则有助于我们进一步明确NMM的药用部位。
            \item[Ex. ] “Vincetoxicum pycnostelma Rhizome and Root”可以进一步细化为：\\
            NMMSN: Vincetoxicum pycnostelma Rhizome \cite{nmm-01mx}; NMMSN-zh: 徐长卿根茎\\
            NMMSN: Vincetoxicum pycnostelma Root \cite{nmm-01my}; NMMSN-zh: 徐长卿根
        \end{rulezh}
        
        \item 特殊形容命名规则。
        \begin{rulezh}
            \item 语法规则。一些NMM须具有某种特有的性状特征或经过某些特殊的产地初加工后方可入药，对于这些NMM，可以在命名组件III中构词部分使用英文形容词、形容词短语、名词或同位语进行命名。
            \item 首字母大写和连词符规则。为便于阅读，特殊形容的首字母应当大写；特殊形容由多个词组成时，中间需要使用连词符“-”连接。
            \item[Ex. ] NMMSN: Zingiber officinale Rhizome \cite{nmm-003g}; NMMSN-zh: 姜根茎\\
            NMMSN: Zingiber officinale Rhizome Fresh \cite{nmm-003m}; NMMSN-zh: 鲜姜根茎
            \item[Ex. ] NMMSN: Curcuma wenyujin Rhizome \cite{nmm-000t}; NMMSN-zh: 温郁金根茎\\
            NMMSN: Curcuma wenyujin Rhizome Freshly-sliced \cite{nmm-0015}; NMMSN-zh: 鲜切片温郁金根茎
            \item 道地药材规则。某些NMM需要特别明确道地产区时，可以使用地理名词作为特殊形容。中国产道地NMM的产地名通常使用中华人民共和国省会名称的标准英文名，NMMSN-zh中使用“省会标准缩写+产”作为其中文对应词。
            \item[Ex. ] NMMSN: Fritillaria thunbergii Bulb Zhejiang \cite{nmm-024c}; NMMSN-zh: 浙产浙贝母鳞茎
        \end{rulezh}

        \item 炮制方法命名规则。
        \begin{rulezh}
            \item 语法规则。PNMM必须包含炮制方法。PNMM基于其对应的ANMM进行命名，通过在命名组件IV额外添加炮制方法所对应的英文形容词、形容词短语、同位语或炮制方法的英文缩写词（词组）进行命名。炮制方法的中文对应词必须以“制”结尾。
            \item[Ex. ] ANMM: NMMSN: Artemisia annual Part-aerial \cite{nmm-0001}; NMMSN-zh: 黄花蒿地上部\\
            PNMM: NMMSN: Artemisia annual Part-aerial Segmented \cite{nmm-0002}; NMMSN-zh: 段制黄花蒿地上部
            \item 首字母大写和连词符规则。为便于阅读，炮制方法的首字母应当大写；炮制方法由多个词组成时，中间需要使用连词符“-”连接。
            \item 多重炮制的NMM命名规则。如果是多重炮制的PNMM，需要使用多个炮制方法，其炮制方法间使用“and”连接，居于更晚炮制过程的炮制方法的词序居后（NMMSN-zh的炮制方法由于有“制”作为词尾，因而无须使用逻辑连词；居于更晚炮制过程的炮制方法的词序居前）。炮制方法和炮制顺序相关，因此炮制方法的词序不可随意改变。
            \item[Ex. ] \cmark NMMSN: Ephedra sinica Stem-herbaceous Segmented and Aquafried-honey \cite{nmm-000b}; NMMSN-zh: 蜜炙制段制草麻黄草质茎\\
            \xmark Ephedra sinica Stem-herbaceous Aquafried-honey and Segmented
            \item 中国天然药材（中药材）炮制方法命名规则。中药材的炮制方法以《中国药典·2020年版·四部》“0231 炮制通则” \cite{chp2020vol4}为基础。炮制方法命名时，如果能明确炮制细类，则优先采用炮制细类进行命名；在不能明确炮制细类时，可以采用炮制大类进行命名。
            \item[Ex. ] NMMSN: Crataegus pinnatifida Fruit Cleaned and Stirfried-golden \cite{nmm-00a1}; NMMSN-zh: 炒黄制净制山楂果实\\
            NMMSN: Crataegus pinnatifida Fruit Cleaned and Stirfried-charred \cite{nmm-00a3}; NMMSN-zh: 炒焦制净制山楂果实
            \item 中药材炮制分类规则。中药材炮制（natural medicinal processing）分3大类：净制 (processing by cleaning)、切制 (processing by cutting)、备制 (processing by preparing)。经过切制的中药材默认已经经过净制。凡需要备制的，其必须首先经过净制或切制。
            \item[Ex. ] \cmark NMMSN: Ephedra sinica Stem-herbaceous Segmented and Aquafried-honey \cite{nmm-000b}; NMMSN-zh: 蜜炙制段制草麻黄草质茎\\
            \xmark Ephedra sinica Stem-herbaceous Aquafried-honey
            \item[Ex. ] \cmark NMMSN: Zingiber officinale Rhizome Cleaned and Stirfried-sand \cite{nmm-003l}; NMMSN-zh: 砂炒制净制姜根茎\\
            \xmark Zingiber officinale Rhizome Stirfried-sand
        \end{rulezh}

        \item 非物种类或物种基源难以确定的NMM的命名规则。若待命名NMM为非物种类或物种基源难以明确，其NMMSN可采用英语习称进行命名。
        \item[Ex. ] NMMSN: Talc \cite{nmm-01xx}; NMMSN-zh: 滑石\\
        NMMSN: Talc Pulverized \cite{nmm-01xz}; NMMSN-zh: 粉制滑石
        
    \end{rulezh}

    \item NMMGN规则。
    \begin{rulezh}
        \item 简洁规则。NMMSN由于命名时需要明确NMM的物种基源、药用部位、特殊描述、炮制方法等信息，因此NMMSN通常较长。为了便于日常使用和临床处方时的便利，每个NMM也均有一个对应的较简短的NMMGN。
        \item 中药材NMMGN-zh的惯常规则。中药材的NMMGN-zh优先采用已经惯用的中药材名。如果中药材出自ChP，ChP收录的中药材的中文名通常即为NMMGN-zh。对于细化物种基源后的单物种基源中药材，为了防止命名冲突，个别中药材的NMMGN-zh开头添加“单”/“独”以区分。
        \item 中药材NMMGN的拼音名优先规则。为了使得中药材的NMMGN和NMMGN-zh呼应，我们在命名中药材的NMMGN时，优先使用中药材NMMGN-zh的拼音名。拼音名不使用声调和空格，使用连词符“-”连接不同汉字的拼音，首字母大写，并使用“v”代替“ü”。
        \item[Ex. ] NMMSN: Erycibe obtusifolia Stem \cite{nmm-0022}; NMMSN-zh: 丁公藤茎; NMMGN: Dan-ding-gong-teng; NMMGN-zh: 单丁公藤（为和“丁工藤”区分，添加“单”字。）\\
        NMMSN: Erycibe obtusifolia vel schmidtii Stem \cite{nmm-0024}; NMMSN-zh: 丁公藤或光叶丁公藤茎; NMMGN: Ding-gong-teng; NMMGN-zh: 丁公藤
        \item[Ex. ] NMMSN: Ligustrum lucidum Fruit \cite{nmm-00d7}; NMMSN-zh: 女贞果实; NMMGN: Nv-zhen-zi; NMMGN-zh: 女贞子
        \item 最短NMMGN-zh规则。NMMGN-zh必须使用2个及以上的汉字进行命名。
        \item[Ex. ] NMMSN: Prunus mume Fruit \cite{nmm-008w}; NMMSN-zh: 梅果实\\
        NMMGN-zh: \cmark 乌梅 \xmark 梅
        \item 先到先得规则。由于NMMGN较短，在其命名时，难免遇到命名冲突的情况。在这种情况下，我们需要遵循先到的NMMGN优先，对于后到的NMMGN，其须加一些额外信息以区分。
        \item[Ex. ] 假设我们已经对以下NMM进行命名：\\
        NMMSN: Ephedra sinica Stem-herbaceous \cite{nmm-0003}; NMMSN-zh: 草麻黄草质茎; NMMGN: Cao-ma-huang; NMMGN-zh: 草麻黄\\
        在上述案例中该NMM的NMMGN-zh并未提及其药用部位信息，因为其是优先命名的，所以省略。但假设我们要进一步将\textit{Ephedra sinica}的根用药，其NMMGN和NMMGN-zh就要适当做出区分：\\
        NMMSN: Ephedra sinica Root \cite{nmm-000g}; NMMSN-zh: 草麻黄根; NMMGN: Cao-ma-huang-gen; NMMGN-zh: 草麻黄根
    \end{rulezh}
    
    \item NMM ID规则。每种NMM被赋予唯一的NMM ID。NMM ID的编码规则为：\lstinline{NMM-XXXX}，其中\lstinline{XXXX}为4位36进制的数字（即0-9，A-Z），从\lstinline{0001}开始递增，止于\lstinline{ZZZZ}，至多可编码$36^4 - 1 = 1,679,615$种NMM。NMM ID大小写不敏感，但在书写时为了方便阅读通常使用全大写。
    \item[Ex. ] NMM-ID: NMM-0001 \cite{nmm-0001}\\
    NMMSN: Artemisia annua Part-aerial\\
    NMMSN-zh: 黄花蒿地上部\\
    NMMGN: Qing-hao\\
    NMMGN-zh: 青蒿
    \item[Ex. ] NMM-ID: NMM-0002 \cite{nmm-0002}\\
    NMMSN: Artemisia annua Part-aerial Segmented\\
    NMMSN-zh: 段制黄花蒿地上部\\
    NMMGN: Qing-hao-duan\\
    NMMGN-zh: 青蒿段

    \item 标准指代规则。
    \begin{rulezh}
        \item 首次完整出现规则。为了保证科学文本中的NMM名称和其实际指代的NMM准确对应，推荐在每个独立文本（如一篇论文、百科等）中首次出现某NMM时，以“NMMSN (NMM-ID, NMMGN)”格式给出NMM的两名一ID。在后续文本中NMM第二次出现时，允许只使用NMMSN、NMMGN或NMM ID进行指代。
        \item[Ex. ] Artemisia annua Part-aerial (NMM-0001, Qing-hao) is a commonly used Chinese natural medicinal material. Artemisia annua Part-aerial has the effect of treating malaria.
        \item[Ex. ] Artemisia annua Part-aerial (NMM-0001, Qing-hao) is a commonly used Chinese natural medicinal material. Qing-hao has the effect of treating malaria.
        \item[Ex. ] Artemisia annua Part-aerial (NMM-0001, Qing-hao) is a commonly used Chinese natural medicinal material. NMM-0001 has the effect of treating malaria.
        \item[Ex. ] 黄花蒿地上部（NMM-0001，青蒿）是一种常用的中药材。黄花蒿地上部可用于治疗疟疾。
        \item[Ex. ] 黄花蒿地上部（NMM-0001，青蒿）是一种常用的中药材。青蒿可用于治疗疟疾。
        \item[Ex. ] 黄花蒿地上部（NMM-0001，青蒿）是一种常用的中药材。NMM-0001可用于治疗疟疾。
        \item 附录完整信息规则。如果某研究中涉及大量NMM，在正文文本中一一给出所有的NMMSN、NMMGN和NMM ID可能导致正文文本过长。这种情况下，可以在正文中仅使用NMMSN或NMMGN或NMM ID中的一种来指代NMM；但在附录中，需要给出“NMM ID -- NMMSN -- NMMGN”的完整列表。
    \end{rulezh}
\end{rulezh}

%% file: chapter/snnmma.tex
\section{Supplementary method: SNNMM Algorithm (SNNMMA)}
\label{appendix:snnmma}

The algorithm and code of SNNMMA has been open-sourced and is available on GitHub (\url{https://github.com/shennong-program/shennongname}). The relevant Python package has been published and released on PyPI (\url{https://pypi.org/project/shennongname}).

This supplementary method primarily elucidates the fundamental principles of the SNNMMA, emphasizing its data input and output aspects. For detailed code implementation, readers are referred to the aforementioned repository.

The input to SNNMMA is a JSON Object. To elucidate, we illustrate with a common natural medicinal material, Mi-ma-huang (蜜麻黄) \boldcolor{(\cref{listing:snnmma-input})}.

\begin{listing}[H]
\begin{sjson}
{
    "nmm_type": "processed",
    "species_origins": [["Ephedra sinica", "草麻黄"], "or", ["Ephedra intermedia", "中麻黄"], "or", ["Ephedra equisetina", "木贼麻黄"]],
    "medicinal_parts": [["stem herbaceous", "草质茎"]],
    "special_descriptions": [],
    "processing_methods": [["segmented", "段制"], "and", ["aquafried honey", "蜜炙制"]]
}
\end{sjson}
\caption{
\textbf{Example of data structure for SNNMMA input.}
}
\label{listing:snnmma-input}
\end{listing}

Within the aforementioned input data structure, users of SNNMMA are required to provide information pertaining to the NMM type along with the associated details for the four kinds of name elements. It is noteworthy that these four types of name elements are collectively stored utilizing a data structure denominated as \lstinline{NmmsnNeData}.

The basic structure is as follows, exemplified by \lstinline{species_origins} \boldcolor{(\cref{listing:snnmma-nmmnedata})}:

\begin{listing}[H]
\begin{sjson}
[
    ["Ephedra sinica", "草麻黄"], 
    "or",
    ["Ephedra intermedia", "中麻黄"], 
    "or", 
    ["Ephedra equisetina", "木贼麻黄"]
]
\end{sjson}
\caption{
\textbf{Example of data structure for NmmsnNeData in SNNMMA.} \\
}
\label{listing:snnmma-nmmnedata}
\end{listing}

The list encompassed in \lstinline{NmmsnNeData} permits the incorporation of multiple name element pairs, each with a data substructure: \lstinline{["name element in English or Latin", "name element in Chinese"]}, which can be interconnected by the logical operator strings \lstinline{"or"} or \lstinline{"and"}.

Subsequently, when \lstinline{NmmsnNeData} is conveyed to SNNMMA, the algorithm autonomously executes a series of processes for each name element type of \lstinline{NmmsnNeData}. This includes string data verification, deduplication, sorting, character transformation, and more. Ultimately, the algorithm calculates and derives the NMM Systematic Name (NMMSN) and NMM Systematic Chinese Name (NMMSN-zh), presenting the results as another JSON Object \boldcolor{(\cref{listing:snnmma-output})}.

\begin{listing}[H]
\begin{sjson}
{
    "success": true,
    "error_msg": "Pipe: construct_nmmsn_spe_ori. Status: warning. Reason: Multiple species origins detected.", 
    "error_msg_en_zh": {
        "en": "Multiple species origins detected.",
        "zh": "检测到多个物种基源。"
    },
    "nmmsn": {
        "nmmsn": "Ephedra equisetina vel intermedia vel sinica Stem-herbaceous Segmented and Aquafried-honey",
        "nmmsn_zh": {
            "zh": "蜜炙制段制木贼麻黄或中麻黄或草麻黄草质茎",
            "pinyin": "mì zhì zhì duàn zhì mù zéi má huáng huò zhōng má huáng huò cǎo má huáng cǎo zhì jīng"
        },
        "nmmsn_name_element": { 
            "nmm_type": "processed",
            "species_origins": [["Ephedra equisetina", "木贼麻黄"], "or", ["Ephedra intermedia", "中麻黄"], "or", ["Ephedra sinica", "草麻黄"]],
            "medicinal_parts": [["stem herbaceous", "草质茎"]],
            "special_descriptions": [],
            "processing_methods": [["segmented", "段制"], "and", ["aquafried honey", "蜜炙制"]]
        },
        "nmmsn_seq": [["Ephedra equisetina vel intermedia vel sinica", "木贼麻黄或中麻黄或草麻黄"], ["Stem-herbaceous", "草质茎"], ["", ""], ["Segmented and Aquafried-honey", "蜜炙制段制"]] 
    }
}
\end{sjson}
\caption{
\textbf{Example of data structure for SNNMMA output (NMMSN construction successful).}
}
\label{listing:snnmma-output}
\end{listing}

The output of the SNNMMA displays the following characteristics: 

Once user data is successfully processed by SNNMMA to construct an NMMSN, the value of \lstinline{success} will be set to \lstinline{true}. Moreover, the resultant information post-NMMSN construction by SNNMMA will be stored under the \lstinline{nmmsn} key.

The specific meanings of each hierarchical key in the SNNMMA output are described as follows:

\begin{itemize}

\item \lstinline{error_msg}: In the SNNMM framework, there exist certain valid yet non-preferred rules. For instance, using multiple species origins for systematic naming of NMM is not recommended. The SNNMMA can automatically detect such anomalies during NMMSN construction. Consequently, there might be instances where the NMMSN is successfully constructed, yet the \lstinline{error_msg} remains populated, recording any issues encountered during the process. These error messages adhere to a standardized format: \lstinline{Pipe: xxx. Status: xxx. Reason: xxx.} and are stored within \lstinline{error_msg}.

\item \lstinline{error_msg_en_zh}: To enhance the user experience for both English and Chinese users, error messages in SNNMMA have been localized. Information pertaining to the \lstinline{Reason} in \lstinline{error_msg} is processed and stored in both English and Chinese within \lstinline{error_msg_en_zh.en} and \lstinline{error_msg_en_zh.zh} respectively. This ensures that even users with programming or language barriers can clearly understand any issues encountered during NMMSN construction by SNNMMA.

\item \lstinline{nmmsn}: This key houses all information directly related to NMMSN.

\item \lstinline{nmmsn.nmmsn}: Represents the successfully constructed NMM Systematic Name.

\item \lstinline{nmmsn.nmmsn_zh}: \lstinline{nmmsn.nmmsn_zh.zh} denotes the successfully constructed NMM Systematic Chinese Name, while \lstinline{nmmsn.nmmsn_zh.pinyin} represents the corresponding pinyin.

\item \lstinline{nmmsn.nmmsn_name_element}: The data structure of this key mirrors the structure of input data \boldcolor{(\cref{listing:snnmma-input})}. However, the order of elements within the \lstinline{NmmsnNeData} data structure might be adjusted or reordered based on the SNNMM rules. For example, the order in \lstinline{species_origins} might change from \mintinline[breaklines]{json}{[["Ephedra sinica", "草麻黄"], "or", ["Ephedra intermedia", "中麻黄"], "or", ["Ephedra equisetina", "木贼麻黄"]]} to \mintinline[breaklines]{json}{[["Ephedra equisetina", "木贼麻黄"], "or", ["Ephedra intermedia", "中麻黄"], "or", ["Ephedra sinica", "草麻黄"]]} due to the alphabetical ordering being \lstinline{e -> i -> s} among the three species origins.

\item \lstinline{nmmsn.nmmsn_seq}: Given that NMMSN comprises four name elements, this key stores the NMMSN corresponding to each name element. This sequenced NMMSN, within ShennongName, can be utilized to distinctively display each name element in a unique color, enhancing user-friendliness.

\end{itemize}

If SNNMMA encounters issues during the NMMSN construction process and fails, the output from SNNMMA will be strikingly similar to that shown in \boldcolor{\cref{listing:snnmma-output}}. However, the value of \lstinline{success} will be set to \lstinline{false}, and the resulting JSON Object will not include the \lstinline{nmmsn} key and its corresponding value \boldcolor{(\cref{listing:snnmma-output-failed})}.

\begin{listing}[H]
\begin{sjson}
{
    "success": false,
    "error_msg": "...", 
    "error_msg_en_zh": {
        "en": "...",
        "zh": "..."
    }
}
\end{sjson}
\caption{
\textbf{Example of data structure for SNNMMA output (NMMSN construction failed).}
}
\label{listing:snnmma-output-failed}
\end{listing}

%% file: chapter/mlmd.tex
\section{Supplementary method: Multilingual Markdown (MLMD)}
\label{appendix:mlmd}

The source code for the MLMD parser has been open-sourced and can be accessed on GitHub: \url{https://github.com/shennong-program/mlmd}.

The associated TypeScript package has been released on npm: \url{https://www.npmjs.com/package/mlmd}

This Supplementary method primarily introduces the core design principles, essential syntax, and usage methods of MLMD. Please refer to the repository above for the detailed syntax and parsing code of MLMD.

\subsection{Core design principles}

Multilingual Markdown (MLMD) is a newly designed lightweight markup language explicitly tailored for managing multilingual text. The syntax of MLMD integrates the strengths of Markdown \cite{CommonMark} while being specially designed for multilingual parallel corpora.

The primary objective of MLMD is to allow users to write and manage multilingual parallel text in an easy-to-read and easy-to-write plain text format.

Thanks to the MLMD syntax that can manage multilingual content in a single document, it can be widely applied in the following scenarios:

\begin{itemize}
    \item Multilingual text writing, content creation
    \item Unified/Structured management, storage of multilingual texts
    \item Multilingual display of multilingual texts
    \item Cross-language text annotation, entity annotation
    \item Cross-language translation, proofreading
    \item Cross-language machine translation
    \item Text analysis/mining of multilingual texts
    \item Natural language processing of multilingual texts
    \item ...
\end{itemize}

\subsection{Writing, storage, and file extension}

The content of MLMD is in plain text, making it possible to compose using any plain text editor.

When saving MLMD content as an independent file, the \lstinline{.mlmd} extension should be used. For instance, if we save MLMD text in a file named \lstinline{abc.mlmd}, then \lstinline{abc} represents the filename and \lstinline{.mlmd} is the standard extension for MLMD.

Since MLMD text is plain text, an independent MLMD text can technically be embedded/stored as a single string within other file formats or databases such as JSON, SQL, MongoDB, etc.

For example, consider the following MLMD text \boldcolor{(\cref{listing:mlmd-hello})}:

\begin{listing}[H]
\begin{mlmd}
{{langs|zh|en}}

你好，世界！
Hello, world!
\end{mlmd}
\caption{
\textbf{MLMD example: Hello, World!}
}
\label{listing:mlmd-hello}
\end{listing}

It can be stored as a single string in a JSON structure \boldcolor{(\cref{listing:mlmd-json})}:

\begin{listing}[H]
\begin{sjson}
{
    "mlmd_str": "{{langs|zh|en}}\n\n你好，世界！\nHello, world!"
}
\end{sjson}
\caption{
\textbf{MLMD stored as a string inside a JSON structure.}
}
\label{listing:mlmd-json}
\end{listing}

\subsection{Essential syntax}

\subsubsection{MLMD language header}

Since MLMD can be used to store parallel corpora of any number of languages, it is required to explicitly define the languages of the text stored in MLMD in the first line of every MLMD file, using the format \lstinline!{{langs|<language_code_1>|<language_code_2>|...}}!. Following the syntax of MLMD, MLMD documents can manage parallel corpora of any number of languages. Thus, non-repeating language codes can be used in the MLMD language header.

It is noteworthy that the first language code in the MLMD language header corresponds to the primary language of that particular MLMD document.

For instance, the language header of a bilingual MLMD document in Chinese and English is as follows \boldcolor{(\cref{listing:mlmd-head-zh-en})}:

\begin{listing}[H]
\begin{mlmd}
{{langs|zh|en}}
\end{mlmd}
\caption{
\textbf{MLMD language header example: Chinese-English.}
}
\label{listing:mlmd-head-zh-en}
\end{listing}

Whereas the language header of a trilingual MLMD document in Chinese, English, and Latin is as follows \boldcolor{(\cref{listing:mlmd-head-zh-en-la})}:

\begin{listing}[H]
\begin{mlmd}
{{langs|zh|en|la}}
\end{mlmd}
\caption{
\textbf{MLMD language header example: Chinese-English-Latin.}
}
\label{listing:mlmd-head-zh-en-la}
\end{listing}

\subsubsection{Multilingual parallel paragraphs}

In MLMD, multilingual parallel corpora are managed at the paragraph level. Each multilingual paragraph is treated as a Block within MLMD. Within each Block, the paragraphs of text in different languages are arranged in the order of language codes and are separated by a line break (when stored as strings, the newline character \lstinline{\n} is used). Different Blocks are separated by a blank line (when stored as strings, two newline characters \lstinline{\n\n} are used). For examples \boldcolor{(\cref{listing:mlmd-zh-en,listing:mlmd-zh-en-la})}:

\begin{listing}[H]
\begin{mlmd}
{{langs|zh|en}}

这是第1段中文。
This is the 1st paragraph in English.

这是第2段中文。
This is the 2nd paragraph in English.
\end{mlmd}
\caption{
\textbf{MLMD multilingual parallel paragraphs example: Chinese-English.}
}
\label{listing:mlmd-zh-en}
\end{listing}

\begin{listing}[H]
\begin{mlmd}
{{langs|zh|en|la}}

这是第1段中文。
This is the 1st paragraph in English.
Hic est paragraphus Latinus primus.

这是第2段中文。
This is the 2nd paragraph in English.
Hic est paragraphus Latinus secundus.
\end{mlmd}
\caption{
\textbf{MLMD multilingual parallel paragraphs example: Chinese-English-Latin.}
}
\label{listing:mlmd-zh-en-la}
\end{listing}

Incidentally, the concept of a Block is one of the cores of MLMD, as when we wish to store data of different structures (such as links, images, etc.) within MLMD, they are essentially treated as individual Blocks. Blocks' management approach of Blocks allows MLMD to be conveniently parsed. The Blocks corresponding to multilingual parallel paragraphs are abbreviated as ``Multi".

\subsubsection{Language-invariant paragraphs}

In multilingual texts, certain text paragraphs remain the same across all languages. We refer to such text as language-invariant paragraphs. When a Block in our saved MLMD document contains only one paragraph of text, it is automatically regarded as a language-invariant text. This type of Block is abbreviated as ``Mono". For example \boldcolor{(\cref{listing:mlmd-mono})}:

\begin{listing}[H]
\begin{mlmd}
{{langs|zh|en}}

这是第1个Block (Multi)中的中文段落。
This is an English paragraph in the 1st Block (Multi).

This is a language-invariant paragraph in English in the 2nd Block (Mono).

这是第3个Block (Mono)中的一段中文的跨语言不变段落。

这是第4个Block (Multi)中的中文段落。
This is an English paragraph in the 4th Block (Multi).
\end{mlmd}
\caption{
\textbf{MLMD language-invariant paragraphs example.}
}
\label{listing:mlmd-mono}
\end{listing}

\subsubsection{Emphasis}

In MLMD, text can be emphasized by wrapping with \lstinline{**} for bold emphasis and \lstinline{*} for italic emphasis. For example \boldcolor{(\cref{listing:mlmd-em})}:

\begin{listing}[H]
\begin{mlmd}
{{langs|zh|en}}

第1个Block (Multi)中的中文**加粗**段落。
A **bold** paragraph in English within the 1st Block (Multi).

A **bold** and *italic* language-invariant paragraph in English in the 2nd Block (Mono).

第3个Block (Mono)中的一段中文的**加粗**和*斜体*的跨语言不变段落。

第4个Block (Multi)中的中文*斜体*段落。
An *italic* paragraph in English within the 4th Block (Multi).
\end{mlmd}
\caption{
\textbf{MLMD emphasis example.}
}
\label{listing:mlmd-em}
\end{listing}

\subsubsection{Headings}

MLMD supports up to six levels of headings. Headings are denoted using the \lstinline{#} symbol. By prefixing a paragraph with 1-6 of the \lstinline{#} symbols, you can designate the heading levels 1-6, respectively. To separate multilingual headings, use the \lstinline{|} symbol. If there is no \lstinline{|}, the heading content is treated as Mono. For example \boldcolor{(\cref{listing:mlmd-heading})}:

\begin{listing}[H]
\begin{mlmd}
{{langs|zh|en}}

# 一级标题 | Heading Level-1

## 二级标题 | Heading Level-2

### 三级标题 | Heading Level-3

#### 四级标题 | Heading Level-4

##### 五级标题 | Heading Level-5

###### 六级标题 | Heading Level-6

# 一级标题

## Heading Level-2
\end{mlmd}
\caption{
\textbf{MLMD headings example.}
}
\label{listing:mlmd-heading}
\end{listing}

\subsubsection{Coreference annotation}

The central linguistic principle of this syntax is ``coreference consistency". It recognizes that different terms or phrases can refer to the same inherent concept or entity in multilingual or monolingual texts. This capacity to discern that, despite variations in form, the essence of the reference remains unchanged underpins this principle. In MLMD, terms with this shared referential quality are described as having ``coreference consistency".

Consider this illustrative example \boldcolor{(\cref{listing:mlmd-no-coref})}:

\begin{listing}[H]
\begin{mlmd}
{{langs|zh|en}}

神农被中国人认为是医药学的始祖。
Shennong is considered the progenitor of medicine and pharmacy by the Chinese.

炎帝被中国人认为是医药学的始祖。
Yan Emperor is considered the progenitor of medicine and pharmacy by the Chinese.
\end{mlmd}
\caption{
\textbf{MLMD coreference consistency example (not annotated).}
}
\label{listing:mlmd-no-coref}
\end{listing}

To those familiar with the historical background, irrespective of the terms ``神农", ``炎帝", ``Shennong", or ``Yan Emperor" being used, they understand these allude to the same historical figure, specifically ``神农" (Shennong). However, these could appear as four separate concepts for those without this knowledge. To convey this textual coreference consistency within the content, MLMD introduces a dedicated coreference annotation syntax. This facilitates easier textual annotations of coreferential relationships and enhances machine translation and AI natural language processing capabilities in understanding coreference consistency. The specific syntax is as follows:

The double bracket notation \lstinline{[[...]]} is utilized for coreference annotation. No spaces must exist between \lstinline{[[} and \lstinline{]]}.

The primary coreference annotations are of two types:

\begin{enumerate}
    \item When the term's expression aligns with its reference: \lstinline{[[term's reference]]}
    \item When there's a divergence between the term's expression and its reference: \lstinline{[[term's reference|term's expression]]}
\end{enumerate}

By employing this approach, the sentences mentioned above can be annotated as \boldcolor{(\cref{listing:mlmd-coref})}:

\begin{listing}[H]
\begin{mlmd}
{{langs|zh|en}}

[[神农]]被中国人认为是医药学的始祖。
[[神农|Shennong]] is considered the progenitor of medicine and pharmacy by the Chinese.

[[神农|炎帝]]被中国人认为是医药学的始祖。
[[神农|Yan Emperor]] is considered the progenitor of medicine and pharmacy by the Chinese.
\end{mlmd}
\caption{
\textbf{MLMD coreference consistency example (annotated).}
}
\label{listing:mlmd-coref}
\end{listing}

In this framework, ``神农" stands as the Primary Term denoting the concept, while ``炎帝," ``Shennong," and ``Yan Emperor" are its coreferential terms. Hence, with the above coreference annotations, even those without the specific historical context can infer that these sentences all point to the same person, specifically ``神农".

\subsubsection{Entity annotation}

In natural language texts, certain terms often represent specific entities. We can achieve a unified annotation by utilizing the coreference annotation syntax to corefer a term to its corresponding entity. In practice, terms can be coreferred to their corresponding entity ID.

For instance, both ``青蒿" and ``Qing-hao" correspond to the NMM ID ``NMM-0001". Hence, we can annotate the entity as follows \boldcolor{(\cref{listing:mlmd-entity})}:

\begin{listing}[H]
\begin{mlmd}
{{langs|zh|en}}

[[NMM-0001|青蒿]]是一种天然药材。
[[NMM-0001|Qing-hao]] is a kind of Natural Medicinal Material.
\end{mlmd}
\caption{
\textbf{MLMD entity annotation example.}
}
\label{listing:mlmd-entity}
\end{listing}

\subsubsection{Comments}

MLMD uses \lstinline{<!-- ... -->} for comments. For example \boldcolor{(\cref{listing:mlmd-comment})}:

\begin{listing}[H]
\begin{mlmd}
{{langs|zh|en}}

第1个Block (Multi)中的中文**加粗**段落。 <!-- 单行注释 -->
A **bold** paragraph in English within the 1st Block (Multi).  <!-- Single-line comment -->

A **bold** and *italic* language-invariant paragraph in English in the 2nd Block (Mono). <!-- Single-line comment -->

<!-- 多行注释
Multi-line comment
-->

第3个Block (Mono)中的一段中文的**加粗**和*斜体*的跨语言不变段落。 <!-- 单行注释 -->

第4个Block (Multi)中的中文*斜体*段落。
An *italic* paragraph in English within the 4th Block (Multi).
\end{mlmd}
\caption{
\textbf{MLMD comment example.}
}
\label{listing:mlmd-comment}
\end{listing}

\subsubsection{Templates}

For texts with specific functionalities, MLMD employs the template syntax \lstinline!{{...}}! for annotation. No spaces are allowed between the double curly brackets \lstinline!{{! and \lstinline!}}!. For instance, the MLMD language header is a type of specialized template.

MLMD can be extended to cater to various unique template functional requirements. Detailed template syntax can be found in the MLMD repository. Here, we emphasize the citation template directly related to this paper.

\subsubsection{Citations}

Users can conveniently add citations to the MLMD text using the citation template: \lstinline!{{ref|@<ref_id>]}}!. This makes MLMD suitable for rigorous academic texts. For detailed information related to the \lstinline{ref_id}, one can store it in BibTeX format using \lstinline!{{ref-citation|bibtex|<ref_citation_info>}}!. The \lstinline{ref_id} must match with the \lstinline{ref_id} in the BibTeX within \lstinline{ref_citation_info}. For example \boldcolor{(\cref{listing:mlmd-cite})}:

\begin{listing}[H]
\begin{mlmd}
{{langs|zh|en}}

AI介导的五阶段科学革命，这一理论在《AI成为主脑科学家》这篇论文中被首次提出{{ref|@ai_masterbrain}}。
The theory of the "Five Stages of AI-involved Scientific Revolution" was first introduced in the paper titled "AI Becomes a Masterbrain Scientist" {{ref|@ai_masterbrain}}.

{{ref-citation|bibtex|
@article{ai_masterbrain,
  title={AI becomes a masterbrain scientist},
  author={YANG, Zijie and WANG, Yukai and ZHANG, Lijing},
  journal={bioRxiv},
  pages={2023--04},
  year={2023},
  publisher={Cold Spring Harbor Laboratory}
}
}}
\end{mlmd}
\caption{
\textbf{MLMD citation example.}
}
\label{listing:mlmd-cite}
\end{listing}

On the ShennongAlpha website, all references are stored in the ShennongKB reference collection. Each reference is assigned a globally unique ID in the format \lstinline{sna-ref-xxx}. Hence, in ShennongAlpha, one can employ a special reference annotation like \lstinline!{{ref|[[sna-ref-1]]}}!. Here, \lstinline{[[sna-ref-1]]} serves as the \lstinline{@<ref_id>}. Since \lstinline{sna-ref-1} is encapsulated within \lstinline![[]]!, it signifies that \lstinline{[[sna-ref-1]]} inherently is a coreference related to the citation. Therefore, ShennongAlpha parses MLMD automatically fetches the reference information about \lstinline{sna-ref-1} from ShennongKB. In this context, there's no further need to supplement the specific reference details through \lstinline!{{ref-citation|bibtex}}!.

\subsection{Machine Translation Compatibility}

MLMD natively supports a novel machine translation approach introduced in this paper: Neural Machine Translation based on Coreference Primary Term (NMT-CPT).

Using coreference annotation to denote the relationships between standard Primary Terms and their translations, NMT-CPT offers users a streamlined and interactive presentation of translation results through the ShennongTranslate user interface. This approach significantly enhances the interpretability of machine translations.

\subsection{HTML Parsing}

Through a parsing flow: MLMD $\rightarrow$ Abstract Syntax Tree (AST) \cite{mdast} $\rightarrow$ HTML, MLMD can be rendered into HTML with rich text formatting, enhancing the user experience while reading its content, ShennongAlpha natively supports HTML rendering of MLMD, offering the bilingual content presentation in four modes: ``Chinese-English", ``English-Chinese," ``Chinese", and ``English" \boldcolor{(\cref{sfig:modes})}.

%% file: chapter/cgs.tex
\section{Supplementary method: Coreference-based Graph Search (CGS)}
\label{appendix:cgs}

\subsection{Foundational CGS}

To elucidate the process of CGS, we illustrate it with a straightforward case:

Consider the set of nodes:
\[ A, B, C, D, E, F \]

The coreference relationships are given as follows:

\begin{align*}
A &\rightarrow B \\
B &\rightarrow \bm{C} \\
D &\rightarrow B \\
E &\rightarrow \bm{F} \\
\end{align*}

Here, \( \bm{C}, \bm{F} \) are manually designated as Primary Terms.

The ``Coreference Primary Term Graph (CPTG)" can be constructed from the above nodes and directed edges. This CPTG must satisfy the following conditions:

\begin{enumerate}
    \item It should be a Directed Acyclic Graph (DAG).
    \item The out-degree of any Primary Term in the graph should be zero.
    \item The out-degree of any node should be less than or equal to one, indicating that no branching nodes are present.
\end{enumerate}

The graph constructed from the given example is illustrated below:

\begin{center}
\begin{tikzpicture}[>=Stealth]
    \node[draw, circle, font=\itshape] (A) at (0,1) {A};
    \node[draw, circle, font=\itshape] (D) at (0,-1) {D};
    \node[draw, circle, font=\itshape] (B) at (2,0) {B};
    \node[draw, rectangle, font={\itshape\bfseries}] (C) at (4,0) {C};

    \node[draw, circle, font=\itshape] (E) at (0,-2) {E};
    \node[draw, rectangle, font={\itshape\bfseries}] (F) at (2,-2) {F};

    \draw[->] (A) -- (B);
    \draw[->] (D) -- (B);
    \draw[->] (B) -- (C);
    \draw[->] (E) -- (F);
\end{tikzpicture}
\end{center}

Once a CPTG is constructed, preliminary graph computations can be conducted to optimize queries to identify the ultimate Primary Term corresponding to any term node.

The results of these computations are stored in a newly created dictionary. Notably, within this dictionary, relationships of directed edges from a Primary Term to itself (e.g., \( \bm{C} \rightarrow \bm{C} \)) are also included. This feature allows the queried term to be recognized as a Primary Term.

\begin{align*}
A &\rightarrow \bm{C} \\
B &\rightarrow \bm{C} \\
D &\rightarrow \bm{C} \\
E &\rightarrow \bm{F} \\
\bm{C} &\rightarrow \bm{C} \\
\bm{F} &\rightarrow \bm{F} \\
\end{align*}

Thus, by searching this dictionary, it can be swiftly determined whether a corresponding Primary Term exists for any given term, including Primary Terms, by verifying its presence in the dictionary. If found, the value corresponding to this term will be its associated Primary Term.

The pseudocode for the Foundational CGS Algorithm is as follows:

\begin{algorithm}[H]
\caption{Foundational CGS Algorithm}
\begin{algorithmic}[1]
\State \textbf{Input:} Set of nodes $N$, Coreference relationships $R$
\State \textbf{Output:} Dictionary $D$ storing the ultimate Primary Terms

\State Initialize the dictionary $D$.
\State Construct the Coreference Primary Term Graph (CPTG) $G$ from the given $N$ and $R$.

\For{each node $n$ in $G$}
    \If{$n$ is not a Primary Term}
        \State Find the ultimate Primary Term $p$ for $n$.
        \If{$p$ exists}
            \State Add the mapping $n \rightarrow p$ to $D$.
        \EndIf
    \Else
        \State Add the self-referencing mapping $n \rightarrow n$ to $D$.
    \EndIf
\EndFor

\State \textbf{For searching:}
\State Given any term $t$, query $D$ to ascertain whether a corresponding Primary Term exists.
\If{$t$ exists in $D$}
    \State Retrieve the corresponding Primary Term $p$ from $D$.
\EndIf

\end{algorithmic}
\end{algorithm}

\subsection{Weighted CGS}

During the construction of the CPTG, directed edges are typically extracted automatically from databases. Consequently, we often cannot satisfy the third requirement of CPTG (i.e., no branching nodes). We can assign a weight to each directed edge based on the Foundational CGS and conduct a weighted graph search in such instances.

Let's illustrate this with a simple example:

Consider the set of nodes:
\[ A, B, C, D, E \]

The coreference relationships are given as follows:

\begin{align*}
A &\xrightarrow{w_{AB}=1} B \\
B &\xrightarrow{w_{BC}=2} \bm{C} \\
D &\xrightarrow{w_{AB}=1} B \\
B &\xrightarrow{w_{AB}=1} \bm{E} \\
\end{align*}

Here, \( \bm{C} \) and \( \bm{E} \) are manually designated as Primary Terms. Each coreference relationship is assigned a weight to signify the importance of that relationship.

A CPTG can be constructed from the above nodes and directed edges. This CPTG meets the first two requirements of a CPTG in Foundational CGS but does not satisfy the third requirement.

The graph constructed from the given example would be as follows:

\begin{center}
\begin{tikzpicture}[>=Stealth]
    \node[draw, circle, font=\itshape] (A) at (0,1) {A};
    \node[draw, circle, font=\itshape] (D) at (0,-1) {D};
    \node[draw, circle, font=\itshape] (B) at (2,0) {B};
    \node[draw, rectangle, font={\itshape\bfseries}] (C) at (4,1) {C};
    \node[draw, rectangle, font={\itshape\bfseries}] (E) at (4,-1) {E};

    \draw[->] (A) -- node[above] {1} (B);
    \draw[->] (D) -- node[above] {1} (B);
    \draw[->] (B) -- node[above] {2} (C);
    \draw[->] (B) -- node[above] {1} (E);
\end{tikzpicture}
\end{center}

In this CPTG, we observe an anomalous branching node \(B\), which points to both \(\bm{C}\) and \(\bm{E}\) simultaneously. Therefore, under these circumstances, when conducting preliminary graph computations to search for the ultimate Primary Term of all nodes, we proceed as follows: For each given node, we acquire all its downstream directed edges. By evaluating the weights of these edges, we select the downstream node corresponding to the edge with the highest weight as the next node. We repeat this process until any Primary Term is reached. 

For instance, when searching for the ultimate Primary Term corresponding to node \(A\), we navigate from \(A\) to its sole downstream node \(B\). However, \(B\) actually has two downstream nodes, \(\bm{C}\) and \(\bm{E}\). At this juncture, since the weight of \(B \rightarrow \bm{C}\) (2) is greater than that of \(B \rightarrow \bm{E}\) (1), we further select \(\bm{C}\) as \(B\)’s downstream node. As \(\bm{C}\) is already a Primary Term, we conclude our graph search, determining that the ultimate Primary Term for \(A\) is \(\bm{C}\), with the graph search path being \(A \rightarrow B \rightarrow \bm{C}\), rather than \(A \rightarrow B \rightarrow \bm{E}\).

Through this method, we can still acquire a dictionary consistent with the Foundational CGS, as illustrated below:

\begin{align*}
A &\rightarrow \bm{C} \\
B &\rightarrow \bm{C} \\
D &\rightarrow \bm{C} \\
\bm{C} &\rightarrow \bm{C} \\
\bm{E} &\rightarrow \bm{E} \\
\end{align*}

The pseudocode for the Weighted CGS Algorithm is as follows:

\begin{algorithm}[H]
\caption{Weighted CGS Algorithm}
\begin{algorithmic}[1]
\State \textbf{Input:} Set of nodes $N$, Coreference relationships $R$, Weights $W$ of directed edges
\State \textbf{Output:} Dictionary $D$ storing the ultimate Primary Terms

\State Initialize the dictionary $D$.
\State Construct the Coreference Primary Term Graph (CPTG) $G$ from the given $N$, $R$, and $W$.

\For{each node $n$ in $G$}
    \If{$n$ is not a Primary Term}
        \State Find the ultimate Primary Term $p$ for $n$ using weighted graph search.
        \If{$p$ exists}
            \State Add the mapping $n \rightarrow p$ to $D$.
        \EndIf
    \Else
        \State Add the self-referencing mapping $n \rightarrow n$ to $D$.
    \EndIf
\EndFor

\State \textbf{For searching:}
\State Given any term $t$, query $D$ to ascertain whether a corresponding Primary Term exists.
\If{$t$ exists in $D$}
    \State Retrieve the corresponding Primary Term $p$ from $D$.
\EndIf

\end{algorithmic}
\end{algorithm}

%% file: refs.bib
@article{serafini2021s,
  title     = {What’s in a Name? Drug Nomenclature and Medicinal Chemistry Trends using INN Publications},
  author    = {Serafini, Marta and Cargnin, Sarah and Massarotti, Alberto and Tron, Gian Cesare and Pirali, Tracey and Genazzani, Armando A},
  journal   = {Journal of Medicinal Chemistry},
  volume    = {64},
  number    = {8},
  pages     = {4410--4429},
  year      = {2021},
  publisher = {ACS Publications}
}

@article{morgan1965generation,
  title     = {The generation of a unique machine description for chemical structures-a technique developed at chemical abstracts service.},
  author    = {Morgan, Harry L},
  journal   = {Journal of chemical documentation},
  volume    = {5},
  number    = {2},
  pages     = {107--113},
  year      = {1965},
  publisher = {ACS Publications}
}

@book{ZhongHuaBenCao1999,
  title={Zhong Hua Ben Cao},
  author={{National Administration of Traditional Chinese Medicine)}},
  publisher={Shanghai Scientific and Technical Publishers},
  year={1999},
}

@book{chp2020vol1,
  title={Pharmacopoeia of the People's Republic of China: 2020 Edition: Volume I},
  author={{Chinese Pharmacopoeia Commission}},
  year={2020},
  publisher={China Medical Science Press}
}

@book{chp2020vol4,
  title={Pharmacopoeia of the People's Republic of China: 2020 Edition: Volume IV},
  author={{Chinese Pharmacopoeia Commission}},
  year={2020},
  publisher={China Medical Science Press}
}

@book{chp2015vol1,
  title={Pharmacopoeia of the People's Republic of China: 2015 Edition: Volume I},
  author={{Chinese Pharmacopoeia Commission}},
  year={2015},
  publisher={China Medical Science Press}
}

@article{kim2023pubchem,
  title={PubChem 2023 update},
  author={Kim, Sunghwan and Chen, Jie and Cheng, Tiejun and Gindulyte, Asta and He, Jia and He, Siqian and Li, Qingliang and Shoemaker, Benjamin A and Thiessen, Paul A and Yu, Bo and others},
  journal={Nucleic Acids Research},
  volume={51},
  number={D1},
  pages={D1373--D1380},
  year={2023},
  publisher={Oxford University Press}
}

@article{harvey2015re,
  title={The re-emergence of natural products for drug discovery in the genomics era},
  author={Harvey, Alan L and Edrada-Ebel, RuAngelie and Quinn, Ronald J},
  journal={Nature reviews drug discovery},
  volume={14},
  number={2},
  pages={111--129},
  year={2015},
  publisher={Nature Publishing Group UK London}
}

@article{atanasov2021natural,
  title={Natural products in drug discovery: advances and opportunities},
  author={Atanasov, Atanas G and Zotchev, Sergey B and Dirsch, Verena M and Supuran, Claudiu T},
  journal={Nature reviews Drug discovery},
  volume={20},
  number={3},
  pages={200--216},
  year={2021},
  publisher={Nature Publishing Group UK London}
}

@article{newman2020natural,
  title={Natural products as sources of new drugs over the nearly four decades from 01/1981 to 09/2019},
  author={Newman, David J and Cragg, Gordon M},
  journal={Journal of natural products},
  volume={83},
  number={3},
  pages={770--803},
  year={2020},
  publisher={ACS Publications}
}

@article{lautie2020unraveling,
  title={Unraveling plant natural chemical diversity for drug discovery purposes},
  author={Lautie, Emmanuelle and Russo, Olivier and Ducrot, Pierre and Boutin, Jean A},
  journal={Frontiers in pharmacology},
  volume={11},
  pages={397},
  year={2020},
  publisher={Frontiers Media SA}
}

@article{tu2016artemisinin,
  title={Artemisinin—a gift from traditional Chinese medicine to the world (Nobel lecture)},
  author={Tu, Youyou},
  journal={Angewandte Chemie International Edition},
  volume={55},
  number={35},
  pages={10210--10226},
  year={2016},
  publisher={Wiley Online Library}
}

@article{pan2014historical,
  title={Historical perspective of traditional indigenous medical practices: the current renaissance and conservation of herbal resources},
  author={Pan, Si-Yuan and Litscher, Gerhard and Gao, Si-Hua and Zhou, Shu-Feng and Yu, Zhi-Ling and Chen, Hou-Qi and Zhang, Shuo-Feng and Tang, Min-Ke and Sun, Jian-Ning and Ko, Kam-Ming and others},
  journal={Evidence-based complementary and alternative medicine},
  volume={2014},
  year={2014},
  publisher={Hindawi}
}

@article{sen2017revival,
  title={Revival, modernization and integration of Indian traditional herbal medicine in clinical practice: Importance, challenges and future},
  author={Sen, Saikat and Chakraborty, Raja},
  journal={Journal of traditional and complementary medicine},
  volume={7},
  number={2},
  pages={234--244},
  year={2017},
  publisher={Elsevier}
}

@misc{china2016,
    author = {{The State Council Information Office of the People's Republic of China}},
    title = {Traditional Chinese Medicine in China},
    url = {http://english.scio.gov.cn/2017-01/17/content_40621689.htm},
    year = {2016}
}

@misc{openai2023gpt4,
      title={GPT-4 Technical Report}, 
      author={OpenAI},
      year={2023},
      eprint={2303.08774},
      archivePrefix={arXiv},
      primaryClass={cs.CL}
}

@article{yang2023ai,
  title={AI becomes a masterbrain scientist},
  author={YANG, Zijie and WANG, Yukai and ZHANG, Lijing},
  journal={bioRxiv},
  pages={2023--04},
  year={2023},
  publisher={Cold Spring Harbor Laboratory}
}

@article{brown2020language,
  title={Language models are few-shot learners},
  author={Brown, Tom and Mann, Benjamin and Ryder, Nick and Subbiah, Melanie and Kaplan, Jared D and Dhariwal, Prafulla and Neelakantan, Arvind and Shyam, Pranav and Sastry, Girish and Askell, Amanda and others},
  journal={Advances in neural information processing systems},
  volume={33},
  pages={1877--1901},
  year={2020}
}

@misc{NetworkX,
  author = {{NetworkX Developers}},
  title = {NetworkX},
  year = {2023},
  publisher = {GitHub},
  journal = {GitHub repository},
  howpublished = {\url{https://github.com/networkx/networkx}},
}

@misc{LangChain,
  author = {{LangChain Developers}},
  title = {LangChain},
  year = {2023},
  publisher = {GitHub},
  journal = {GitHub repository},
  howpublished = {\url{https://github.com/langchain-ai/langchain}},
}

@article{wei2022chain,
  title={Chain-of-thought prompting elicits reasoning in large language models},
  author={Wei, Jason and Wang, Xuezhi and Schuurmans, Dale and Bosma, Maarten and Xia, Fei and Chi, Ed and Le, Quoc V and Zhou, Denny and others},
  journal={Advances in Neural Information Processing Systems},
  volume={35},
  pages={24824--24837},
  year={2022}
}

@misc{Jieba,
  author = {{Jieba Developers}},
  title = {Jieba},
  year = {2020},
  journal = {GitHub repository},
  howpublished = {\url{https://github.com/fxsjy/jieba}},
}

@inproceedings{li2022prompt,
  title={Prompt-driven neural machine translation},
  author={Li, Yafu and Yin, Yongjing and Li, Jing and Zhang, Yue},
  booktitle={Findings of the Association for Computational Linguistics: ACL 2022},
  pages={2579--2590},
  year={2022}
}

@inproceedings{Transformer,
 author = {Ashish Vaswani and
Noam Shazeer and
Niki Parmar and
Jakob Uszkoreit and
Llion Jones and
Aidan N. Gomez and
Lukasz Kaiser and
Illia Polosukhin},
 booktitle = {NeurIPS2017},
 pages = {5998--6008},
 title = {Attention is All you Need},
 year = {2017}
}

@article{GPTonMT,
  author       = {Amr Hendy and
                  Mohamed Abdelrehim and
                  Amr Sharaf and
                  Vikas Raunak and
                  Mohamed Gabr and
                  Hitokazu Matsushita and
                  Young Jin Kim and
                  Mohamed Afify and
                  Hany Hassan Awadalla},
  title        = {How Good Are {GPT} Models at Machine Translation? {A} Comprehensive
                  Evaluation},
  journal      = {CoRR},
  volume       = {abs/2302.09210},
  year         = {2023},
  url          = {https://doi.org/10.48550/arXiv.2302.09210},
}

@article{Survey_Hallu,
  author       = {Ziwei Ji and
                  Nayeon Lee and
                  Rita Frieske and
                  Tiezheng Yu and
                  Dan Su and
                  Yan Xu and
                  Etsuko Ishii and
                  Yejin Bang and
                  Andrea Madotto and
                  Pascale Fung},
  title        = {Survey of Hallucination in Natural Language Generation},
  journal      = {{ACM} Comput. Surv.},
  volume       = {55},
  number       = {12},
  pages        = {248:1--248:38},
  year         = {2023},
  url          = {https://doi.org/10.1145/3571730},
  doi          = {10.1145/3571730},
  bibsource    = {dblp computer science bibliography, https://dblp.org}
}

@article{peng2023check,
  title={Check your facts and try again: Improving large language models with external knowledge and automated feedback},
  author={Peng, Baolin and Galley, Michel and He, Pengcheng and Cheng, Hao and Xie, Yujia and Hu, Yu and Huang, Qiuyuan and Liden, Lars and Yu, Zhou and Chen, Weizhu and others},
  journal={arXiv preprint arXiv:2302.12813},
  year={2023}
}

@article{nakano2021webgpt,
  title={Webgpt: Browser-assisted question-answering with human feedback},
  author={Nakano, Reiichiro and Hilton, Jacob and Balaji, Suchir and Wu, Jeff and Ouyang, Long and Kim, Christina and Hesse, Christopher and Jain, Shantanu and Kosaraju, Vineet and Saunders, William and others},
  journal={arXiv preprint arXiv:2112.09332},
  year={2021}
}

@article{touvron2023llama,
  title={Llama 2: Open foundation and fine-tuned chat models},
  author={Touvron, Hugo and Martin, Louis and Stone, Kevin and Albert, Peter and Almahairi, Amjad and Babaei, Yasmine and Bashlykov, Nikolay and Batra, Soumya and Bhargava, Prajjwal and Bhosale, Shruti and others},
  journal={arXiv preprint arXiv:2307.09288},
  year={2023}
}

@misc{CommonMark,
  title = {CommonMark},
  year = {2023},
  howpublished = {\url{https://commonmark.org/}},
}

@misc{mdast,
  author = {{mdast Developers}},
  title = {mdast: Markdown Abstract Syntax Tree},
  year = {2023},
  url = {https://github.com/syntax-tree/mdast},
  note = {GitHub repository}
}

@article{leonti2017traditional,
  title={Traditional Mediterranean and European herbal medicines},
  author={Leonti, Marco and Verpoorte, Robert},
  journal={Journal of ethnopharmacology},
  volume={199},
  pages={161--167},
  year={2017},
  publisher={Elsevier}
}

@article{bultum2019etm,
  title={ETM-DB: integrated Ethiopian traditional herbal medicine and phytochemicals database},
  author={Bultum, Lemessa Etana and Woyessa, Assefa Mussa and Lee, Doheon},
  journal={BMC complementary and alternative medicine},
  volume={19},
  pages={1--11},
  year={2019},
  publisher={Springer}
}

@article{mahomoodally2013traditional,
  title={Traditional medicines in Africa: an appraisal of ten potent African medicinal plants},
  author={Mahomoodally, M Fawzi and others},
  journal={Evidence-based complementary and alternative medicine},
  volume={2013},
  year={2013},
  publisher={Hindawi}
}

@article{naghizadeh2020unaprod,
  title={UNaProd: a universal natural product database for Materia Medica of Iranian traditional medicine},
  author={Naghizadeh, Ayeh and Hamzeheian, Donya and Akbari, Shaghayegh and Mohammadi, Fahimeh and Otoufat, Tohid and Asgari, Saeme and Zarei, Azadeh and Noroozi, Samane and Nasiri, Najmeh and Salamat, Mahdi and others},
  journal={Evidence-based complementary and alternative medicine},
  volume={2020},
  year={2020},
  publisher={Hindawi}
}

@article{joos2012herbal,
  title={Herbal medicine in primary healthcare in Germany: the Patient's perspective},
  author={Joos, Stefanie and Glassen, Katharina and Musselmann, Berthold and others},
  journal={Evidence-Based Complementary and Alternative Medicine},
  volume={2012},
  year={2012},
  publisher={Hindawi}
}

@article{dominguez2015mexican,
  title={Mexican Traditional Medicine: Traditions of yesterdey and Phytomedicines for Tomorrow},
  author={Dom{\'\i}nguez, Fabiola and Alonso-Castro, Angel Josabad and Anaya, Maricruz and Gonz{\'a}lez-Trujano, Ma Eva and Salgado-Ceballos, Hermelinda and Orozco-Su{\'a}rez, Sandra},
  journal={Therapeutic Medicinal Plants: From Lab to the Market},
  pages={10--46},
  year={2015},
  publisher={CRC Press Boca Raton, Florida}
}

@article{dutra2016medicinal,
  title={Medicinal plants in Brazil: Pharmacological studies, drug discovery, challenges and perspectives},
  author={Dutra, Rafael C and Campos, Maria M and Santos, Adair RS and Calixto, Jo{\~a}o B},
  journal={Pharmacological research},
  volume={112},
  pages={4--29},
  year={2016},
  publisher={Elsevier}
}

@article{motoo2011traditional,
  title={Traditional Japanese medicine, Kampo: its history and current status},
  author={Motoo, Yoshiharu and Seki, Takashi and Tsutani, Kiichiro},
  journal={Chinese Journal of Integrative Medicine},
  volume={17},
  number={2},
  pages={85--87},
  year={2011},
  publisher={Springer}
}

@article{park2012traditional,
  title={Traditional medicine in China, Korea, and Japan: a brief introduction and comparison},
  author={Park, Hye-Lim and Lee, Hun-Soo and Shin, Byung-Cheul and Liu, Jian-Ping and Shang, Qinghua and Yamashita, Hitoshi and Lim, Byungmook and others},
  journal={Evidence-based complementary and alternative medicine},
  volume={2012},
  year={2012},
  publisher={Hindawi}
}

@article{azaizeh2010traditional,
  title={Traditional Arabic and Islamic medicine, a re-emerging health aid},
  author={Azaizeh, Hassan and Saad, Bashar and Cooper, Edwin and Said, Omar and others},
  journal={Evidence-Based Complementary and Alternative Medicine},
  volume={7},
  pages={419--424},
  year={2010},
  publisher={Hindawi}
}

@book{inp2010vol3,
  title={Indian Pharmacopoeia: 2010 Edition: Volume III},
  author={{Indian Pharmacopoeia Commission}},
  year={2010},
  publisher={The Indian Pharmacopoeia Commission, Indian Pharmacopoeia Laboratory, Government of India, Ministry of Health \& Family Welfare}
}

@misc{dayiorg,
    author = {{China Medical Information Platform}},
    title = {China Medical Information Platform},
    url = {www.dayi.org.cn},
}

@article{chen2018characterization,
  title={Characterization of the chemical space of known and readily obtainable natural products},
  author={Chen, Ya and Garcia de Lomana, Marina and Friedrich, Nils-Ole and Kirchmair, Johannes},
  journal={Journal of chemical information and modeling},
  volume={58},
  number={8},
  pages={1518--1532},
  year={2018},
  publisher={ACS Publications}
}

@article{jang1946ch,
  title={Ch'ang Shan, a Chinese antimalarial herb},
  author={Jang, Chang-Shaw and Fu, FY and Wang, CY and Huang, KC and Lu, G and Chou, TC},
  journal={Science},
  volume={103},
  number={2663},
  pages={59--59},
  year={1946},
  publisher={American Association for the Advancement of Science}
}

@article{zhang1973preliminary,
  title={Preliminary clinical observations of 6 cases of leukemia treated by “Ailin solution”},
  author={Zhang, T and Zhang, P and Wang, S and Han, T},
  journal={Medicine and Pharmacy of Heilongjiang},
  pages={66--67},
  year={1973}
}

@book{leigh2011principles,
  title={Principles of chemical nomenclature: a guide to IUPAC recommendations},
  author={Leigh, Geoffrey J},
  year={2011},
  publisher={Royal Society of Chemistry}
}

@misc{nmm-0016,
  author = {ShennongAlpha},
  title = {ShennongAlpha Knowledge: NMM-0016, Curcuma wenyujin Rhizome Freshly-sliced Cleaned},
  url = {https://shennongalpha.westlake.edu.cn/en-zh/knowledge/nmm-0016},
  note = {Accessed: 2024-05-01},
}

@misc{nmm-000b,
  author = {ShennongAlpha},
  title = {ShennongAlpha Knowledge: NMM-000B, Ephedra sinica Stem-herbaceous Segmented and Aquafried-honey},
  url = {https://shennongalpha.westlake.edu.cn/en-zh/knowledge/nmm-000b},
  note = {Accessed: 2024-05-01},
}

@misc{nmm-0006,
  author = {ShennongAlpha},
  title = {ShennongAlpha Knowledge: NMM-0006, Ephedra equisetina vel intermedia vel sinica Stem-herbaceous},
  url = {https://shennongalpha.westlake.edu.cn/en-zh/knowledge/nmm-0006},
  note = {Accessed: 2024-05-01},
}

@misc{nmm-01yf,
  author = {ShennongAlpha},
  title = {ShennongAlpha Knowledge: NMM-01YF, Taraxacum unspecified Herb},
  url = {https://shennongalpha.westlake.edu.cn/en-zh/knowledge/nmm-01yf},
  note = {Accessed: 2024-05-01},
}

@misc{Taraxacum,
  author = {{Catalogue of Life}},
  title = {Taraxacum},
  url = {https://www.catalogueoflife.org/data/taxon/7SSF},
  note = {Accessed: 2024-05-01},
}

@misc{nmm-0003,
  author = {ShennongAlpha},
  title = {ShennongAlpha Knowledge: NMM-0003, Ephedra sinica Stem-herbaceous},
  url = {https://shennongalpha.westlake.edu.cn/en-zh/knowledge/nmm-0003},
  note = {Accessed: 2024-05-01},
}

@misc{nmm-000g,
  author = {ShennongAlpha},
  title = {ShennongAlpha Knowledge: NMM-000G, Ephedra sinica Root},
  url = {https://shennongalpha.westlake.edu.cn/en-zh/knowledge/nmm-000g},
  note = {Accessed: 2024-05-01},
}

@misc{nmm-0020,
  author = {ShennongAlpha},
  title = {ShennongAlpha Knowledge: NMM-0020, Solidago decurrens Herb},
  url = {https://shennongalpha.westlake.edu.cn/en-zh/knowledge/nmm-0020},
  note = {Accessed: 2024-05-01},
}

@misc{nmm-001l,
  author = {ShennongAlpha},
  title = {ShennongAlpha Knowledge: NMM-001L, Panax ginseng Rhizome and Root},
  url = {https://shennongalpha.westlake.edu.cn/en-zh/knowledge/nmm-001l},
  note = {Accessed: 2024-05-01},
}

@misc{nmm-012w,
  author = {ShennongAlpha},
  title = {ShennongAlpha Knowledge: NMM-012W, Tetradium ruticarpum Fruit},
  url = {https://shennongalpha.westlake.edu.cn/en-zh/knowledge/nmm-012w},
  note = {Accessed: 2024-05-01},
}

@misc{nmm-00ac,
  author = {ShennongAlpha},
  title = {ShennongAlpha Knowledge: NMM-00AC, Cremastra appendiculata vel Pleione bulbocodioides vel yunnanensis Pseudobulb},
  url = {https://shennongalpha.westlake.edu.cn/en-zh/knowledge/nmm-00ac},
  note = {Accessed: 2024-05-01},
}

@misc{nmm-0004,
  author = {ShennongAlpha},
  title = {ShennongAlpha Knowledge: NMM-0004, Ephedra intermedia Stem-herbaceous},
  url = {https://shennongalpha.westlake.edu.cn/en-zh/knowledge/nmm-0004},
  note = {Accessed: 2024-05-01},
}

@misc{nmm-0005,
  author = {ShennongAlpha},
  title = {ShennongAlpha Knowledge: NMM-0005, Ephedra equisetina Stem-herbaceous},
  url = {https://shennongalpha.westlake.edu.cn/en-zh/knowledge/nmm-0005},
  note = {Accessed: 2024-05-01},
}

@misc{nmm-009x,
  author = {ShennongAlpha},
  title = {ShennongAlpha Knowledge: NMM-009X, Crataegus pinnatifida Fruit},
  url = {https://shennongalpha.westlake.edu.cn/en-zh/knowledge/nmm-009x},
  note = {Accessed: 2024-05-01},
}

@misc{nmm-009y,
  author = {ShennongAlpha},
  title = {ShennongAlpha Knowledge: NMM-009Y, Crataegus pinnatifida var major Fruit},
  url = {https://shennongalpha.westlake.edu.cn/en-zh/knowledge/nmm-009y},
  note = {Accessed: 2024-05-01},
}

@misc{nmm-005x,
  author = {ShennongAlpha},
  title = {ShennongAlpha Knowledge: NMM-005X, Ziziphus jujuba var spinosa Seed},
  url = {https://shennongalpha.westlake.edu.cn/en-zh/knowledge/nmm-005x},
  note = {Accessed: 2024-05-01},
}

@misc{nmm-001s,
  author = {ShennongAlpha},
  title = {ShennongAlpha Knowledge: NMM-001S, Panax ginseng Leaf},
  url = {https://shennongalpha.westlake.edu.cn/en-zh/knowledge/nmm-001s},
  note = {Accessed: 2024-05-01},
}

@misc{nmm-01mz,
  author = {ShennongAlpha},
  title = {ShennongAlpha Knowledge: NMM-01MZ, Vincetoxicum pycnostelma Rhizome and Root},
  url = {https://shennongalpha.westlake.edu.cn/en-zh/knowledge/nmm-01mz},
  note = {Accessed: 2024-05-01},
}

@misc{nmm-01mx,
  author = {ShennongAlpha},
  title = {ShennongAlpha Knowledge: NMM-01MX, Vincetoxicum pycnostelma Rhizome},
  url = {https://shennongalpha.westlake.edu.cn/en-zh/knowledge/nmm-01mx},
  note = {Accessed: 2024-05-01},
}

@misc{nmm-01my,
  author = {ShennongAlpha},
  title = {ShennongAlpha Knowledge: NMM-01MY, Vincetoxicum pycnostelma Root},
  url = {https://shennongalpha.westlake.edu.cn/en-zh/knowledge/nmm-01my},
  note = {Accessed: 2024-05-01},
}

@misc{nmm-003g,
  author = {ShennongAlpha},
  title = {ShennongAlpha Knowledge: NMM-003G, Zingiber officinale Rhizome},
  url = {https://shennongalpha.westlake.edu.cn/en-zh/knowledge/nmm-003g},
  note = {Accessed: 2024-05-01},
}

@misc{nmm-003m,
  author = {ShennongAlpha},
  title = {ShennongAlpha Knowledge: NMM-003M, Zingiber officinale Rhizome Fresh},
  url = {https://shennongalpha.westlake.edu.cn/en-zh/knowledge/nmm-003m},
  note = {Accessed: 2024-05-01},
}

@misc{nmm-000t,
  author = {ShennongAlpha},
  title = {ShennongAlpha Knowledge: NMM-000T, Curcuma wenyujin Rhizome},
  url = {https://shennongalpha.westlake.edu.cn/en-zh/knowledge/nmm-000t},
  note = {Accessed: 2024-05-01},
}

@misc{nmm-0015,
  author = {ShennongAlpha},
  title = {ShennongAlpha Knowledge: NMM-0015, Curcuma wenyujin Rhizome Freshly-sliced},
  url = {https://shennongalpha.westlake.edu.cn/en-zh/knowledge/nmm-0015},
  note = {Accessed: 2024-05-01},
}

@misc{nmm-024c,
  author = {ShennongAlpha},
  title = {ShennongAlpha Knowledge: NMM-024C, Fritillaria thunbergii Bulb Zhejiang},
  url = {https://shennongalpha.westlake.edu.cn/en-zh/knowledge/nmm-024c},
  note = {Accessed: 2024-05-14},
}

@misc{nmm-0001,
  author = {ShennongAlpha},
  title = {ShennongAlpha Knowledge: NMM-0001, Artemisia annua Part-aerial},
  url = {https://shennongalpha.westlake.edu.cn/en-zh/knowledge/nmm-0001},
  note = {Accessed: 2024-05-01},
}

@misc{nmm-0002,
  author = {ShennongAlpha},
  title = {ShennongAlpha Knowledge: NMM-0002, Artemisia annua Part-aerial Segmented},
  url = {https://shennongalpha.westlake.edu.cn/en-zh/knowledge/nmm-0002},
  note = {Accessed: 2024-05-01},
}

@misc{nmm-00a1,
  author = {ShennongAlpha},
  title = {ShennongAlpha Knowledge: NMM-00A1, Crataegus pinnatifida Fruit Cleaned and Stirfried-golden},
  url = {https://shennongalpha.westlake.edu.cn/en-zh/knowledge/nmm-00a1},
  note = {Accessed: 2024-05-01},
}

@misc{nmm-00a3,
  author = {ShennongAlpha},
  title = {ShennongAlpha Knowledge: NMM-00A3, Crataegus pinnatifida Fruit Cleaned and Stirfried-charred},
  url = {https://shennongalpha.westlake.edu.cn/en-zh/knowledge/nmm-00a3},
  note = {Accessed: 2024-05-01},
}

@misc{nmm-003l,
  author = {ShennongAlpha},
  title = {ShennongAlpha Knowledge: NMM-003L, Zingiber officinale Rhizome Cleaned and Stirfried-sand},
  url = {https://shennongalpha.westlake.edu.cn/en-zh/knowledge/nmm-003l},
  note = {Accessed: 2024-05-01},
}

@misc{nmm-01xx,
  author = {ShennongAlpha},
  title = {ShennongAlpha Knowledge: NMM-01XX, Talc},
  url = {https://shennongalpha.westlake.edu.cn/en-zh/knowledge/nmm-01xx},
  note = {Accessed: 2024-05-01},
}

@misc{nmm-01xz,
  author = {ShennongAlpha},
  title = {ShennongAlpha Knowledge: NMM-01XZ, Talc Pulverized},
  url = {https://shennongalpha.westlake.edu.cn/en-zh/knowledge/nmm-01xz},
  note = {Accessed: 2024-05-01},
}

@misc{nmm-0022,
  author = {ShennongAlpha},
  title = {ShennongAlpha Knowledge: NMM-0022, Erycibe obtusifolia Stem},
  url = {https://shennongalpha.westlake.edu.cn/en-zh/knowledge/nmm-0022},
  note = {Accessed: 2024-05-01},
}

@misc{nmm-0024,
  author = {ShennongAlpha},
  title = {ShennongAlpha Knowledge: NMM-0024, Erycibe obtusifolia vel schmidtii Stem},
  url = {https://shennongalpha.westlake.edu.cn/en-zh/knowledge/nmm-0024},
  note = {Accessed: 2024-05-01},
}

@misc{nmm-00d7,
  author = {ShennongAlpha},
  title = {ShennongAlpha Knowledge: NMM-00D7, Ligustrum lucidum Fruit},
  url = {https://shennongalpha.westlake.edu.cn/en-zh/knowledge/nmm-00d7},
  note = {Accessed: 2024-05-01},
}

@misc{nmm-008w,
  author = {ShennongAlpha},
  title = {ShennongAlpha Knowledge: NMM-008W, Prunus mume Fruit},
  url = {https://shennongalpha.westlake.edu.cn/en-zh/knowledge/nmm-008w},
  note = {Accessed: 2024-05-01},
}
